\documentclass[times,twocolumn,final]{elsarticle}
\usepackage{medima}
\usepackage{framed,multirow}
\usepackage{stackengine}
\newcommand\barbelow[1]{\stackunder[1.2pt]{$#1$}{\rule{.8ex}{.075ex}}}
\usepackage{epsfig}
\usepackage{amsmath}
\usepackage{lipsum}
\usepackage{enumitem}
\usepackage[toc, page]{appendix}

\usepackage{booktabs}
\usepackage{amssymb}
\newcommand{\E}{\mathbb{E}}
\usepackage{latexsym}
\usepackage{array, makecell}
\usepackage{bm}
\usepackage{xspace}
\usepackage{nicematrix}
\usepackage{subfigure}
\usepackage[T1]{fontenc}
\usepackage{graphicx}
\usepackage{multirow}
\usepackage{microtype}
\usepackage{url}
\usepackage{xcolor}
\usepackage[colorlinks = true,
            linkcolor = tealblue,
            urlcolor  = tealblue,
            citecolor = tealblue,
            anchorcolor = tealblue]{hyperref}

\DeclareMathOperator*{\argmax}{arg\,max\xspace}

\DeclareMathOperator*{\trace}{tr}

\definecolor{newcolor1}{rgb}{.0, .502, .675}
\definecolor{candyAppleRed}{RGB}{255 8 0}

\makeatletter
\AtBeginDocument{\def\@citecolor{newcolor1}}
\AtBeginDocument{\def\@linkcolor{newcolor1}}
\AtBeginDocument{\def\@anchorcolor{newcolor1}}
\AtBeginDocument{\def\@filecolor{newcolor1}}
\AtBeginDocument{\def\@urlcolor{newcolor1}}
\AtBeginDocument{\def\@menucolor{newcolor1}}
\AtBeginDocument{\def\@pagecolor{newcolor1}}
\makeatother

\newif\ifcb
\cbtrue

\newcolumntype{?}{!{\vrule width 1pt}}
\usepackage{color, colortbl}
\definecolor{Gray}{gray}{0.9}
\definecolor{newcolor}{rgb}{.8,.349,.1}
\newcommand{\KL}{\mathrm{KL}}

\journal{}%Medical Image Analysis}

\begin{document}

\verso{Chen \textit{et~al.}}

\begin{frontmatter}

\title{Unsupervised learning of spatially varying regularization for diffeomorphic image registration}

\author[1]{Junyu \snm{Chen}\corref{cor1}}
\author[2]{Shuwen \snm{Wei}}
\author[3]{Yihao \snm{Liu}}
\author[2]{Zhangxing \snm{Bian}}
\author[4]{Yufan \snm{He}}
\author[2]{Aaron \snm{Carass}}
\author[1]{Harrison \snm{Bai}}
\author[1]{Yong \snm{Du}}
\address[1]{Department of Radiology and Radiological Science, Johns Hopkins School of Medicine, MD, USA}
\address[2]{Image Analysis and Communications Laboratory, Department of Electrical and Computer Engineering, Johns Hopkins University, MD, USA}
\address[3]{Department of Electrical and Computer Engineering, Vanderbilt University, TN, USA}
\address[4]{NVIDIA Corporation, Santa Clara, CA, USA}
\cortext[cor1]{Corresponding author. E-mail address: 
 jchen245@jhmi.edu.}
\received{xxxx}
\finalform{xxxx}
\accepted{xxxx}
\availableonline{xxxx}
\communicated{xxxx}

\begin{abstract}
Spatially varying regularization accommodates the deformation variations that may be necessary for different anatomical regions during deformable image registration. Historically, optimization-based registration models have harnessed spatially varying regularization to address anatomical subtleties. However, most modern deep learning-based models tend to gravitate towards spatially invariant regularization, wherein a homogenous regularization strength is applied across the entire image, potentially disregarding localized variations. In this paper, we propose a hierarchical probabilistic model that integrates a prior distribution on the deformation regularization strength, enabling the end-to-end learning of a spatially varying deformation regularizer directly from the data. The proposed method is straightforward to implement and easily integrates with various registration network architectures. Additionally, automatic tuning of hyperparameters is achieved through Bayesian optimization, allowing efficient identification of optimal hyperparameters for any given registration task. Comprehensive evaluations on publicly available datasets demonstrate that the proposed method significantly improves registration performance and enhances the interpretability of deep learning-based registration, all while maintaining smooth deformations. Our code is freely available at \url{http://bit.ly/3BrXGxz}.
\end{abstract}

\begin{keyword}
%% Keywords
\KWD Deformable Image Registration \sep Deep Neural Networks \sep Spatially Varying Regularization
\end{keyword}

\end{frontmatter}

%\linenumbers


\section{Introduction}
Deformable image registration~(DIR) is an important component in many medical imaging applications. The objective is to estimate a smooth deformation that aligns a moving image with a fixed image to maximize their similarity. Optimization-based DIR methods, such as LDDMM~\citep{beg2005computing}, SyN~\citep{avants2008symmetric}, Demons~\citep{thirion1998image, vercauteren2009diffeomorphic}, and FLASH~\citep{zhang2019fast}, are built on rigorous mathematical frameworks and have traditionally been the preferred methods for accomplishing DIR tasks. However, these methods present certain limitations: they often require time-consuming optimization steps, and their instance-specific optimization scheme makes it challenging to integrate auxiliary or prior information into the registration process.

In recent years, the advancement of deep neural networks~(DNNs) and their success in processing image data have sparked increased interest in developing DNN-based methods for medical image registration.
Deep learning-based registration methods train a DNN on an image dataset to optimize a global objective function, usually in the form of a similarity measure, with a regularizer added to enforce the spatial smoothness of the deformation.
Such methods circumvent the laborious iterative optimization procedure, enabling faster and sometimes superior deformation mappings in a fraction of the time required by traditional methods.
For a comprehensive review of recent developments in deep learning-based medical image registration, the readers are directed to~\citep{chen2024survey}.
In the remainder of the paper, we use the terms "learning-based methods" to denote "deep learning-based methods" for ease of discussion.

In learning-based DIR, most existing methods adopt the objective function of traditional methods to serve as loss functions for training DNNs. Mathematically, this is expressed as:
\begin{equation}
\mathcal{E}(\pmb{m}\circ\phi, \pmb{f}) + \lambda \mathcal{R}(\phi),
\end{equation}
where $\phi$ is a deformation field, the first term, $\mathcal{E}(\pmb{m}\circ\phi, \pmb{f})$, is a similarity metric between the deformed moving image, denoted as $\pmb{m}\circ\phi$, and the fixed image $\pmb{f}$. The term $\mathcal{R}(\phi)$ acts as a regularizer that encourages the spatial smoothness of the deformation field $\phi$. The hyperparameter $\lambda$ controls the degree of regularization and is often designated as a constant and \textit{spatially-invariant} value~\citep{balakrishnan2019voxelmorph, dalca2019unsupervised, de2017end, chen2021vitvnet,chen2022transmorph,liu2022coordinate, kim2021cyclemorph,liu2024vector,chen2024survey}. However, employing a constant regularization strength assumes that all anatomical structures necessitate a similar level of regularization, neglecting the possibility that optimal regularization could vary based on the particular images at hand. For example, the anatomical structure of an individual may better resemble certain individuals than others~\citep{simpson2012probabilistic}. As such, a constant regularization strategy might not always be the most effective. To address this, some methods have been proposed to condition $\lambda$ within the DNN framework, allowing for the adjustment of the regularization strength during the test time~\citep{mok2021conditional, hoopes2021hypermorph}. Although this offers a degree of adaptability, the challenge of determining the optimal regularization strength for the given image still persists.
Additionally, these works applied a \textit{spatially-invariant} regularization strength to the entire image, which does not account for the variations of deformation that may be necessary for different regions of the image~\citep{niethammer2019metric}.
An example of this can be seen in brain scan registration, where the brain ventricles can vary in size between different patients, leading to different scales of deformation in the ventricles compared to other parts of the brain~\citep{niethammer2019metric}. Similarly, when inhale-to-exhale images of the lung are aligned, larger lung deformations are anticipated than those of surrounding tissue~\citep{shen2019region}. \textcolor{black}{Such lung motion commonly involves sliding motion at anatomical interfaces. For instance, at the boundary between lung tissue and adjacent rib bones, the lung may exhibit discontinuous sliding movements, while the ribs remain nearly stationary or move in the opposite direction.} Furthermore, less structured anatomical regions, such as white matter in brain MRI or liver and lungs in CT, may require smoother deformation to counteract noise. In contrast, regions with more intricate structures may require more complex deformations for accurate registration. 

\textcolor{black}{Efforts have been made to develop registration models that accommodate complex and spatially varying anatomical motion. These include optimization-based approaches~\citep{schmidt2009slipping, schmidt2012estimation, yin2010lung, risser2013piecewise, stefanescu2004grid, vialard2014spatially, gerig2014spatially, kabus2006variational}, as well as deep learning-based methods that either allow discontinuous motion at organ boundaries by leveraging anatomical label maps~\citep{chen2021deep, chen2022joint, lu2023discontinuity}, or learn spatially varying regularization directly from data~\citep{niethammer2019metric, shen2019region, wang2023conditional}.
Among traditional optimization-based methods, many have been tailored for lung registration to account for sliding motion. These approaches typically enforce deformation smoothness in the normal direction at lung boundaries while permitting tangential discontinuities, thereby avoiding penalties on physiologically plausible motions~\citep{schmidt2009slipping, schmidt2012estimation, yin2010lung, risser2013piecewise, pace2013locally}. In contrast, learning-based methods that incorporate anatomical label maps focus on enabling discontinuities at tissue interfaces but generally apply a spatially invariant regularization across the image~\citep{chen2021deep, chen2022joint, lu2023discontinuity}. Only a few deep learning-based methods have explicitly modeled spatially varying regularization within a DNN framework~\citep{niethammer2019metric, shen2019region, wang2023conditional}.}
In \citep{niethammer2019metric}, Niethammer \emph{et al.} proposed using DNNs to predict locally adaptive weights for multi-Gaussian kernels, but their method requires pre-setting the variance of the kernels and is not effective for end-to-end registration networks.
\citet{shen2019region} expanded on the work of \citep{niethammer2019metric} by introducing an end-to-end training scheme to learn a spatially varying regularizer and the initial momentum in a spatiotemporal velocity setting.
This method has the ability to track the deformation of regions and estimate a unique regularizer for each time point. However, the implementation of this method is not straightforward and cannot be seamlessly integrated with existing DNN-based methods.

\textcolor{black}{In this paper, we propose a novel DNN-based method for end-to-end learning of spatially varying regularization in deformable image registration. Our motivation is primarily technical, aiming to enhance flexibility beyond conventional global deformation regularization by enabling the registration network to adapt regularization strength at the voxel level. This spatially varying regularization is learned directly from image data in an unsupervised manner, automatically balancing local deformation smoothness with local image similarity. Although we do not explicitly incorporate anatomical priors to enforce specific tissue-level deformation constraints, our approach implicitly accommodates complex deformation patterns, such as sliding motions, by observing voxel-level similarity information. Hyperparameters controlling the regularization strength are efficiently tuned using a validation dataset through an automated Bayesian optimization pipeline, thus avoiding computationally expensive grid searches.}

The main contributions of this work are as follows:
\begin{itemize}
    \item We propose a spatially varying regularization method for diffeomorphic image registration through unsupervised learning. This approach leverages a hierarchical probabilistic formulation of the image registration problem, enabling a novel training scheme for the registration network. At test time, the method generates a subject-specific regularization map that controls regularization levels at the voxel level.
    \item \textcolor{black}{The proposed method improves the registration performance of learning-based image registration, enables flexible handling of complex local motions, and provides an additional layer of interpretability.}
    \item The proposed method is easy to implement and seamlessly integrates as a plug-and-play module with other learning-based registration models.
    \item Comprehensive evaluations on publicly available datasets across various anatomical regions demonstrate that the proposed method improves the registration accuracy of the DNN backbone while preserving low deformation irregularities.
    \item To promote reproducible research, we provide the community with access to the source code and pre-trained models, which are publicly available at: \url{http://bit.ly/3BrXGxz}.
\end{itemize} 

The remainder of the paper is organized as follows. Section~\ref{sec:background} discusses related work. Section~\ref{sec:methods} describes the proposed methodology. The experimental setup, implementation details, and datasets used in this study are discussed in Sect.~\ref{sec:experiments}. Section~\ref{sec:results} presents the experimental results. The findings drawn from these results are discussed in Sect.~\ref{sec:discussion}, and Sect.~\ref{sec:conclusion} concludes the paper.

\section{Related Works}
\label{sec:background}
\subsection{Diffusion Regularizer}
\label{sec:diff_reg}
In learning-based image registration, the diffusion regularizer is often used to impose deformation smoothness~\citep{balakrishnan2019voxelmorph, dalca2019unsupervised, kim2021cyclemorph, chen2022transmorph, chen2021vitvnet, liu2022coordinate}. It is expressed as:
\begin{equation}
\label{eqn:diff}
    \mathcal{R}(\phi) = \sum_{\mathbf{p}\in\Omega}\vert\nabla \pmb{u}(\mathbf{p})\vert^2,
\end{equation}
where $\pmb{u}$ denotes the displacement field such that the deformation field $\phi=\pmb{id}+\pmb{u}$, $\pmb{id}$ denotes the identity, $\mathbf{p}=(p^x,p^y,p^z)$ is the voxel location, and $\Omega\subset\mathbb{R}^3$ represents the 3D spatial domain. The operator, $\nabla$, computes the spatial gradients of each component of the displacement field.
Specifically, $\nabla\pmb{u}(\mathbf{p})=(\frac{\partial\pmb{u}(\mathbf{p})}{\partial x}, \frac{\partial\pmb{u}(\mathbf{p})}{\partial y}, \frac{\partial\pmb{u}(\mathbf{p})}{\partial z})$, which can be approximated through finite differences.
For example, with the forward difference approximation, we have $\frac{\partial\pmb{u}(\mathbf{p})}{\partial x}\approx\pmb{u}(p^{x}+1, p^{y}, p^{z})-\pmb{u}(p^{x}, p^{y}, p^{z})$. As shown in \citet{dalca2019learning, dalca2019unsupervised}, this regularizer can be derived from the maximum a posteriori estimation of the variable $\pmb{u}$ assuming that the prior distribution on $\pmb{u}$ is a multivariate normal distribution with mean $\pmb{\mu}=\mathbf{0}$ and covariance $\mathbf{\Sigma}$, expressed as: $p(\pmb{u}) \propto \mathcal{N}(\pmb{u};\barbelow{\mathbf{0}},\mathbf{\Sigma})$, where $\mathbf{\Sigma}^{-1}=\pmb{\Lambda}_{\pmb{u}}=\lambda \mathbf{L}$, and $\mathbf{L}=\mathbf{D}-\mathbf{A}$ is the Laplacian of a neighborhood graph defined on the image grid. Here, $\mathbf{D}$ is the degree matrix and $\mathbf{A}$ is the adjacency matrix. The parameter $\lambda$ controls the smoothness of the displacement. The logarithm of this prior distribution can be simplified to Eqn. \ref{eqn:diff} as follows:
\begin{equation}
\label{eqn:diff_derive}
\begin{split}
   \log p(\pmb{u})&=\frac{1}{2}\log\vert\mathbf{\Sigma}^{-1}\vert-\frac{1}{2} \pmb{u}^\intercal\mathbf{\Sigma}^{-1}\pmb{u} + \text{const.}\\
   &=\frac{1}{2}\log\vert\pmb{\Lambda}_{\pmb{u}}\vert-\frac{1}{2} \pmb{u}^\intercal\pmb{\Lambda}_{\pmb{u}}\pmb{u} + \text{const.}\\
   %&\propto-
   &= -\frac{\lambda}{2} \pmb{u}^\intercal\mathbf{L}\pmb{u} + \text{const.} \\
   &= -\frac{\lambda}{2}\sum_{\mathbf{p}\in\Omega}\vert\nabla \pmb{u}(\mathbf{p})\vert^2 + \text{const.},
\end{split}  
\end{equation}
by noting the fact that $\log\vert\pmb{\Lambda}_{\pmb{u}}\vert$ is a constant, where $\vert\pmb{\Lambda}_{\pmb{u}}\vert = \mathrm{det}(\pmb{\Lambda}_{\pmb{u}})$. The parameter $\lambda$ regulates the smoothness of the deformation by varying the covariance of the normal distribution. When $\lambda$ increases, the covariance $\mathbf{\Sigma}$ decreases, resulting in greater similarity among neighboring displacements. This can also be understood as applying a larger variance Gaussian kernel to the displacement field, leading to a smoother deformation.

\subsection{Probabilistic Inference of Regularization}
\cite{simpson2012probabilistic} introduced a hierarchical probabilistic model that adaptively determines the level of regularization for each subject based on available data, rather than manually tuning the regularization hyperparameter for each subject. It uses a free-form deformation approach, where the transformation is characterized by a set of control points, $\pmb{\omega}$. The registration process is mathematically represented as $\pmb{f}=\pmb{m}\circ\phi(\pmb{\omega})+\pmb{\epsilon}$. Here, the likelihood of the fixed image is described by a normal distribution, and prior distributions are placed on both the transformation parameters $\pmb{\omega}$ and the image noise $\pmb{\epsilon}$. Specifically, the prior for $\pmb{\omega}$ is modeled using a multivariate normal distribution, as detailed in Sect.~\ref{sec:diff_reg}, while a gamma distribution is used for the prior modeling of $\pmb{\epsilon}$. Additionally, a scalar spatial precision parameter $\lambda$, which arises from the multivariate normal distribution of $\pmb{\omega}$, is modeled with a prior gamma distribution and serves to control the regularization level. 
%For inference, the method employs a mean-field variational Bayes approach, optimizing an objective function that comprises the log-likelihood and the negative Kullback-Leibler divergence between the approximate posterior distributions and the assigned priors. 
This optimization-based method is a full probabilistic model that also captures the uncertainty associated with the transformation parameters. Although it allows for the inference of different regularization parameters for each given image pair, the regularization remains spatially invariant across the image.
In contrast, in this paper, we aim to infer varying levels of regularization strength at the voxel level using a deep learning framework.
This enables us to implement spatially varying regularization and offers a significantly faster computational time compared to traditional optimization-based methods.

\subsection{Learning Spatially Varying Regularizer}
\label{sec:spt_vary_gauss}
\subsubsection{Unsupervised Learning for Regularization Weights}
In \citet{niethammer2019metric}, a method for spatially varying regularization through metric learning was proposed. The approach involves learning locally adaptive weights for multiple Gaussian kernels with different standard deviations, in the form: $\sum_{i=0}^{N-1} w_i(\mathbf{p}) G_i$.
Here, $w_i(\mathbf{p})$ represents the weight of the $i^{th}$ Gaussian kernel $G$ at the voxel location $\mathbf{p}$, and they satisfy $\sum_{i=0}^{N-1} w_i(\mathbf{p})=1$.
A neural network $f$, with parameters $\theta$, predicts these adaptive weights using input from the image pair and the initial momentum $m_0$ generated by a traditional registration method.
This can be represented as $[w_0,\ldots,w_{N-1}]=f_\theta(\pmb{m}, \pmb{f}, m_0)$. Then, the deformation is smoothed by convolving the weighted multi-Gaussian kernel with $m_0$:
\begin{equation}
\sum_{i=0}^{N-1} \sqrt{w_i(\mathbf{p})} \int_\mathbf{q}G_i(\mathbf{p}-\mathbf{q})\sqrt{w_i(\mathbf{q})}m_0(\mathbf{q})d\mathbf{q}.
\end{equation}
During network training, two loss functions were placed over $w$'s to promote smoothness in the momentum and adaptive weights.
These loss functions included an optimal mass transport~(OMT) loss and a total variational~(TV) loss. The OMT loss forces the network to prioritize the use of the Gaussian kernel with the largest variance, while the TV loss encourages weight changes coinciding with image edges. Although successful in implementing spatially varying regularization, this method is not suitable for end-to-end registration networks because it requires initial momentum for network input. Additionally, the number and standard deviations of the Gaussian kernels, as well as the weighting parameters for OMT and TV losses, are hyperparameters that need extensive training cycles to optimize manually.

\subsubsection{Label Map-Guided Regularization Methods}
\textcolor{black}{With the rise of deep learning, supervised and semi-supervised approaches have leveraged anatomical label maps, which delineate structural boundaries, to guide image registration methods by imposing spatially varying regularization and facilitating discontinuous deformations~\citep{chen2021deep, chen2022joint, lu2023discontinuity, wang2023conditional}. Specifically, methods proposed in~\citep{chen2021deep, chen2022joint} explicitly partition the deformation estimation by having the registration network generate separate deformation fields for each anatomical region, which are subsequently masked by corresponding anatomical label maps and combined to produce the final deformation. In contrast, \citet{wang2023conditional} applied spatially varying regularization by assigning different regularization weights to displacement gradients based on anatomical labels. Additionally, \citet{lu2023discontinuity} used anatomical labels to identify interfaces between anatomical structures and selectively relaxed regularization constraints at these boundaries.}

%A parallel independent work~\citep{wang2023conditional} also explores spatially varying regularization, where varying regularization weights are allocated to different anatomical regions. Their approach presents a different mechanism that conditions a predefined weight map in the network architecture. To implement their approach, anatomical label maps of the input images are required, and the weights are directly imposed on the displacement gradients. 

\textcolor{black}{While these approaches have demonstrated the potential of spatially varying regularization, their major limitation lies in the reliance on anatomical label maps. Although obtaining segmentations has become relatively easier due to well-established segmentation networks (e.g., \citep{isensee2021nnu, huo20193d, he2024vista3d, ma2024segment}), or even joint segmentation-registration learning frameworks (e.g., \citep{xu2019deepatlas, khor2023anatomically, he2020deep}), these methods are inherently limited by the accuracy and completeness of the anatomical labels. Coarse or incomplete label maps, which may fail to delineate finer substructures, restrict these methods' ability to apply suitable local regularization.}
Conversely, in our approach, we aim to determine the spatially varying regularization weight directly from the data without leaning on anatomical label maps.
\begin{figure}[t]
\begin{center}
\includegraphics[width=0.25\textwidth]{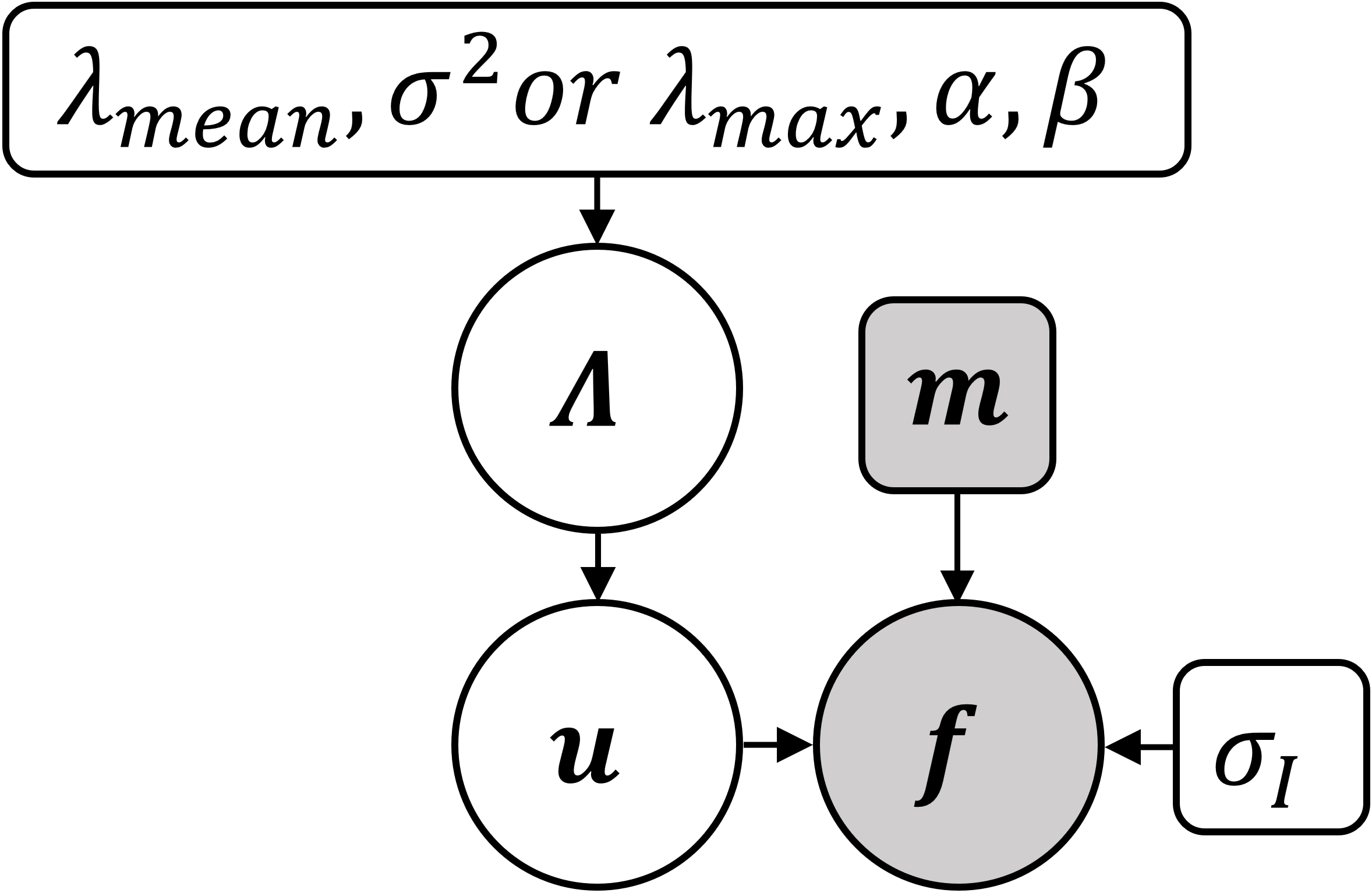}
\end{center}
   \caption{The probabilistic dependencies of the model parameters are graphically illustrated in the diagram. Random variables are depicted as circles, while rounded squares represent model parameters. Shaded quantities indicate observed elements \textit{at test time}, and the plate indicates replication given different samples.}
\label{fig:inference_fig}
\end{figure}

\section{Methods}
\label{sec:methods}
We define the moving and fixed images, respectively, as $\pmb{m}$ and $\pmb{f}$. These images are defined in a 3D spatial domain, denoted as $\pmb{m}, \pmb{f}\in\mathbb{R}^{H\times W\times D}$. Here, we aim to develop a spatially varying deformation regularizer for end-to-end learning of deformable image registration, rather than a spatially invariant regularization for the entire deformation as used in almost all existing learning-based registration methods (e.g., \cite{balakrishnan2019voxelmorph, dalca2019unsupervised, kim2021cyclemorph, chen2022transmorph, mok2021conditional, liu2022coordinate, chen2024survey}). \textcolor{black}{Conventional global regularizers (e.g., diffusion regularization as discussed in Sect.~\ref{sec:diff_reg}) impose uniform smoothness constraints, whereas our method introduces spatially varying regularization by modeling its distribution through probabilistic assumptions.} \textcolor{black}{Specifically, we formulate the learning of spatially varying regularization under a maximum-a-posteriori (MAP) framework, resulting in a novel loss function with a new variable---the spatial weight volume---that controls local regularization strength at each voxel. This spatial weight volume is estimated using a lightweight decoder head integrated into the registration DNN. The associated hyperparameters are efficiently selected for each registration task using Bayesian optimization.} The details of this approach are expounded in the subsequent sections.

\begin{figure}[!b]
    \centering
    \includegraphics[width=0.35\textwidth]{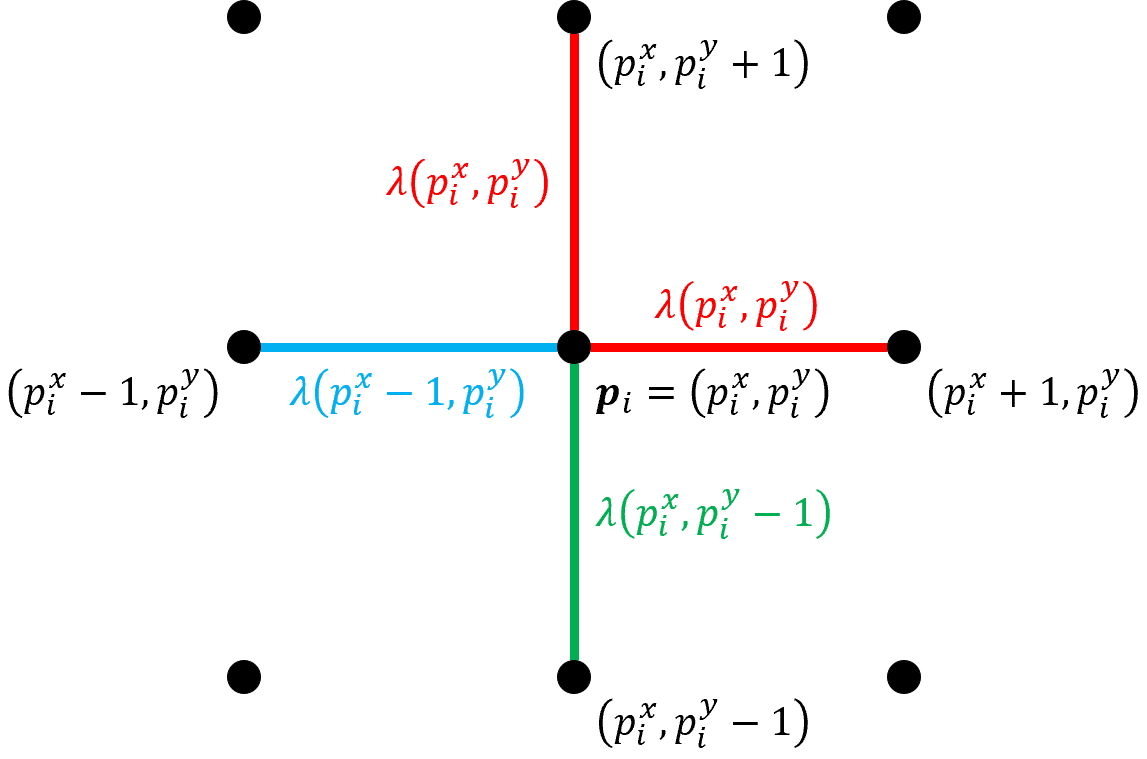}
    \caption{Weighted graph Laplacian on a 2D image grid centered at $\mathbf{p}_i$. Red edges represent forward adjacency. Blue and green edges indicate backward adjacency. Each color corresponds to a distinct edge weight.}
    \label{fig:laplacian}
\end{figure}

\begin{figure*}[t]
\begin{center}
\includegraphics[width=0.7\textwidth]{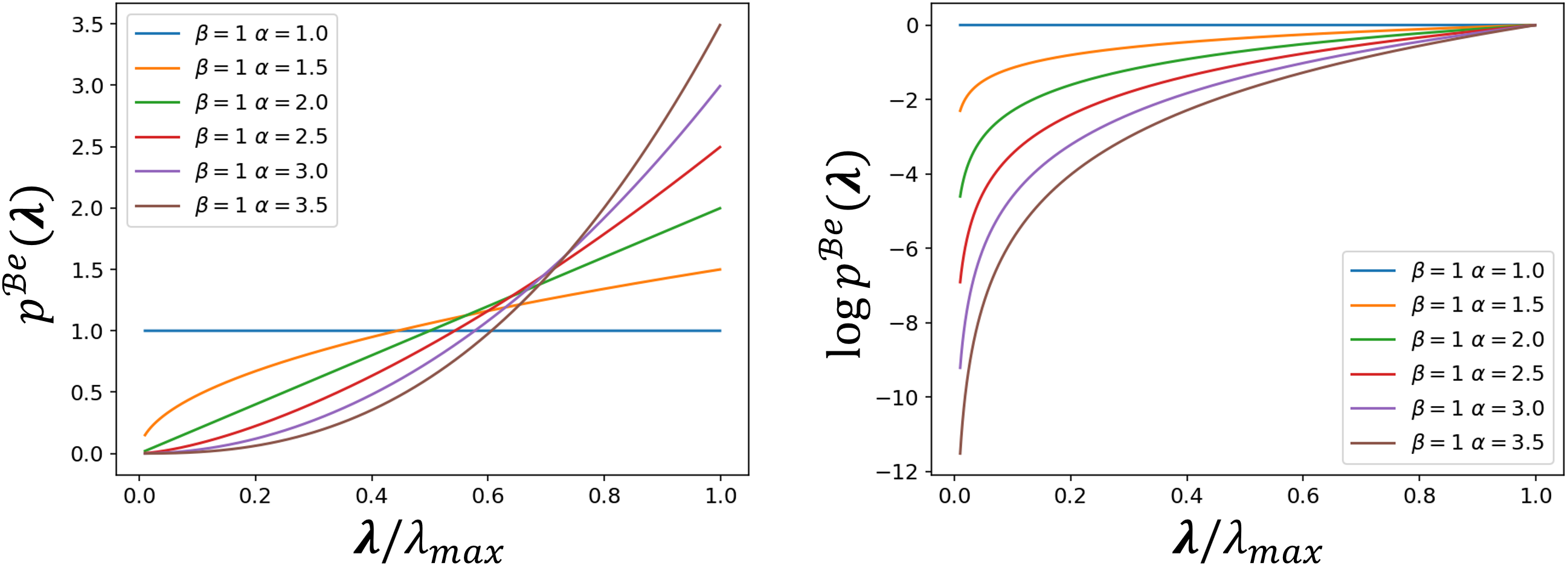}
\end{center}
   \caption{The plots of $p^{\mathcal{B}e}(\pmb{\lambda})$ and $\log p^{\mathcal{B}e}(\pmb{\lambda})$ under the assumption of beta distribution, with varying shape parameter $\alpha$.}
\label{fig:beta}
\end{figure*}

\subsection{Hierarchical Model}
Figure~\ref{fig:inference_fig} depicts the graphical illustration of the proposed hierarchical model. Our objective is to compute the joint posterior probability of registration, denoted by $p(\pmb{u}, \pmb{\Lambda}|\pmb{f}; \pmb{m})$, through a MAP framework. Here, $\pmb{u}$ is a $N_c \times 3$ matrix that symbolizes the displacement field, while $\pmb{\Lambda}$ is a $N_c\times N_c$ matrix that functions as the spatial precision (i.e., the inverse covariance). $N_c$ denotes the total number of voxels (i.e., $N_c=HWD$). We model $\pmb{\Lambda}$ as a \textit{weighted} graph Laplacian matrix defined over the image grid. This matrix encodes spatial variations in the degree of regularization by assigning weights to the edges between adjacent voxels. 

Leveraging Bayes' theorem, we can express this joint posterior distribution as being directly proportional to the product of the hyperprior distribution of the spatial precision (\emph{i.e.}, $p(\pmb{\Lambda})$), the population distribution of the displacement field given the spatial regularization (\emph{i.e.}, $p(\pmb{u}|\pmb{\Lambda})$), and the data likelihood (\emph{i.e.}, $p(\pmb{f}|\pmb{u}, \pmb{\Lambda}; \pmb{m})$). Mathematically, this relationship can be expressed as:
\textcolor{black}{
\begin{equation}
\begin{split}
p(\pmb{u}, \pmb{\Lambda} | \pmb{f}; \pmb{m}) &= \frac{p(\pmb{u}, \pmb{\Lambda}; \pmb{m}) p(\pmb{f} | \pmb{u}, \pmb{\Lambda}; \pmb{m})}{p(\pmb{f}; \pmb{m})} \\
&= \frac{p(\pmb{\Lambda}; \pmb{m}) p(\pmb{u} | \pmb{\Lambda}; \pmb{m}) p(\pmb{f} | \pmb{u}, \pmb{\Lambda}; \pmb{m})}{p(\pmb{f}; \pmb{m})} \\
&\propto p(\pmb{\Lambda}) p(\pmb{u} | \pmb{\Lambda}) p(\pmb{f} | \pmb{u}, \pmb{\Lambda}; \pmb{m}),
\end{split}
\end{equation}We simplify $p(\pmb{\Lambda}; \pmb{m})$ and $p(\pmb{u} | \pmb{\Lambda}; \pmb{m})$ to $p(\pmb{\Lambda})$ and $p(\pmb{u} | \pmb{\Lambda})$ because $\pmb{m}$ is an input, not inferred or marginalized over, and we assume these distributions are \textit{static}, not functions of $\pmb{f}$ and $\pmb{m}$. Our DNN predicts a point estimate (i.e., the mode) of this posterior for $\pmb{u}$ and $\pmb{\Lambda}$ from $\pmb{f}$ and $\pmb{m}$, achieving data-specific regularization via the posterior. We jointly optimize the logarithm of the joint posterior over $\pmb{u}$ and $\pmb{\Lambda}$ using a Type-II MAP approach.} This leads to:
\begin{equation}
\label{eqn:log_post}
\begin{split}
    \argmax_{\pmb{u},\pmb{\Lambda}}\log p(\pmb{u}, \pmb{\Lambda}|\pmb{f}; \pmb{m})=\argmax_{\pmb{u},\pmb{\Lambda}}\log &p(\pmb{\Lambda})+\log p(\pmb{u}|\pmb{\Lambda})\\
    &+\log p(\pmb{f}|\pmb{u}, \pmb{\Lambda}; \pmb{m}).
\end{split}
\end{equation}

\textcolor{black}{In conventional MAP estimation, the optimization problem must be solved separately for each image pair. In contrast, we adopt the delta variational Bayes approach, also known as amortized MAP~\citep{sonderby2016amortised, shu2018amortized}, which trains a neural network to directly predict the optimal MAP estimates for a given input pair. In our formulation, the variational joint posterior $q_\phi(\pmb{u},\pmb{\Lambda}|\pmb{f};\pmb{m})$ is parameterized by a neural network, and the training objective minimizes KL divergences to match the joint prior $p(\pmb u, \pmb\Lambda)=p(\pmb u|\pmb\Lambda)p(\pmb\Lambda)$. A detailed derivation is provided in \ref{sec:derivation_DVB}.}

In this work, we model the data likelihood using the Boltzmann distribution:
\begin{equation}
    p(\pmb{f}|\pmb{u}, \pmb{\Lambda}; \pmb{m}) \propto \mathrm{exp}(-\sigma_I \mathcal{D}_{NCC}(\pmb{f}, \pmb{m}\circ\phi_{\pmb{u}})),
\end{equation}
where $\sigma_I$ is a user-defined hyperparameter (we set $\sigma_I=1$ throughout this study), $\phi_{\pmb{u}}$ denotes the deformation field that warps $\pmb{m}$ to $\pmb{f}$, and $\mathcal{D}_{NCC}(\cdot)$ denotes the negative normalized cross-correlation~(NCC).
The population and hyperprior distributions are described in the following sections.

\subsection{Population Distribution}
\label{sec:pop_dist}
In accordance with \citet{dalca2019unsupervised} and \citet{simpson2012probabilistic} and Sect.~\ref{sec:diff_reg}, we model the population distribution, $p(\pmb{u}|\pmb{\Lambda})$, by using a multivariate normal distribution:
\begin{equation}
    \pmb{u}|\pmb{\Lambda}\sim\mathcal{N}(\pmb{u};\barbelow{\mathbf{0}}, \pmb{\Lambda}^{-1})=\frac{|\pmb{\Lambda}|^\frac{3}{2}}{(2\pi)^\frac{3N_c}{2}}\mathrm{exp}(-\frac{1}{2} \trace (\pmb{u}^\intercal\pmb{\Lambda}\pmb{u})),
\end{equation}
and 
\begin{equation}
    \pmb{\Lambda} = \mathbf{D}-\mathbf{A},
\end{equation}
where $\mathbf{D}$ and $\mathbf{A}$ are the degree matrix and the adjacency matrix of a weighted graph Laplacian, respectively, \textcolor{black}{and $|\cdot|$ denotes the matrix determinant.}

A representation of the weighted graph on a 2D image grid centered at $\mathbf{p}_i$ is shown in Fig.~\ref{fig:laplacian}.
The dots show the voxel in the neighborhood of $\mathbf{p}_i$, and the edges show the adjacency and corresponding weights between $\mathbf{p}_i$ and its neighbors.
Specifically, the red edges show the forward adjacency, and the blue and green edges show the backward adjacency. 
The edge weights were designed such that edges with forward adjacency were assigned equal weights, e.g., $\lambda(p_i^x, p_i^y)$, as highlighted in red in Fig.~\ref{fig:laplacian}.
Based on this design, the adjacency matrix $\mathbf{A}$ is defined such that its element $\mathbf{A}_{ij}$ is:
\begin{equation}
  \mathbf{A}_{ij} =
    \begin{cases}
      \lambda(\mathbf{p}_i) & \text{if $\mathbf{p}_j$ is forward adjacent to $\mathbf{p}_i$}\\
      0 & \text{otherwise}
    \end{cases},
\end{equation}
where $\mathbf{p}_i$ and $\mathbf{p}_j$ correspond to voxel locations indexed by $i$ and $j$, respectively, which from 1 to $N_c$. The forward adjacency in 3D denotes the condition that $\mathbf{p}_j$ is adjacent to $\mathbf{p}_i$ in the positive $x$, $y$, or $z$ direction (i.e., $\mathbf{p}_j\in\{(p^{x}_i+1, p^{y}_i, p^{z}_i), (p^{x}_i, p^{y}_i+1, p^{z}_i), (p^{x}_i, p^{y}_i, p^{z}_i+1)\}$). This adjacency design arises from the computation of forward differences for the gradient operation in the diffusion regularizer (Eqn.~\ref{eqn:diff}). Note that the choice of edge weights and adjacency can be adapted for different finite-difference approximations. The diagonal degree matrix $\mathbf{D}$ is similarly defined as:
\begin{equation}
  \mathbf{D}_{ij} =
    \begin{cases}
      \sum_{k=1}^{N_c}\pmb{A}_{ik} & \text{if $i=j$}\\
      0 & \text{otherwise}
    \end{cases}.  
\end{equation}
%where the value 6 arises from considering 6-connected neighbors in a 3D grid. 
In these equations, $\lambda(\mathbf{p})$ represents the regularization weight of the edges that are forward connected with the specific voxel location $\mathbf{p}$, which modulates the regularization strength at that location. With these considerations, we formulate the log-probability as follows:
\begin{equation}
\label{eqn:prior}
\begin{split}
    \log p(\pmb{u}|\pmb{\Lambda}) & \propto 3\log\vert\pmb{\Lambda}\vert-\trace(\pmb{u}^\intercal\pmb{\Lambda}\pmb{u})\\
    &=3\log\vert\pmb{\Lambda}\vert-\sum_{\mathbf{p} \in \Omega} \lambda(\mathbf{p}) |\nabla\pmb{u}(\mathbf{p})|^2,
\end{split}
\end{equation}
\textcolor{black}{where $\vert\pmb{\Lambda}\vert$ denotes the determinant of $\pmb{\Lambda}$,} is related to the covariance of the multivariate normal distribution (as discussed in Sect.~\ref{sec:diff_reg}). A higher value of $\lambda(\mathbf{p})$ virtually results in a stronger Gaussian smoothing at voxel location $\mathbf{p}$. 

\textcolor{black}{In practical implementation, the first term, $3\log\vert\pmb{\Lambda}\vert$, is omitted and this exclusion is justified by three main factors. Firstly, the computation of $\log\vert\pmb{\Lambda}\vert$ is non-trivial, computationally intensive, and prone to numerical instability due to the determinant computation of a large sparse matrix with dimensions $N_c\times N_c$. Secondly, while this term primarily enhances spatial precision (or reduces covariance) in the multivariate normal distribution governing $\pmb{u}$, an informative hyperprior is already imposed on $\pmb{\Lambda}$ to regulate spatial precision (as discussed in Sect.~\ref{sec:hyperprior}), making its inclusion unnecessary. Lastly, the prior elements in $|\pmb{\Lambda}|$ are not independent but coupled, and this coupling is modeled through a multivariate normal distribution, facilitated by the upsampling operation (see Sect.~\ref{sec:hyperprior} for details). Thus, instead of incorporating the higher-order term $\log\vert\pmb{\Lambda}\vert$, we rely on $p(\pmb{\lambda})$, which characterizes the individual elements in $\pmb{\Lambda}$, leading to a more concise and computationally efficient representation of the prior.}

The second term in Eqn.~\ref{eqn:prior} essentially constitutes a spatially varying version of the diffusion regularizer, which can be interpreted as the application of spatially-modulated multi-Gaussian kernels to the deformation --- an approach resonating with the method proposed in \citet{niethammer2019metric} (Sect.~\ref{sec:spt_vary_gauss}).
This can be attributed to the fact that the convolution of multiple Gaussian kernels essentially results in a Gaussian kernel.
However, we argue that the method proposed here is more apt for end-to-end training than the method in \citet{niethammer2019metric}, since we introduce a hyperprior distribution to $\pmb{\Lambda}$ and leverage the DNN to directly predict $\lambda(\mathbf{p})$ --- in other words, the Gaussian covariance --- based on the given data.
Thus, unlike the approach in \citet{niethammer2019metric}, the proposed method obviates the need to predetermine the number and variances of the Gaussians.
\begin{figure*}[t]
\begin{center}
\includegraphics[width=0.95\textwidth]{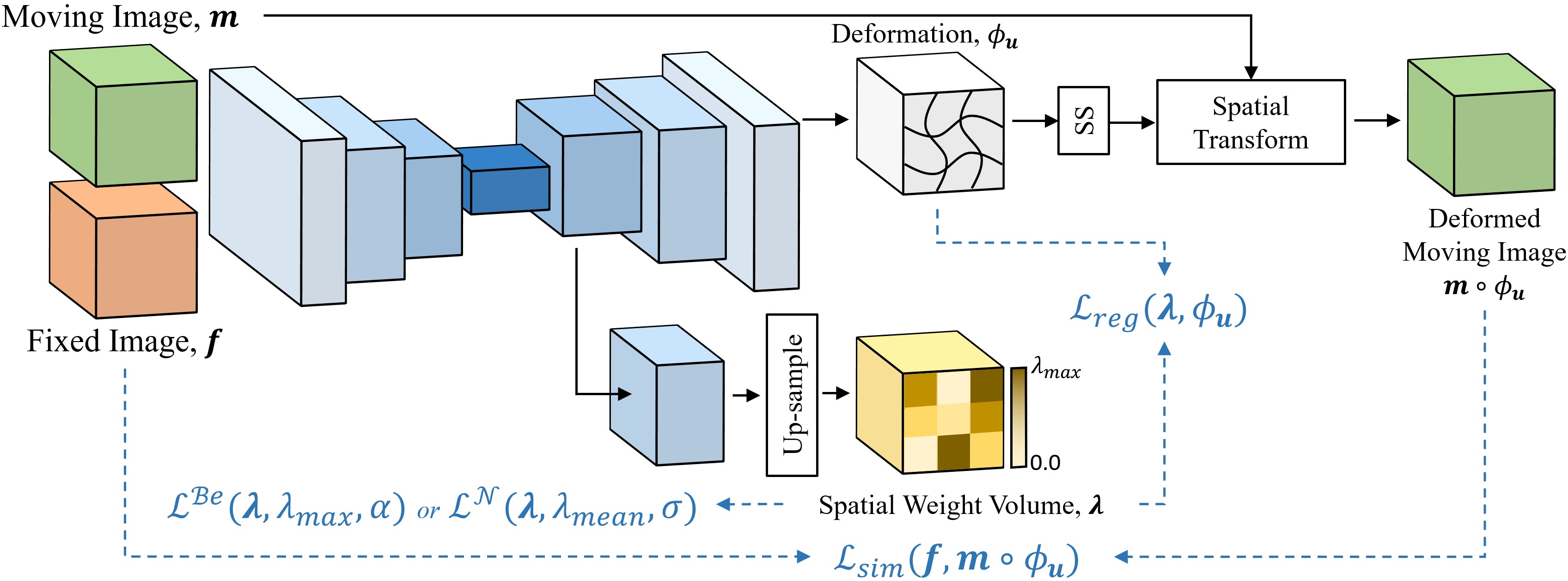}
\end{center}
   \caption{The overall framework of the proposed method for unsupervised learning of spatially varying regularization for diffeomorphic image registration. The method introduces lightweight ConvNet blocks that take the feature maps from the bottleneck layers of the registration network to generate a spatial weight volume $\pmb{\lambda})$, assigning regularization at the voxel level. The spatial weight volume is used to regularize the deformation during training. Additionally, the spatial weight volume is itself regularized by an auxiliary term (i.e., $\mathcal{L}^{\mathcal{N} \mathrm{or} \mathcal{B}e}(\pmb{\lambda})$), promoting smoother deformations when appropriate. While the framework is demonstrated using \texttt{TransMorph} as the registration backbone, it is agnostic to the choice of registration networks and can be easily adapted to other architectures.}
\label{fig:framework}
\end{figure*}
\subsection{Hyperprior Distribution}
\label{sec:hyperprior}
The hyperprior distribution of spatial precision, $\pmb{\Lambda}$, is probabilistically modeled via the scale family of distributions. The underlying intuition is that the strength of regularization across voxel locations adheres to a specific distribution, promoting regularization but permitting weaker regularization strength at certain voxel locations. It is important to note that the Laplacian matrix $\pmb{\Lambda}$ is intrinsically related to the spatial weights $\pmb{\lambda}$. Specifically, $\pmb{\Lambda}$ is a deterministic function of $\pmb{\lambda}$, constructed based on the edge weights represented by $\pmb{\lambda}$ and the graph structure. As a result, modeling the distribution $p(\pmb{\lambda})$ effectively captures the hyperprior distribution $p(\pmb{\Lambda})$. 

In the context of this paper, the prior knowledge for $\pmb{\lambda}$ emerges from the general preference to favor smooth deformations. That is, among all the potential solutions to the ill-posed problem of DIR, those that result in smooth deformations are particularly preferred.
This preference suggests that $\lambda(\mathbf{p})$ is more likely to have larger values to impose overall smoothness in the deformation field. Here, we consider two assumptions that $\pmb{\lambda}$ is either normal- or beta-distributed, although alternative distribution types can also be considered depending on the application.
Furthermore, ensuring spatial smoothness in $\pmb{\lambda}$ is crucial to avert abrupt changes in regularization, which in turn could affect the smoothness of the deformation field.
To impose smoothness in $\pmb{\lambda}$, we integrate a multivariate normal distribution in addition to either the normal or beta distribution to model $p(\pmb{\lambda})$. The resultant hyperprior distribution is expressed as the multiplication of two distributions: the first couples the elements of $\pmb{\lambda}$ together, while the second characterizes the independent attributes of each element within $\pmb{\lambda}$:
\begin{equation}
    p(\pmb{\lambda})\propto\mathcal{N}(\pmb{\lambda};\barbelow{\mathbf{0}},\mathbf{\Sigma}_\lambda) p^{\mathcal{N} \mathrm{or} \mathcal{B}e}(\pmb{\lambda}).
\end{equation}
Analogous to Eqn.~\ref{eqn:diff_derive}, the multivariate normal distribution simplifies to a diffusion regularizer placed over $\pmb{\lambda}$. Consequently, the log-hyperprior can be expressed as:
\begin{equation}
\begin{split}
    \log p(\pmb{\lambda})\propto& -\frac{\eta}{2}\sum_{\mathbf{p}\in\Omega}\vert\nabla \pmb{\lambda}(\mathbf{p})\vert^2 + \log p^{\mathcal{N} \mathrm{or} \mathcal{B}e}(\pmb{\lambda})+ \text{const.}.\\
\end{split}
\end{equation}
In practice, this diffusion regularizer introduces a hyperparameter (i.e., $\eta$) that requires careful tuning. Rather than delving into this complexity, we employ a relaxed version of this regularizer: we estimate $\pmb{\lambda}$ at a lower resolution and then upsampling it using bilinear interpolation. This effectively imparts the necessary smoothness to $\pmb{\lambda}$ without additional complications. The modeling of $p^{\mathcal{N} \mathrm{or} \mathcal{B}e}(\pmb{\lambda})$ is described in subsequent sections.

\subsubsection{Normal Distribution}
We first opt for the scale family of normal distribution to model $\pmb{\lambda}$, owing to its simplicity and also motivated by the conjugate relationship that this normal prior holds with our earlier assumption of a multivariate normal distribution for the population distribution, $p(\pmb{u}|\pmb{\lambda})$. The definition is as follows:
\begin{equation}
\begin{split}
    p^\mathcal{N}(\pmb{\lambda})&\propto\mathcal{N}(\frac{\pmb{\lambda}}{\lambda_{mean}};\barbelow{\mathbf{1}}, \sigma^2) \\
    &=\frac{1}{(\sqrt{2\pi}\sigma)^{N_c}}\exp\left({-\frac{1}{2\sigma^2}(\frac{\pmb{\lambda}}{\lambda_{mean}}-\mathbf{1})^\intercal (\frac{\pmb{\lambda}}{\lambda_{mean}}-\mathbf{1})}\right),
\end{split}
\end{equation}
and
\begin{equation}  
\label{eqn:log_gauss}
\begin{split}
    \log p^{\mathcal{N}}(\pmb{\lambda})
    &\propto -\frac{1}{2\sigma^2}(\frac{\pmb{\lambda}}{\lambda_{mean}}-\mathbf{1})^\intercal (\frac{\pmb{\lambda}}{\lambda_{mean}}-\mathbf{1}) + \text{const.}\\
    &=-\frac{1}{2\sigma^2}\sum_{\mathbf{p} \in \Omega}(\frac{{\lambda}(\mathbf{p})}{\lambda_{mean}}-1)^2 + \text{const.},
\end{split}
\end{equation}
where $\sigma$ is the standard deviation and $\lambda_{mean}$ acts as a scale parameter that determines the overall regularization strength.
Notably, a larger $\lambda_{mean}$ virtually inflates the mean of the normal distribution, yielding larger values of $\lambda(\mathbf{p})$ and, therefore, stronger regularization, and vice versa.
\textcolor{black}{As outlined in Eqn.~\ref{eqn:prior}, $\pmb{\lambda}$ is associated with covariance, hence it cannot have values less than 0, resulting in a one-sided truncated normal distribution for the prior. In practice, we enforce this constraint by applying a ReLU activation, i.e., $\lambda(\pmb{p}) = \text{ReLU}(z(\pmb{p}))$, where $z(\pmb{p})$ is the DNN output. Additionally, we ensure numerical stability during training by selecting $\sigma^2 > 0$ and $\lambda_{mean} > 0$. However, this truncation disrupts the conjugacy with the Gaussian population distribution assumption.
Additionally, the absence of a specific upper limit can cause excessive regularization, particularly at the tail end of the distribution, complicating the selection of optimal values for $\lambda_{mean}$ and $\sigma$.}
In the hyperparameter optimization process, we optimize $\sigma' = \frac{1}{2\sigma^2}$ directly, rather than $\sigma$, to simplify the tuning process.

In the following section, we propose using the beta distribution to model the hyperprior distribution of $\pmb{\lambda}$, which presents a more controlled approach to regularization by defining the upper and lower bounds for the values $\pmb{\lambda}$ can take.

\subsubsection{Beta Distribution}
Our choice of the beta distribution for the hyperprior arises from its ability to facilitate an upper bound for regularization strength, while also providing the flexibility to give a higher likelihood to the user-defined upper bound. Therefore, we adopt the scale family of beta distribution for $\pmb{\lambda}$, which can be mathematically expressed as:
\begin{equation}
\begin{split}
    p^{\mathcal{B}e}(\pmb{\lambda})&\propto\prod_{\mathbf{p}\in\Omega} \mathcal{B}e \left( \frac{\lambda(\mathbf{p})}{\lambda_{max}};\alpha,\beta \right)\\
    &=\prod_{\mathbf{p}\in\Omega}\frac{\Gamma(\alpha+\beta)}{\Gamma(\alpha)\Gamma(\beta)} \left(\frac{\lambda(\mathbf{p})}{\lambda_{max}} \right)^{\alpha-1} \left(1 - \frac{\lambda(\mathbf{p})}{\lambda_{max}} \right)^{\beta-1},
\end{split}
\end{equation}
and
\begin{equation}
\label{eqn:log_beta}
\begin{split}
    &\log p^{\mathcal{B}e}(\pmb{\lambda}) \\
    &\propto\sum_{\mathbf{p} \in \Omega} \left( (\alpha-1) \log \frac{\lambda(\mathbf{p})}{\lambda_{max}}+(\beta-1)  \log(1-\frac{\lambda(\mathbf{p})}{\lambda_{max}}) \right) + \text{const.}
\end{split}
\end{equation}
where $\lambda(\mathbf{p})\in[0, \lambda_{max}]$, $\lambda_{max}>0$ is a hyperparameter that controls the maximum regularization strength by scaling the distribution, $\Gamma(\cdot)$ denotes the Gamma function, and $\alpha$ and $\beta$ are shape parameters that define the shape of the beta distribution. We set $\beta=1$ and choose an $\alpha$ value greater than or equal to 1 (i.e., $\alpha\geq 1$), encouraging a higher regularization strength (i.e., a larger value of lambda) to be more probable. \textcolor{black}{Then, the second term vanishes because $\beta-1=0$. This simplifies the expression to:
\begin{equation}
\label{eqn:beta_dist}
    \log p^{\mathcal{B}e}(\pmb{\lambda})\propto(\alpha-1)\sum_{\mathbf{p} \in \Omega}\log\frac{\lambda(\mathbf{p})}{\lambda_{max}} + \text{const.}. 
\end{equation}}
This specific instance of the beta distribution is also known as a power-function distribution. \textcolor{black}{In practical implementation, a DNN outputs $z(\pmb{p})$, which is passed through a Sigmoid activation to obtain $\frac{\lambda(\mathbf{p})}{\lambda_{max}}$, i.e., $\frac{\lambda(\mathbf{p})}{\lambda_{max}} = \text{Sigmoid}(z(\pmb{p}))$, ensuring that $\lambda(\mathbf{p})$ remains within $[0, \lambda_{max}]$, thereby ensuring numerical stability.} As shown in Fig. \ref{fig:beta}, altering $\alpha$ effectively allows us to modulate the penalty applied to $\lambda(\mathbf{p})$.
Specifically, an $\alpha$ value of 1 yields a uniform distribution, inflicting no penalty, thus allowing $\lambda(\mathbf{p})$ to assume any value within its range. A larger $\alpha$, however, pushes $\lambda(\mathbf{p})$ towards $\lambda_{max}$. Unlike a normal distribution model for $p(\pmb{\lambda})$, the beta distribution extends the advantage of bounding the regularization strength within a specified range, with a minimum of 0, implying no regularization, and a maximum set by $\lambda_{max}$. To simplify hyperparameter optimization, we optimize $\alpha' = \alpha - 1$ instead of directly tuning $\alpha$.

\textcolor{black}{It is also important to note the intrinsic connection between $\log p^{\mathcal{B}e}(\pmb{\lambda})$ (Eqn.~\ref{eqn:beta_dist}) and binary entropy loss. As $\frac{\lambda(\mathbf{p})}{\lambda_{max}}\in[0,1]$ can be viewed as a "probability" map produced by the network, with a target probability map of $1$ and the weight of the loss controlled by $1-\alpha$.} \textcolor{black}{We also note the conceptual similarity between this beta prior and the optimal mass transport (OMT) loss in~\citep{niethammer2019metric}, both of which encourage larger regularization strengths via a log-loss form. In our approach, the loss directly promotes larger values in the spatial weight map, resulting in stronger regularization at each voxel.}

\subsection{Loss Function}
\label{sec:loss}
In this subsection, we elaborate on how the loss function is formulated, following from Eqn.~\ref{eqn:log_post}. The aim is to maximize the log joint posterior, which is equivalent to minimizing its negative. Therefore, the loss function can be expressed as:
\begin{equation}
\label{eqn:loss}
\begin{split}
    \mathcal{L}(\pmb{m}, \pmb{f}, \pmb{u}, \pmb{\lambda})=\mathcal{D}_{NCC}(\pmb{f}, \pmb{m}\circ\phi_{\pmb{u}})+\sum_{\mathbf{p}\in\Omega}& \lambda(\mathbf{p}) |\nabla\pmb{u}(\mathbf{p})|^2\\
    -&\mathcal{L}^{\mathcal{N} \mathrm{or} \mathcal{B}e}(\pmb{\lambda}).
\end{split}
\end{equation}
For the last term, $\mathcal{L}^{\mathcal{N} \mathrm{or} \mathcal{B}e}(\pmb{\lambda})$, we specify it based on either Eqn. \ref{eqn:log_gauss} or Eqn. \ref{eqn:log_beta}, depending on whether a normal or beta distribution model is employed, i.e., $\mathcal{L}^{\mathcal{N} \mathrm{or} \mathcal{B}e}(\pmb{\lambda}) = \log p^{\mathcal{N} \mathrm{or} \mathcal{B}e}(\pmb{\lambda})$.
%the hyperprior $p(\pmb{\lambda})$ considers each $\lambda(\mathbf{p})$ as being independent, whereas the prior elements in $|\pmb{\Lambda}|$ are coupled together. Therefore, it helps simplify the prior by excluding the higher order term in $\log\vert\pmb{\Lambda}\vert$. 

\subsection{Registration Neural Network}
Figure \ref{fig:framework} shows the overall framework of the proposed method. We used a neural network to take in moving and fixed images, $\pmb{m}$ and $\pmb{f}$.
The network outputs a deformation field $\phi_{\pmb{u}}\in\mathbb{R}^{3\times H\times W\times D}$ that warps $\pmb{m}$ to $\pmb{f}$, as well as a spatial weight volume $\pmb{\lambda}\in\mathbb{R}^{H\times W\times D}$. We then used this weight volume to apply spatially varying regularization strength to different voxels through a \textit{weighted diffusion regularizer} (i.e., the second term in Eqn.~\ref{eqn:loss}).
The backbone registration network we employed builds upon our previously proposed \texttt{TransMorph}~\citep{chen2022unsupervised, chen2022transmorph}, which has shown effective results in multiple registration applications. It is important to note that the proposed spatially varying regularization framework is not restricted to this architecture and can be seamlessly extended to other network designs.
The \texttt{TransMorph} network employs a Swin Transformer encoder~\citep{liu2021swin} and a ConvNet decoder, which generates a set of deformation fields. These fields are then composed to form the final deformation field.
As shown in Fig.~\ref{fig:framework}, an additional convolution block comprising three convolutional layers is introduced within the decoder of the \texttt{TransMorph} and applied to the 1/4 resolution branch, resulting in the generation of the weight volume $\pmb{\lambda}$.
The final weight volume is subsequently up-sampled using bilinear interpolation to align with the resolution of the deformation field.
As mentioned in Sect.~\ref{sec:hyperprior}, by generating the weight volume at a lower resolution, we were able to reduce computational workload and introduce spatial smoothness in the weight volume, which aligns with the physical interpretation of these weights.

\subsection{Diffeomorphic Image Registration}
As the proposed network learns to impose spatially varying regularization, certain regions may experience weak or even absent regularization, which could result in less realistic deformations. To address this, we enforce diffeomorphic image registration by interpreting the displacement field $\pmb{u}$, estimated by the neural network, as an approximation of a stationary velocity field $\pmb{v}$. We implement a time-stationary setting by exponentiating this velocity field using the \textit{scaling-and-squaring}~(SS) approach~\citep{arsigny2006log, ashburner2007fast}, mathematically expressed as: 
\begin{equation}
\label{eqn:SS}
    \phi = \exp{(\pmb{v})} = \left(\exp{(2^{-N}\pmb{v})}\right)^{2^N},
\end{equation}
where $\pmb{v}$ is derived from $\pmb{u}$ and used to accumulate transformations in a smooth and invertible way.
We opt for the SS approach for its straightforward implementation. However, the proposed method can also accommodate a time-varying setting if desired.

For the remainder of this paper, the proposed method is denoted as \texttt{TM-SPR} to emphasize the use of the spatially varying regularizer~(SPR).

\subsection{Automatic Hyperparameter Tuning}
The proposed method requires tuning two hyperparameters: $\lambda_{mean}$ and $\sigma'$ when using a Gaussian prior, or $\lambda_{max}$ and $\alpha'$ when using a beta prior. Instead of employing computationally expensive grid search, we employed Bayesian optimization~(BO) through the Tree-Structured Parzen Estimator~(TPE) for hyperparameter optimization~\citep{bergstra2011algorithms, watanabe2023tree}. BO optimizes the objective function (specifically the Dice score in our case) on the validation dataset by building a probabilistic model and using it to determine the next set of hyperparameters to sample. This approach is more efficient than grid search because it selectively explores the hyperparameter space, focusing on regions that are more likely to yield improvements. Furthermore, BO incorporates pruning, which enables early stopping of unpromising trials, further enhancing sample efficiency~\citep{watanabe2023tree, ozaki2020multiobjective}. For this study, we implemented BO using the open source Optuna package~\citep{akiba2019optuna}, which fully automates hyperparameter tuning and analysis.

\begin{figure}[!t]
\begin{center}
\includegraphics[width=0.45\textwidth]{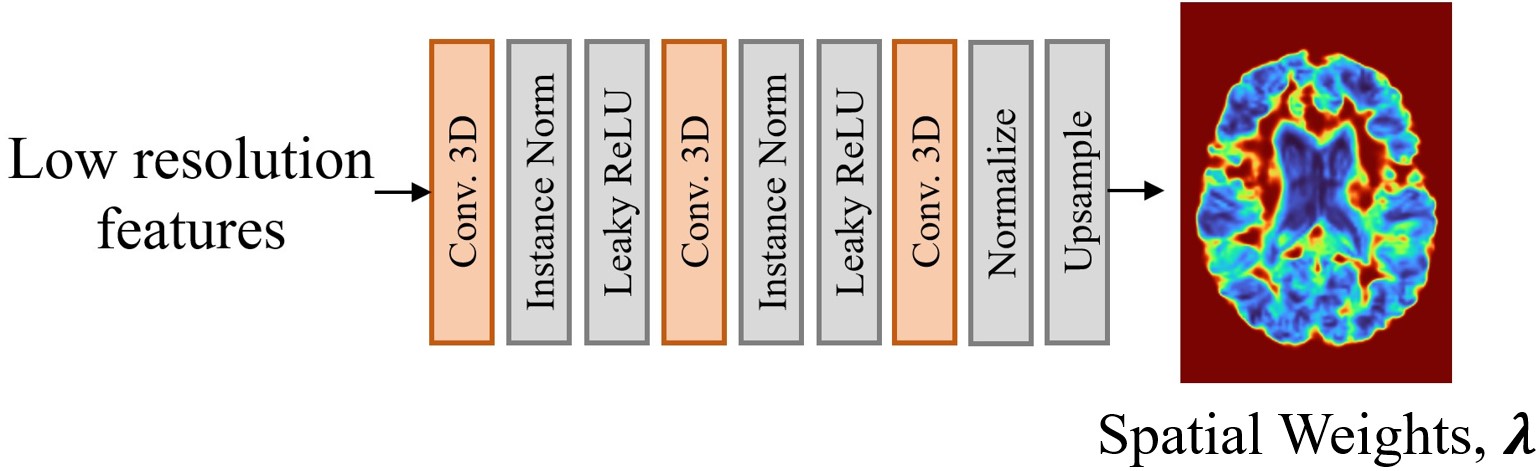}
\end{center}
   \caption{The schematic of the ConvNet block that produces the spatial weight volume, $\pmb{\lambda}$.}
\label{fig:ConvNet_for_w}
\end{figure}

\begin{figure*}[t]
\begin{center}
\includegraphics[width=0.95\textwidth]{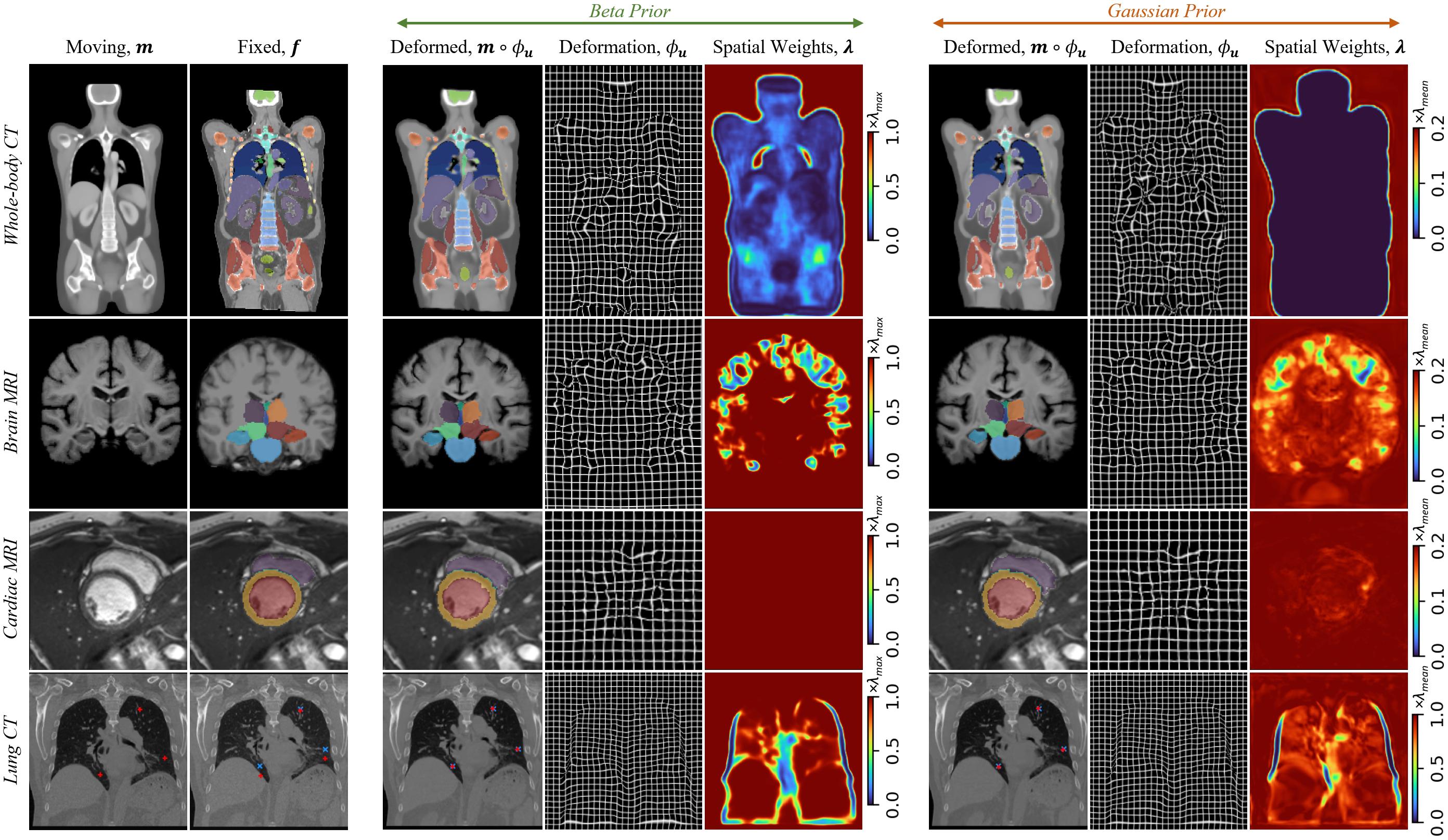}
\end{center}
   \caption{Qualitative results of the proposed method across three registration tasks: atlas-to-subject whole-body CT registration~(autoPET), inter-subject brain MRI registration~(IXI), intra-subject cardiac MRI registration~(ACDC M\&Ms), \textcolor{black}{and intra-subject lung CT registration~(4DCT)} from top to bottom. The left panel displays the moving image and fixed image, with landmarks or selected label maps overlaid on the fixed image. The middle panel illustrates the deformed moving image, deformation field, and spatial weight volume generated by the proposed method using the beta prior. The right panel presents the corresponding results obtained using the Gaussian prior.}
\label{fig:qual_results}
\end{figure*}
\section{Experiments}
\label{sec:experiments}
\subsection{Dataset and Pre-processing}
The proposed registration method was extensively evaluated using three publicly available data sets that cover a broad spectrum of anatomical regions and imaging applications. The descriptions of each dataset, along with the associated pre-processing steps, are provided in the subsequent paragraphs.
\paragraph{IXI} The IXI dataset\footnote{https://brain-development.org/ixi-dataset/} comprises 576 T1-weighted brain MRI images, with a distribution of 403 volumes for training, 58 volumes for validation, and 115 volumes for testing.
Furthermore, a moving image, which was a brain atlas image obtained from \citet{kim2021cyclemorph}, was used for the atlas-to-patient registration task. All images went through a standard preprocessing pipeline through FreeSurfer~\citep{fischl2012freesurfer}.
This involved resampling to a uniform voxel size of $1\times1\times1$ mm, affine registration, skull-stripping, and intensity normalization. After preprocessing, the image volumes were consistently cropped to a size of $160 \times 192 \times 224$. Anatomical label maps that include more than 30 anatomical structures were produced using FreeSurfer to evaluate registration performances. The registration task associated with this dataset is atlas-to-patient registration.

\paragraph{ACDC and M\&Ms} The ACDC dataset~\citep{bernard2018deep} includes 150 subjects, while the M\&Ms dataset~\citep{campello2021multi} comprises 320 subjects, making a total of 470 subjects. 
Each subject is represented by two frames corresponding to the end-diastolic~(ED) and end-systolic~(ES) phases. The datasets provide manual delineations of the left ventricle~(LV) and right ventricle~(RV) blood pools, as well as the left ventricular myocardium~(MYO).
Data were split into training, validation, and testing sets with 259, 61, and 150 subjects, respectively.
All images were resampled to a uniform spatial resolution of $1.5\times1.5\times3.5$ mm and adjusted to a volume of interest~(VOI) of $128\times128\times32$, centered around the heart.
Furthermore, the intensity values of the images were normalized using a min-max normalization to scale between 0 and 1. The primary registration challenge in this dataset involves intra-subject registration between the ED and ES stages.

\paragraph{autoPET} The autoPET dataset~\citep{gatidis2022whole} comprises whole-body PET/CT scans from 900 cancer patients.
However, only CT data were used in this study. Of the total, 628 were allocated for training, 90 for validation, and 178 for testing purposes.
The segmentation of 104 classes in CT images was facilitated by the TotalSegmentor~\citep{wasserthal2022totalsegmentator}.
Our study focused on 103 of these classes, excluding the face, which were subsequently grouped into 29 labels by merging similar structures.
Resampling was first applied to all images, setting the voxel size to $2.8\times2.8\times3.8$ mm. Subsequently, the images were affinely aligned with a common reference, chosen randomly from the patient pool, using the ANTs package~\citep{avants2009advanced}.
The post-alignment images were then cropped to dimensions $192\times192\times256$. Intensity values were truncated to fall within the [-300, 300] range and subsequently normalized to [0, 1], enhancing the contrast for soft tissues.
A CT atlas and associated anatomical label maps were constructed in~\citet{chen2023constructing} using an atlas construction method described in~\citep{dalca2019learning}.
The registration task associated with this dataset is atlas-to-subject registration.

\paragraph{4DCT} \textcolor{black}{The 4DCT dataset~\citep{castillo2009framework, castillo2009four} provides 4D CT breathing sequences with annotated landmarks. It consists of thoracic 4DCT scans from 10 subjects, acquired during radiotherapy planning for thoracic malignancies. Each scan includes 300 annotated landmarks. Due to variations in image size and resolution, we resampled and cropped/zero-padded the images to ensure that the lung region of each subject fits within a standardized spatial dimension of $224\times224\times224$. Specifically, Case 1 was resampled to a uniform voxel size of $1.14\times1.14\times1.14$ mm, Cases 2–5 to $1.4\times1.4\times1.4$ mm, and Cases 6–10 to $1.6\times1.6\times1.6$ mm. The landmarks were transformed accordingly by first defining a sphere of radius 2.5 voxels around each landmark in the original image. After resampling and cropping/ padding, the centroid of the transformed sphere was used as the updated landmark position. The registration task for this dataset involves intra-patient registration. In this study, we focus on registering images from the extreme breathing phases, specifically from "\textit{T00}" to "\textit{T50}". Given the limited number of image pairs (10), the dataset is less suitable for training learning-based methods. Instead, we employ instance-specific optimization of registration DNNs~\citep{balakrishnan2019voxelmorph,siebert2024convexadam,tian2024unigradicon} to adapt them for this registration task. Specifically, we pretrained the DNNs using randomly generated shapes of size $224\times224\times224$, as demonstrated in~\citet{chen2025pretraining}, and then applied instance-specific optimization on each image pair. Cases 4–7 were designated as the validation set for tuning regularization hyperparameters. Further details on the training strategy are provided in Sect.~\ref{sec:imp_details}.}  

\begin{table}[t]
\centering
\caption{Search range of hyperparameters used in the Bayesian hyperparameter optimization.}
\fontsize{7.5}{9}\selectfont
\begin{tabular}{c cccc}
 \toprule
 Dataset & $\lambda_{mean}$ & $\sigma'$ & $\lambda_{max}$ & $\alpha'$ \\
 \cmidrule(lr){1-1}\cmidrule(lr){2-2}\cmidrule(lr){3-3}\cmidrule(lr){4-4}\cmidrule(lr){5-5}
 \rowcolor{Gray}
Whole-body CT~(\textit{autoPET}) & {[}0.8, 3{]}  & {[}0, 2{]} & {[}0.8, 3{]} & {[}0, 0.1{]}\\

Brain MRI (\textit{IXI}) & {[}0.5, 5{]} & {[}0, 4{]} & {[}0.5, 5{]} & {[}0, 0.2{]}\\
\rowcolor{Gray}
Cardiac MRI (\textit{ACDC M\&Ms}) & {[}0.8, 3{]} & {[}0, 2{]} & {[}0.8, 3{]} & {[}0, 0.6{]}\\
\textcolor{black}{Lung CT (\textit{4DCT})} & \textcolor{black}{{[}0.01, 3{]}} & \textcolor{black}{{[}0, 0.5{]}} & \textcolor{black}{{[}0.01, 3{]}} & \textcolor{black}{{[}0, 0.3{]}}\\
 \hline

\end{tabular}
\label{tab:BO_details}
\end{table}

\subsection{Baseline Methods}
To comprehensively evaluate the benefits of the proposed spatially varying regularization, we compared it with several well-established baseline methods. These included two traditional optimization-based approaches: \texttt{SyN} from the ANTs package~\citep{avants2008symmetric}, widely considered the state-of-the-art for neuroimaging, and \texttt{deedsBCV}~\citep{heinrich2013mrf}, a leading method for abdominal registration. 

We also assessed our method against \texttt{VoxelMorph}~\citep{balakrishnan2019voxelmorph}, a classical unsupervised ConvNet-based method, enhanced with augmented parameter scaling as suggested in \citep{jian2024mamba} for improved performance. Additionally, we conducted comparisons with \texttt{TransMorph}~\citep{chen2022transmorph}, the backbone network used in the proposed method, to assess the impact of introducing spatially varying regularization. For both \texttt{VoxelMorph} and \texttt{TransMorph}, we employed commonly used hyperparameter settings during training~\citep{balakrishnan2019voxelmorph, meng2024corr}, setting the weights for the image similarity measure~(i.e., NCC) and the deformation regularization (i.e., diffusion regularizer) equally to 1. 

Furthermore, we evaluated the proposed approach against \texttt{HyperMorph}~\citep{hoopes2022learning}, a hyperparameter optimization framework that selects regularization parameters based on a validation dataset. For a fair comparison, we extended \texttt{HyperMorph} to \texttt{TransMorph} by integrating its hyperparameter learning framework while maintaining the same backbone network as the proposed method, denoted as \texttt{Hyper-TransMorph}. This ensures that performance differences can be attributed specifically to spatially varying regularization. 

\textcolor{black}{A special consideration applies to lung CT registration on the 4DCT dataset. Due to the dataset's small size, we adapted learning-based registration DNNs by pretraining on synthetic images of random shapes~\citep{chen2025pretraining}, followed by instance-specific optimization. As a result, the learning-based methods effectively function as optimization-based approaches. However, this adaptation prevents the use of hypernetwork-based methods, such as \texttt{HyperMorph} and \texttt{HyperTransMorph}, since their training and validation sets (i.e., synthetic shapes and lung CT images) would have mismatched distributions. Consequently, these methods were excluded from the comparative evaluation for lung CT registration.}

For all deep learning-based models, we consistently used NCC as the image similarity measure and the diffusion regularizer for deformation regularization. The detailed implementation of each method is provided below.

\begin{itemize}
    \item \texttt{SyN}~\citep{avants2008symmetric}: A diffeomorphic image registration method from the ANTs package. We employed the "\texttt{antsRegistrationSyN.sh}" pipeline optimized for atlas-based registration with the "\texttt{SyNOnly}" option and cross-correlation as the similarity measure.
    \item \texttt{deedsBCV}~\citep{heinrich2013mrf}: An optimization-based method using discrete optimization strategies with MIND-SSC~\citep{heinrich2012mind} as the similarity measure. For the whole-body CT registration task, the grid spacing, search radius, and quantization step were set to $[8, 7, 6, 5, 4]$, $[8, 7, 6, 5, 4]$, and $[5, 4, 3, 2, 1]$, respectively, based on default settings optimized for abdominal registration. For brain MRI registration, we followed recommendations in~\citep{hoffmann2021synthmorph} for neuroimaging, setting the grid spacing, search radius, and quantization step to $[6, 5, 4, 3, 2]$, $[6, 5, 4, 3, 2]$, and $[5, 4, 3, 2, 1]$, respectively. For cardiac MRI registration, we used the same hyperparameters as brain MRI registration, considering the smaller anatomical variability in intra-patient images. \textcolor{black}{For lung CT registration, we used the same hyperparameters as whole-body CT registration, as the same parameters demonstrated robust performance on lung CT registration in~\citet{heinrich2013mrf}.}
    \item \textcolor{black}{\texttt{ConvexAdam}~\citep{siebert2024convexadam}: A recently proposed optimization-based method that combines coupled convex optimization for initial registration with Adam optimization for instance-specific refinement of the deformation fields. We used its MIND-SSC~\citep{heinrich2012mind} variant in this study. For whole-body CT registration, we applied the recommended "\textit{AbdomenCTCT}" settings. For brain MRI registration, we used the "\textit{OASIS}" settings. For cardiac MRI registration, we used the default settings. For lung CT registration, we adopted the recommended "\textit{NLST}" settings. }
    \item \texttt{VoxelMorph}~\citep{balakrishnan2019voxelmorph}: An unsupervised learning-based model with a U-Net backbone~\citep{ronneberger2015u}. We adopted an enhanced version with an expanded model size, as proposed in~\citep{jian2024mamba}, setting the encoder and decoder channels to $[16, 32, 64, 96, 128]$ and $[128, 96, 64, 32]$, respectively.
    \item \texttt{HyperMorph}~\citep{hoopes2021hypermorph}: A framework for learning regularization hyperparameters using an auxiliary network. The auxiliary network comprises multiple small sub-networks, each consisting of five fully connected layers with [32, 64, 64, 128, 128] units, activated by ReLU, followed by a linearly activated fully connected layer to predict the parameters for a specific trainable layer in the registration network. The number of these sub-networks matches the number of trainable layers in the registration network. In this study, \texttt{HyperMorph} was applied to \texttt{VoxelMorph}, as described in the original work. The regularization hyperparameter $\lambda$ was provided as input to the auxiliary network, while the weight for the image similarity term in the loss function was set to $1-\lambda$ during training, as was done similarly in~\citet{hoopes2021hypermorph}.
    \item \texttt{TransMorph}~\citep{chen2022transmorph}: A Transformer-based registration model. We employed an improved version~\citep{chen2022unsupervised}, which demonstrated superior performance compared to the original model and served as our baseline.
    \item \texttt{Hyper-TransMorph}: An extension of \texttt{TransMorph} integrating the \texttt{HyperMorph} framework to enable hyperparameter learning while retaining the Transformer-based backbone.
\end{itemize}

\subsection{Implementation Details}
\label{sec:imp_details}

\textcolor{black}{For all registration tasks except lung CT, we trained each learning-based model} for 500 epochs using the Adam optimizer~\citep{kingma2014adam} with a batch size of 1 and a learning rate of 0.0001.
The number of integration steps for SS (i.e., $N$ in Eqn.~\ref{eqn:SS}) to enforce diffeomorphic registration was also set to 7. 
The implementation was carried out using the PyTorch framework, and training was performed on NVIDIA GPUs, including the RTX 3090, RTX Titan, RTX A6000, and H100.

\textcolor{black}{For lung CT registration on the 4DCT dataset, we first pretrained the registration DNNs (i.e., \texttt{VoxelMorph}, \texttt{TransMorph}, and \texttt{TM-SPR}) using pairs of synthetic images generated via Perlin noise~\citep{perlin2002improving}. This pretraining strategy has been shown to be effective for training registration networks~\citep{hoffmann2021synthmorph}, and our recent work~\citep{chen2025pretraining} further demonstrated its benefit as an initialization method for lung CT image registration. Specifically, multiple noise channels of mixed scales were produced, then combined with an "\textit{argmax}" operation to generate images containing random shapes. We added random levels of Gaussian noise to these images and simultaneously generated random deformation fields from Perlin noise to create corresponding image pairs. Pretraining proceeded for 200 epochs, each consisting of 3,000 randomly generated image pairs, using a loss function with equal weights on NCC and diffusion regularization. We employed the Adam optimizer with a batch size of 1. Afterward, we performed instance-specific optimization on each 4DCT image pair by iteratively updating the pretrained DNN for 400 iterations.}

For hyperparameter tuning, we conducted 50 trials for each experiment using Optuna. The specific search ranges for the hyperparameters, $\lambda_{mean}$ and $\sigma'$ for the Gaussian prior, or $\lambda_{max}$ and $\alpha'$ for the beta prior, are detailed in Table~\ref{tab:BO_details}. Furthermore, we applied a median pruner with a startup trial of 5, a warm-up step of 30, and an interval step of 10 to optimize the search process throughout the study.

\subsection{Evaluation Metrics} To evaluate the registration performance, we measured the overlap of anatomical label maps between the fixed image and the deformed moving image using the Dice coefficient. Given the focus on unsupervised image registration in this study, label maps were not used during the training phase. Thus, the Dice score serves as an indirect but adequate metric to assess registration accuracy. \textcolor{black}{We applied this metric to the whole-body CT, brain MRI, and cardiac MRI registration tasks. For lung CT registration, we measured target registration error (TRE), defined as the distance between displaced landmarks from the moving image and the corresponding landmarks in the fixed image.} To assess the smoothness of the deformation fields, we used the percentage of non-positive Jacobian determinants (\%$\vert J\vert\leq0$) and the non-diffeomorphic volume~(\%NDV), as described in \citep{liu2022finite}. These metrics provide more accurate evaluations of the invertibility of the deformation under the finite-difference approximation.
\begin{table}[t]
\centering
\caption{Quantitative results for atlas-to-subject whole-body CT registration on the autoPET dataset. Top-performing results are highlighted in \textbf{bold}, and second-best results are indicated in \textit{italic}. Statistically significant improvements, measured using the Wilcoxon signed-rank test, of the TransMorph variant over the backbone \texttt{TransMorph} are denoted by ** (**: p-value < 0.001; *: p-value < 0.01).}
\fontsize{9}{10.5}\selectfont
    \begin{tabular}{ c  c c c }
 \toprule
 \multicolumn{4}{c}{\textit{Whole-body CT Registration (autoPET)}}\\
 \cmidrule(lr){1-4}
 Method & Dice$\uparrow$ & \%$\vert J\vert\leq0\downarrow$ & \%NDV$\downarrow$ \\
 \cmidrule(lr){1-1}\cmidrule(lr){2-2}\cmidrule(lr){3-3}\cmidrule(lr){4-4}
 \rowcolor{Gray}
 Initial & 0.403$\pm$0.252 & - & -\\
 
 \texttt{SyN} & 0.633$\pm$0.263 &0.00\%&0.00\%\\
 \rowcolor{Gray}
 \texttt{deedsBCV} & 0.631$\pm$0.254 &0.00\%&0.00\%\\

 \textcolor{black}{\texttt{ConvexAdam}}&\textcolor{black}{0.625$\pm$0.260} &\textcolor{black}{0.00\%}&\textcolor{black}{0.00\%}\\
\rowcolor{Gray}
 \texttt{VoxelMorph} & 0.622$\pm$0.273 &0.03\% &0.00\%\\

 \texttt{HyperMorph} & 0.638$\pm$0.277 & 0.07\% & 0.38\%\\
  \rowcolor{Gray}
 \texttt{TransMorph} & 0.661$\pm$0.267 & 0.00\% & 0.00\%\\
 \textcolor{black}{\texttt{Hyper-TransMorph}} & \textcolor{black}{\textit{0.674$\pm$0.267}**} & \textcolor{black}{0.00\%} & \textcolor{black}{0.00\%}\\
 \cmidrule(lr){1-1}\cmidrule(lr){2-2}\cmidrule(lr){3-3}\cmidrule(lr){4-4}
 \rowcolor{Gray}
 \texttt{TM-SPR}$_{\text{\textit{Gaussian}}}$ & 0.670$\pm$0.272** & 0.00\% & 0.00\%\\

 \texttt{TM-SPR}$_{\text{\textit{Beta}}}$ & \textbf{0.677$\pm$0.268}** & 0.00\% & 0.00\%\\
 \hline
\end{tabular}
\label{tab:autoPET}
\end{table}

\begin{figure*}[t]
\begin{center}
\includegraphics[width=0.98\textwidth]{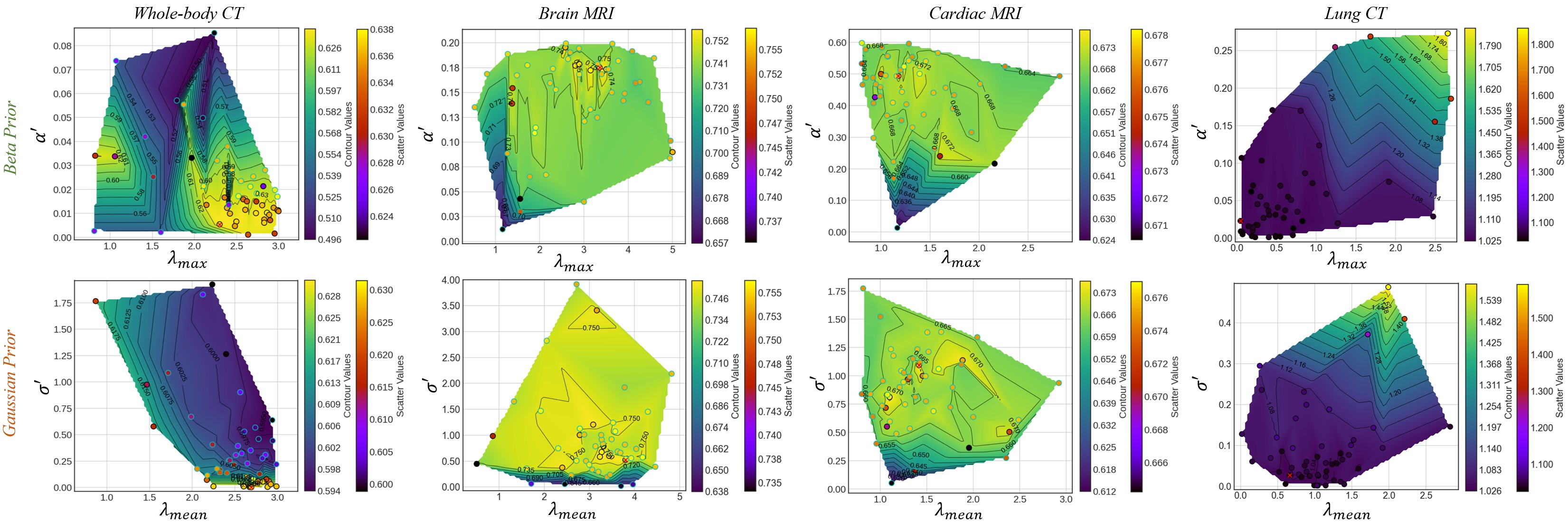}
\end{center}
   \caption{Hyperparameter surface maps obtained through Bayesian Optimization~(BO) for $\alpha'$ and $\lambda_{max}$ (beta prior) and $\sigma'$ and $\lambda_{mean}$ (Gaussian prior), evaluated on the validation set across \textcolor{black}{four} registration tasks: whole-body CT registration (\textit{autoPET}), brain MRI registration (\textit{IXI}), cardiac MRI registration (\textit{ACDC M\&Ms}), \textcolor{black}{and lung CT registration (\textit{4DCT})} from left to right. Scatter points with black edges represent trials that completed the full training cycle, while those with cyan edges indicate trials pruned before completion. \textcolor{black}{Note that for the first three tasks, the objective is to maximize DSC, whereas for the lung CT task, the optimization minimizes TRE.}}
\label{fig:hyper_contour}
\end{figure*}

\section{Results}
\label{sec:results}
\subsection{Whole-body Image Registration Results on the autoPET Dataset}
\paragraph{1) Qualitative Evaluation} The top row of Fig.~\ref{fig:qual_results} presents the qualitative results of the proposed spatially varying regularization with optimized hyperparameters derived under two priors: beta (middle panel) and Gaussian (right panel). The spatial weight volume, $\pmb{\lambda})$, learned under both priors consistently favors smaller regularization weights within the body region. The Gaussian prior represents an extreme case, applying near-zero regularization within the body, whereas the beta prior assigns stronger regularization weights to central regions of organs where intensity differences between the moving and fixed images are smaller, such as in the lungs and pelvis.
This behavior highlights the beta prior's potential to adaptively modulate regularization based on local image characteristics in whole-body image registration. 

The regions of low regularization observed with both priors underscore the importance of the proposed approach in enforcing diffeomorphic registration within a time-stationary framework. Without such a strategy, deformations within the body are likely to become irregular and unrealistic, which is highly undesirable for medical image registration~\citep{chen2024survey}. Additional qualitative comparisons between different registration methods are shown in Fig.~\ref{fig:autopet_qual} in the appendix.

\paragraph{2) Quantitative Evaluation} Table~\ref{tab:autoPET} presents quantitative comparisons between different registration methods. The results indicate that learning-based models outperform optimization-based methods for this task, with \texttt{SyN}, \texttt{deedsBCV}, \textcolor{black}{and the recently proposed \texttt{ConvexAdam}} achieving mean Dice scores of 0.633, 0.631, \textcolor{black}{and 0.625}, respectively. Among learning-based models, \texttt{TransMorph}-based approaches outperform ConvNet-based models, i.e., \texttt{VoxelMorph} and \texttt{HyperMorph}, which optimizes the regularization hyperparameter $\lambda$ (as defined in Eqn.~\ref{eqn:diff_derive}). Specifically, \texttt{VoxelMorph} achieves a mean Dice score of 0.621, and \texttt{HyperMorph} improves this to 0.638. In comparison, \texttt{TransMorph}, which uses a fixed $\lambda$ value of 1 without optimization, achieves a higher mean Dice score of 0.661. 

The three \texttt{TransMorph}-based models capable of selecting optimal hyperparameters (i.e., \texttt{Hyper-TransMorph}, \texttt{TM-SPR}$_{\text{\textit{Gaussian}}}$, and \texttt{TM-SPR}$_{\text{\textit{Beta}}}$) demonstrate statistically significant improvements over \texttt{TransMorph} with a fixed regularization parameter ($\lambda$ in Eqn.~\ref{eqn:diff_derive} set to 1) and no hyperparameter optimization.
Among these models, \texttt{TM-SPR}$_{\text{\textit{Beta}}}$ achieves the highest mean Dice score of 0.677, followed by \texttt{Hyper-TransMorph} at 0.674, and \texttt{TM-SPR}$_{\text{\textit{Gaussian}}}$ at 0.670. Although the performance of \texttt{TM-SPR}$_{\text{\textit{Beta}}}$ and \texttt{Hyper-TransMorph} is very similar, a statistically significant difference is observed between the two under the Wilcoxon signed-rank test ($p=0.009$).

These results highlight the prominence of the proposed spatially varying regularization for whole-body CT registration. The comparable mean Dice scores of \texttt{Hyper-TransMorph} and \texttt{TM-SPR} models can be attributed to the nature of the task, which favors minimal regularization (i.e., $\lambda$ approaches 0). For instance, the optimal hyperparameter for \texttt{Hyper-TransMorph} is 0.06, while for the spatially varying regularization models under the beta and Gaussian priors, the parameters enforcing larger values in the spatial weight volume $\pmb{\lambda})$ (i.e., $\alpha'$ and $\sigma'$) are also close to zero, at 0.027 and 0.005, respectively, as shown in Table~\ref{tab:optimal_hyper}. Consequently, the spatial weights within the body regions remain quite small and near zero, as illustrated in Fig.~\ref{fig:hyper_contour}. Theoretically, when deformation regularization is minimal or absent, \texttt{Hyper-TransMorph}, \texttt{TM-SPR}$_{\text{\textit{Beta}}}$, and \texttt{TM-SPR}$_{\text{\textit{Gaussian}}}$ effectively converge to the baseline \texttt{TransMorph} model with no deformation regularization (i,e, $\lambda=0$). This convergence explains the similar performance observed among these models. 

A detailed comparison of Dice scores across various methods for different anatomical structures in whole-body CT is illustrated using box plots in Fig.~\ref{fig:autopet_box} in the appendix.

In terms of regularity of deformation, optimization-based methods produced highly regularized deformations. Among learning-based methods, \texttt{VoxelMorph} and \texttt{HyperMorph} exhibited greater deformation irregularities due to the absence of diffeomorphism constraints. In contrast, \texttt{TransMorph}-based methods produced highly regularized deformations, benefiting from the imposed diffeomorphic registration achieved through the time-stationary velocity framework.

\begin{table}[t]
\centering
\caption{Quantitative results for atlas-to-subject brain MRI registration on the IXI dataset. Top-performing results are highlighted in \textbf{bold}, and second-best results are indicated in \textit{italic}. Statistically significant improvements, measured using the Wilcoxon signed-rank test, of the TransMorph variant over the backbone \texttt{TransMorph} are denoted by ** (**: p-value < 0.001; *: p-value < 0.01).}
\fontsize{9}{10.5}\selectfont
    \begin{tabular}{ c  c c c }
 \toprule
 \multicolumn{4}{c}{\textit{Brain MRI Registration (IXI)}}\\
 \cmidrule(lr){1-4}
 Method & Dice$\uparrow$ & \%$\vert J\vert\leq0\downarrow$ & \%NDV$\downarrow$ \\
 \cmidrule(lr){1-1}\cmidrule(lr){2-2}\cmidrule(lr){3-3}\cmidrule(lr){4-4}
 \rowcolor{Gray}
 Initial & 0.386$\pm$0.195 & - & -\\
 
 \texttt{SyN} & 0.758$\pm$0.126 &0.00\%&0.00\%\\
 \rowcolor{Gray}
 \texttt{deedsBCV} & 0.740$\pm$0.127 & 0.05\%& 0.02\%\\
 
 \textcolor{black}{\texttt{ConvexAdam}} & \textcolor{black}{0.749$\pm$0.126} & \textcolor{black}{0.06\%}& \textcolor{black}{0.01\%}\\
 \rowcolor{Gray}
 \texttt{VoxelMorph} & 0.744$\pm$0.131 &4.33\%&1.74\%\\

 \texttt{HyperMorph} & 0.753$\pm$0.131 & 0.38\% & 0.08\%\\
 \rowcolor{Gray}
 \texttt{TransMorph} & 0.757$\pm$0.126 & 0.00\% & 0.00\%\\
 \textcolor{black}{\texttt{Hyper-TransMorph}} & \textcolor{black}{0.762$\pm$0.127**} & \textcolor{black}{0.00\%} & \textcolor{black}{0.00\%}\\
 \cmidrule(lr){1-1}\cmidrule(lr){2-2}\cmidrule(lr){3-3}\cmidrule(lr){4-4}
 \rowcolor{Gray}
 \texttt{TM-SPR}$_{\text{\textit{Gaussian}}}$ & \textbf{0.767$\pm$0.127}** & 0.00\% & 0.00\%\\

 \texttt{TM-SPR}$_{\text{\textit{Beta}}}$ & \textit{0.767$\pm$0.129}** & 0.00\% & 0.00\%\\
 \hline
\end{tabular}
\label{tab:IXI}
\end{table}

\begin{table}[t]
\centering
\caption{Quantitative results for intra-subject cardiac MRI registration on the ACDC and the M\&Ms datasets. Top-performing results are highlighted in \textbf{bold}, and second-best results are indicated in \textit{italic}. Statistically significant improvements, measured using the Wilcoxon signed-rank test, of the TransMorph variant over the backbone \texttt{TransMorph} are denoted by ** (**: p-value < 0.001; *: p-value < 0.01).}
\fontsize{9}{10.5}\selectfont
    \begin{tabular}{ c c c c }
 \toprule
 \multicolumn{4}{c}{\textit{Cardiac MRI Registration (ACDC M\&Ms)}}\\
 \cmidrule(lr){1-4}
 Method & Dice$\uparrow$ & \%$\vert J\vert\leq0\downarrow$ & \%NDV$\downarrow$ \\
 \cmidrule(lr){1-1}\cmidrule(lr){2-2}\cmidrule(lr){3-3}\cmidrule(lr){4-4}
 \rowcolor{Gray}
 Initial & 0.537$\pm$0.166 & - & -\\
 
 \texttt{SyN} &0.678$\pm$0.138 &0.00\%&0.00\%\\
 %\texttt{FireANTs} &&&\\
 \rowcolor{Gray}
 \texttt{deedsBCV}&\textbf{0.740$\pm$0.121} &0.00\%&0.00\%\\

\textcolor{black}{\texttt{ConvexAdam}}&\textcolor{black}{0.689$\pm$0.142} &\textcolor{black}{0.00\%}&\textcolor{black}{0.00\%}\\
 \rowcolor{Gray}
 \texttt{VoxelMorph} & 0.689$\pm$0.128 & 0.44\%& 0.17\%\\
 
 \texttt{HyperMorph} & 0.647$\pm$0.141 & 0.08\% & 0.20\% \\
 \rowcolor{Gray}
 \texttt{TransMorph} & 0.711$\pm$0.126 & 0.00\% & 0.00\%\\
\textcolor{black}{\texttt{Hyper-TransMorph}} & \textcolor{black}{0.698$\pm$0.127} & \textcolor{black}{0.00\%} & \textcolor{black}{0.00\%}\\
 \cmidrule(lr){1-1}\cmidrule(lr){2-2}\cmidrule(lr){3-3}\cmidrule(lr){4-4}
 \rowcolor{Gray}
 \texttt{TM-SPR}$_{\text{\textit{Gaussian}}}$ & 0.733$\pm$0.120** & 0.00\% & 0.00\%\\
 
 \texttt{TM-SPR}$_{\text{\textit{Beta}}}$ & \textit{0.735$\pm$0.119}** & 0.00\% & 0.00\%\\
 \hline
\end{tabular}
\label{tab:ACDC}
\end{table}

\begin{table*}[t]
\centering
\caption{\textcolor{black}{Quantitative results of the target registration error (TRE) for 10 cases from the intra-subject Lung CT registration on the 4DCT dataset. The average TRE (Avg. TRE) is obtained by averaging the mean TRE across all 10 subjects. Cases 4–7 were used as the validation set for \texttt{TransMorph}, \texttt{TM-SPR}$_{\text{\textit{Gaussian}}}$, and \texttt{TM-SPR}$_{\text{\textit{Beta}}}$ to optimize hyperparameters. Top-performing results are highlighted in \textbf{bold}, and second-best results are indicated in \textit{italic}.}}
\fontsize{7.5}{9}\selectfont
    \begin{tabular}{ c  c c c c c c c c c}
 \toprule
 \multicolumn{10}{c}{\textit{Lung CT Registration (4DCT)}}\\
 \cmidrule(lr){1-10}
 Case No. & Resolution (mm$^3$) & Initial & \texttt{SyN} & \texttt{deedsBCV} & \texttt{ConvexAdam} & \texttt{VoxelMorph}& \texttt{TransMorph} & \texttt{TM-SPR}$_{\text{\textit{Gaussian}}}$ &\texttt{TM-SPR}$_{\text{\textit{Beta}}}$ \\
 \cmidrule(lr){1-1}\cmidrule(lr){2-2}\cmidrule(lr){3-3}\cmidrule(lr){4-4}\cmidrule(lr){5-5}\cmidrule(lr){6-6}\cmidrule(lr){7-7}\cmidrule(lr){8-8}\cmidrule(lr){9-9}\cmidrule(lr){10-10}
 \rowcolor{Gray}
 1 & 1.14$\times$1.14$\times$1.14 & 3.938$\pm$2.809 & 1.193$\pm$0.612 &1.079$\pm$0.646 &1.073$\pm$0.576 & 1.054$\pm$0.555 & 1.065$\pm$0.596 & \textbf{1.039$\pm$0.543}&\textit{1.045$\pm$0.557} \\
 
 2 & 1.4$\times$1.4$\times$1.4 & 4.366$\pm$3.932 & 1.586$\pm$0.646 &1.024$\pm$0.679 & 1.027$\pm$0.658& 1.027$\pm$0.588  &1.037$\pm$0.523 &\textbf{0.987$\pm$0.523}& \textit{0.995$\pm$0.530} \\
\rowcolor{Gray}
 3 & 1.4$\times$1.4$\times$1.4 &6.977$\pm$4.100 & 1.857$\pm$1.393 &1.349$\pm$0.825 & 1.257$\pm$0.702 & 1.279$\pm$0.659 & 1.390$\pm$0.713 &\textit{1.224$\pm$0.666}& \textbf{1.218$\pm$0.676}\\

 4 & 1.4$\times$1.4$\times$1.4 &9.906$\pm$4.882 & 1.973$\pm$1.549 & 1.543$\pm$1.003 &\textit{1.483$\pm$1.004} & 1.591$\pm$1.013  & 1.509$\pm$0.986 &\textbf{1.469$\pm$0.974}& 1.491$\pm$0.962 \\
 \rowcolor{Gray}
 5 & 1.4$\times$1.4$\times$1.4 &7.525$\pm$5.540 & 2.627$\pm$2.476 & 1.621$\pm$1.479 & 1.503$\pm$1.412 & 1.544$\pm$1.300 & 1.531$\pm$1.363 &\textbf{1.463$\pm$1.331}& \textit{1.465$\pm$1.329} \\

 6 & 1.6$\times$1.6$\times$1.6 &10.957$\pm$7.017 & 3.688$\pm$3.349 & 1.898$\pm$1.404 &1.686$\pm$1.332 & 1.542$\pm$0.867 & \textit{1.529$\pm$0.863} &1.548$\pm$0.892& \textbf{1.523$\pm$0.844} \\
 \rowcolor{Gray}
 7 & 1.6$\times$1.6$\times$1.6 &11.084$\pm$7.456 & 5.469$\pm$5.100 & 2.029$\pm$1.699 &2.204$\pm$2.496 & 1.599$\pm$0.912 & 1.477$\pm$0.797 & \textbf{1.444$\pm$0.799}& \textit{1.471$\pm$0.808} \\

 8 & 1.6$\times$1.6$\times$1.6 &14.983$\pm$8.958 & 9.411$\pm$8.473 & 2.018$\pm$2.308 &3.554$\pm$5.563 & 1.876$\pm$1.930 & 1.593$\pm$1.309 & \textbf{1.551$\pm$1.207}& \textit{1.583$\pm$1.218} \\
 \rowcolor{Gray}
 9 & 1.6$\times$1.6$\times$1.6 &7.993$\pm$4.017 & 4.477$\pm$2.926 & 1.710$\pm$1.296 &1.535$\pm$1.135 & 1.456$\pm$0.808 & 1.468$\pm$0.868 & \textbf{1.382$\pm$0.771} & \textit{1.417$\pm$0.824} \\

 10 & 1.6$\times$1.6$\times$1.6 &7.356$\pm$6.398 & 3.001$\pm$3.324 & 1.860$\pm$1.790 &1.626$\pm$1.869 & 1.659$\pm$1.280 & 1.482$\pm$1.059 & \textbf{1.457$\pm$0.918} &\textit{1.462$\pm$0.998}\\
 
 \cmidrule(lr){1-1}\cmidrule(lr){2-2}\cmidrule(lr){3-3}\cmidrule(lr){4-4}\cmidrule(lr){5-5}\cmidrule(lr){6-6}\cmidrule(lr){7-7}\cmidrule(lr){8-8}\cmidrule(lr){9-9}\cmidrule(lr){10-10}
 \rowcolor{Gray}
 &Avg. TRE$\downarrow$ & 8.509$\pm$3.154 & 3.528$\pm$2.342 & 1.613$\pm$0.346 & 1.695$\pm$0.697 & 1.462$\pm$0.255 &1.408$\pm$0.185&\textbf{1.356$\pm$0.193}& \textit{1.367$\pm$0.196} \\
 &\%$\vert J\vert\leq0\downarrow$& 0.00\% & 0.00\% & 0.00\% & 0.00\%&0.00\%  & 0.00\% &0.00\%& 0.00\%\\
 \rowcolor{Gray}
 &\%NDV$\downarrow$& 0.00\% & 0.00\% & 0.00\% & 0.00\% &0.00\%  & 0.00\%&0.00\%& 0.00\%\\
 \hline
\end{tabular}
\label{tab:4DCT}
\end{table*}

\subsection{Brain Image Registration Results on the IXI Dataset}
\paragraph{1) Qualitative Evaluation} The \textcolor{black}{second} row of Fig.~\ref{fig:qual_results} shows the qualitative results of the proposed spatially varying regularization applied to brain MRI registration, with results derived from the beta prior in the middle panel and the Gaussian prior in the right panel. For both priors, the learned spatial weight volume (i.e., $\pmb{\lambda}$) is observed to yield smaller values toward the cortical regions of the brain, where greater shape variability exists across images and patients, resulting in weaker deformation regularization. In contrast, the values in the $\pmb{\lambda}$ increase in the deeper subcortical regions, indicating a stronger deformation regularization in these areas. 

These patterns are more pronounced under the beta prior, which provides a $\pmb{\lambda})$ that is more structured and interpretable. In contrast, while the Gaussian prior demonstrates a similar trend, its $\pmb{\lambda}$ appears to be more complex and less coherent compared to those derived from the beta prior. Importantly, the regions with near-zero values in $\pmb{\lambda}$ are consistent across both priors, predominantly located near the cortical regions. 

Despite weaker regularization in certain areas, the deformation fields remain overall smooth, as shown in Fig.~\ref{fig:qual_results}. This smoothness can be attributed to two key factors. First, the overall regularization strength per voxel is relatively high, as indicated by the values $\lambda_{max} = 3.354$ and $\lambda_{mean} = 3.796$ in Table\ref{tab:optimal_hyper}. Although some regions exhibit smaller spatial weights, the combination of larger $\lambda_{max}$ and $\lambda_{mean}$ with the spatial weight volume, $\pmb{\lambda}$, collectively enforces strong regularization across voxels, resulting in an overall smooth deformation field. Second, the time-stationary framework effectively enforces diffeomorphic registration, which further contributes to the observed smoothness. 

Additional qualitative comparisons of different registration methods on the brain MRI registration task are shown in Fig.~\ref{fig:ixi_qual} in the appendix.

\begin{table*}[t]
\centering
\caption{Optimal regularization hyperparameters identified through grid search for \texttt{HyperMorph} and \texttt{Hyper-TransMorph}, and BO for \texttt{TM-SPR}$_{\text{\textit{Gaussian}}}$ and \texttt{TM-SPR}$_{\text{\textit{Beta}}}$. \textcolor{black}{Note that since we used instance-specific optimization for Lung CT registration task, where the hyperparameter learning frameworks are not applicable, we therefore did not include \texttt{HyperMorph} and \texttt{Hyper-TransMorph} for this experiment.}}
\fontsize{8}{9.5}\selectfont
    \begin{tabular}{ c c c c c c c c c }
 \toprule
 &\multicolumn{2}{c}{\textit{Whole-body CT Registration}}&\multicolumn{2}{c}{\textit{Brain MRI Registration}}&\multicolumn{2}{c}{\textit{Cardiac MRI Registration}}&\multicolumn{2}{c}{\textcolor{black}{\textit{Lung CT Registration}}}\\
 \cmidrule(lr){1-1}\cmidrule(lr){2-3}\cmidrule(lr){4-5}\cmidrule(lr){6-7}\cmidrule(lr){8-9}
 \rowcolor{Gray}
 \texttt{HyperMorph} & \multicolumn{2}{c}{$\lambda=0.25$} & \multicolumn{2}{c}{$\lambda=0.75$} &\multicolumn{2}{c}{$\lambda=0.65$} &\multicolumn{2}{c}{\textcolor{black}{-}} \\
 \texttt{Hyper-TransMorph} & \multicolumn{2}{c}{$\lambda=0.06$}& \multicolumn{2}{c}{$\lambda=0.71$}  &\multicolumn{2}{c}{$\lambda=0.58$} &\multicolumn{2}{c}{\textcolor{black}{-}}  \\
 \rowcolor{Gray}
 \texttt{TM-SPR}$_{\text{\textit{Gaussian}}}$ & \textcolor{black}{$\sigma'=0.002$} & \textcolor{black}{$\lambda_{mean}=2.691$}& $\sigma'=0.525$ & $\lambda_{mean}=3.796$ & $\sigma'=1.416$ & $\lambda_{mean}=1.094$ & \textcolor{black}{$\sigma'=0.024$} & \textcolor{black}{$\lambda_{mean}=1.148$} \\
 
 \texttt{TM-SPR}$_{\text{\textit{Beta}}}$ & $\alpha'=0.005$ & $\lambda_{max}=2.298$ & $\alpha'=0.175$ & $\lambda_{max}=3.354$& \textcolor{black}{$\alpha'=0.493$} & \textcolor{black}{$\lambda_{max}=1.177$} & \textcolor{black}{$\alpha'=0.062$} & \textcolor{black}{$\lambda_{max}=0.181$}\\
 \hline
\end{tabular}
\label{tab:optimal_hyper}
\end{table*}

\paragraph{2) Quantitative Evaluation} Table~\ref{tab:IXI} presents the quantitative results for brain MRI registration. Optimization-based methods demonstrate strong performance for this task, with \texttt{SyN} achieving a mean Dice score of 0.758, \texttt{deedsBCV} achieving 0.740, \textcolor{black}{and \texttt{ConvexAdam} achieving 0.749}. Notably, \texttt{SyN} significantly surpasses learning-based methods such as \texttt{VoxelMorph} and \texttt{HyperMorph} and performs comparably to \texttt{TransMorph}. It is worth mentioning that both \texttt{VoxelMorph} and \texttt{HyperMorph} were augmented with scaled-up model parameters, as suggested in recent research~\citep{jian2024mamba}, yet \texttt{SyN} maintains robust performance. Its robust performance is likely due to the use of the "\texttt{antsRegistrationSyN.sh}" pipeline, specifically optimized for atlas-to-subject brain MRI registration, rather than the default registration pipeline used in our previous study~\citep{chen2022transmorph}. 

Among learning-based methods, \texttt{TransMorph} with a fixed regularization parameter ($\lambda = 1$) outperforms both \texttt{VoxelMorph} and \texttt{HyperMorph}, with the latter optimized for regularization hyperparameters. Specifically, \texttt{TransMorph} achieves a mean Dice score of 0.757, compared to 0.744 for \texttt{VoxelMorph} and 0.753 for \texttt{HyperMorph}. When comparing hyperparameter learning methods, all approaches significantly outperform their backbone \texttt{TransMorph}. In particular, \texttt{TM-SPR}$_{\text{\textit{Gaussian}}}$ achieves the highest mean Dice score of 0.767, closely followed by \texttt{TM-SPR}$_{\text{\textit{Beta}}}$, which also achieves 0.767 but with slightly higher variance, and \texttt{Hyper-TransMorph} at 0.762. The Wilcoxon signed-rank test reveals a significant difference between \texttt{TM-SPR}$_{\text{\textit{Gaussian}}}$ and \texttt{Hyper-TransMorph} with $p\ll0.001$.

Interestingly, the $\lambda$ value for \texttt{Hyper-TransMorph} (0.58) is comparable to the $\lambda_{max}$ value for \texttt{TM-SPR}$_{\text{\textit{Beta}}}$ (0.493), as shown in Table~\ref{tab:optimal_hyper}. This indicates that the primary distinction between the two methods lies in their approach to regularization: \texttt{Hyper-TransMorph} applies a spatially-uniform regularization strength, while \texttt{TM-SPR}$_{\text{\textit{Beta}}}$ employs spatially varying regularization. The superior mean Dice score achieved by \texttt{TM-SPR}$_{\text{\textit{Beta}}}$ suggests that this improvement is attributable to the spatially varying, subject-adaptive deformation regularization, which facilitates a more precise alignment of brain structures.

A detailed comparison of Dice scores across various methods for different brain structures is illustrated using box plots in Fig.~\ref{fig:ixi_box} in the appendix.

In terms of deformation regularity, \texttt{deedsBCV} produced minor folded voxels and non-diffeomorphic volumes. As observed in the previous task, \texttt{VoxelMorph} and \texttt{HyperMorph} exhibit greater deformation irregularity. In contrast, \texttt{TransMorph}-based methods consistently produce highly regularized deformations.

\subsection{Cardiac Image Registration Results on the ACDC and M\&Ms Datasets}
\paragraph{1) Qualitative Evaluation} The \textcolor{black}{third} row of Fig.~\ref{fig:qual_results} illustrates the qualitative results of cardiac MRI registration, with the middle panel showing results derived using the beta prior and the right panel showing results using the Gaussian prior. Unlike other tasks, the learned spatially varying regularization exhibits an almost piecewise constant pattern throughout the image, resulting in spatial weights that are large and nearly uniform across voxels. This observation is more pronounced with the beta prior, where the spatial weights appear more consistent. 

Despite producing similar spatial weight volumes, it is important to note that the regularization imposed at each voxel differs between the two priors. Specifically, the Gaussian prior enforces relatively stronger regularization, as indicated by $\lambda_{mean} = 1.094$ and $\lambda_{max} = 0.493$, shown in Table~\ref{tab:optimal_hyper}. Additional qualitative comparisons of various registration methods for the cardiac MRI registration task are presented in Fig.~\ref{fig:acdcmm_qual} in the appendix.

\paragraph{2) Quantitative Evaluation} Table~\ref{tab:ACDC} presents the quantitative comparisons for this task. Among optimization-based methods, while \texttt{SyN} achieves a lower score of 0.678, \texttt{deedsBCV} achieves the highest mean Dice score of 0.740, outperforming all other methods, including learning-based approaches. This result indicates the superiority of \texttt{deedsBCV} for unsupervised image registration on cardiac MRI.

For learning-based baseline methods, consistent with other tasks, \texttt{TransMorph}, with a fixed regularization hyperparameter set to 1, outperforms both \texttt{VoxelMorph} and \texttt{HyperMorph}. Specifically, \texttt{TransMorph} achieves a mean Dice score of 0.711, compared to 0.689 for \texttt{VoxelMorph} and 0.647 for \texttt{HyperMorph}. 

\begin{figure*}[t]
\begin{center}
\includegraphics[width=0.95\textwidth]{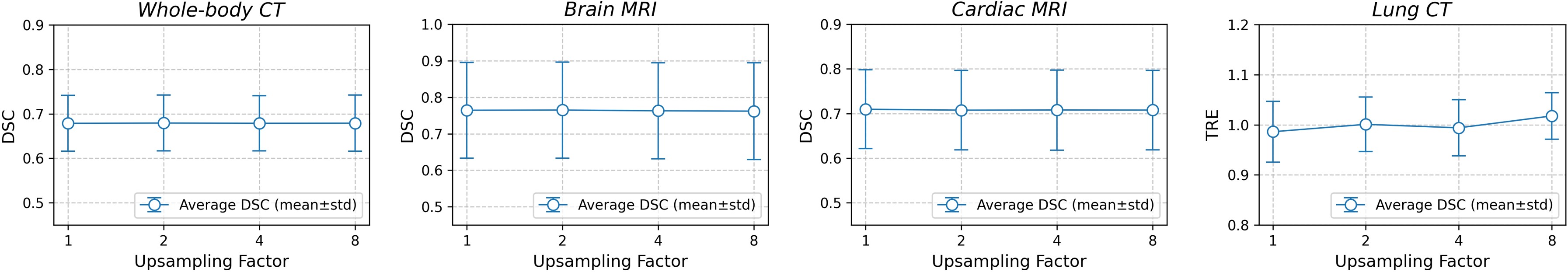}
\end{center}
   \caption{\textcolor{black}{Impact of upsampling factors on registration performance. Validation performance is shown for each registration task using different upsampling factors applied during the estimation of the spatial weight volume.}}
\label{fig:upsamp}
\end{figure*}

Notably, among methods that optimize regularization hyperparameters, \texttt{Hyper-TransMorph} underperforms its backbone, \texttt{TransMorph}, with a significantly lower mean Dice score of 0.698. Similarly, \texttt{HyperMorph} achieves lower performance compared to its backbone, \texttt{VoxelMorph}. In contrast, both \texttt{TM-SPR}$_{\text{\textit{Gaussian}}}$ and \texttt{TM-SPR}$_{\text{\textit{Beta}}}$ achieve significantly higher mean Dice scores of 0.733 and 0.735, respectively. These results indicate that the proposed method offers a more robust augmentation of the backbone model compared to spatially invariant regularization, resulting in significant performance improvements for this task.

A detailed comparison of Dice scores across various methods for the three cardiac structures is illustrated using box plots in Fig.~\ref{fig:acdcmm_box} in the appendix.

As observed in other tasks, all \texttt{TransMorph}-based models enforce diffeomorphism, resulting in highly regularized deformations with no folded voxels or non-diffeomorphic volumes. This consistency highlights the robustness of these methods in producing anatomically plausible deformations.

\subsection{Lung Image Registration Results on the 4DCT Dataset}
\paragraph{1) Qualitative Evaluation}
\textcolor{black}{The fourth row of Fig.\ref{fig:qual_results} presents qualitative results for lung CT registration, with landmarks overlaid on each image. The middle and right panels show the registration outcomes using the beta and Gaussian priors, respectively. Due to respiratory motion, lung deformation exhibits sliding effects, where the lung structures shrink while adjacent tissues, such as rib bones, remain largely stationary. For both priors, the estimated spatial weight maps display low or near-zero values at lung boundaries, thus permitting sharp transitions in deformation without penalization, effectively accommodating sliding motion. Additional qualitative results are provided in Fig.\ref{fig:lungct_qual} in the appendix.}

\paragraph{2) Quantitative Evaluation}
\textcolor{black}{Table~\ref{tab:4DCT} provides detailed quantitative comparisons of TRE across methods for each case in the 4DCT dataset. Cases 4–7 were used as the validation set for determining optimal hyperparameters in the learning-based approaches. Among the optimization-based methods, \texttt{deedsBCV} attained the best performance (mean TRE = 1.613 mm), outperforming both \texttt{SyN} (mean TRE = 3.528 mm) and \texttt{ConvexAdam} (mean TRE = 1.695 mm).}

\textcolor{black}{All learning-based methods achieved superior performance compared to optimization-based methods, largely due to the instance-specific optimization step and robust initialization provided by network pretraining~\citep{chen2025pretraining}. Hyperparameters for the regularization weights of the diffusion regularizer in \texttt{VoxelMorph} and \texttt{TransMorph} were optimized using BO over 50 trials, yielding optimal weights of 0.0136 and 0.0281, respectively. Quantitatively, \texttt{VoxelMorph} attained a mean TRE of 1.462 mm, while \texttt{TransMorph} achieved a better mean TRE of 1.408 mm.}

\textcolor{black}{Our proposed spatially varying regularization models (\texttt{TM-SPR}) delivered the best overall performance, with mean TREs below 1.4 mm, which is smaller than the average voxel size (1.47 mm). This implies an average registration error of less than one voxel. Specifically, \texttt{TM-SPR}$_{\text{\textit{Gaussian}}}$ achieved a mean TRE of 1.356 mm, and \texttt{TM-SPR}$_{\text{\textit{Beta}}}$ attained a mean TRE of 1.367 mm. Only in two cases (Case 3 and Case 6) did \texttt{TM-SPR}$_{\text{\textit{Beta}}}$ outperform \texttt{TM-SPR}$_{\text{\textit{Gaussian}}}$.}

\textcolor{black}{With instance-specific optimization, registration of a $224\times224\times224$ image pair requires 15.8 minutes over 400 epochs, averaged across ten runs.} We provide further discussions and additional qualitative examples related to discontinuity-preserving registration in lung CT data in Sect.~\ref{sec:discont_reg}.

\section{Discussion}
\label{sec:discussion}
\subsection{Hyperparameter Analysis}
In the following subsections, we outline the hyperparameter search process for the baseline models, \texttt{HyperMorph} and \texttt{Hyper-TransMorph}, focusing on the regularization hyperparameter $\lambda$. Additionally, we detail the hyperparameter optimization process for the proposed spatially varying regularization method, emphasizing the selection of $\alpha'$ and $\sigma'$ to control the spatial coherence of the spatial weight volume ($\pmb{\lambda})$), as well as $\lambda_{max}$ and $\lambda_{mean}$ to regulate the overall strength of the deformation regularization.
\subsubsection{Hyperparameter Selection for HyperMorph and Hyper-TransMorph}
The hyperparameter learning framework introduced by \texttt{HyperMorph} enables the efficient determination of optimal regularization hyperparameters through a single training session. This approach involves inputting different hyperparameter values into the auxiliary network, which predicts the corresponding model parameters for the registration network. The best model parameters are then selected on the basis of the registration accuracy achieved in the validation set.

For this study, we performed a dense grid search for the optimal regularization hyperparameter $\lambda$ within the range $[0, 1]$, using a step size of 0.01. The identified optimal $\lambda$ values for both \texttt{HyperMorph} and \texttt{Hyper-TransMorph} across the three registration tasks are summarized in Table~\ref{tab:optimal_hyper}. Notably, the optimal $\lambda$ values for brain and cardiac MRI registration tasks are similar for both models. However, for whole-body CT registration, \texttt{Hyper-TransMorph} favors a much smaller $\lambda$ (0.06), indicating minimal or no regularization, whereas \texttt{HyperMorph} selects a relatively higher $\lambda$ of 0.25, suggesting stronger regularization.

Figure~\ref{fig:hypermorph} in the appendix shows the dense grid search results. Although the overall trends in performance curves are consistent between the two models, with a sharp decline in performance observed beyond $\lambda = 0.8$, differences emerge in their peak performance and the "turning points" where performance begins to drop significantly. Additionally, \texttt{HyperMorph} exhibits less variability in performance compared to \texttt{Hyper-TransMorph}. These differences suggest that the underlying network architecture significantly influences the optimal hyperparameter values, highlighting that the best hyperparameter setting for one deep learning architecture may not generalize well to others. Although further exploration of this observation is beyond the scope of this study, it opens an intriguing avenue for future research.

\begin{figure*}[!tbp]
  \centering
  \begin{minipage}[b]{0.48\textwidth}
  \centering
    \includegraphics[width=0.8\textwidth]{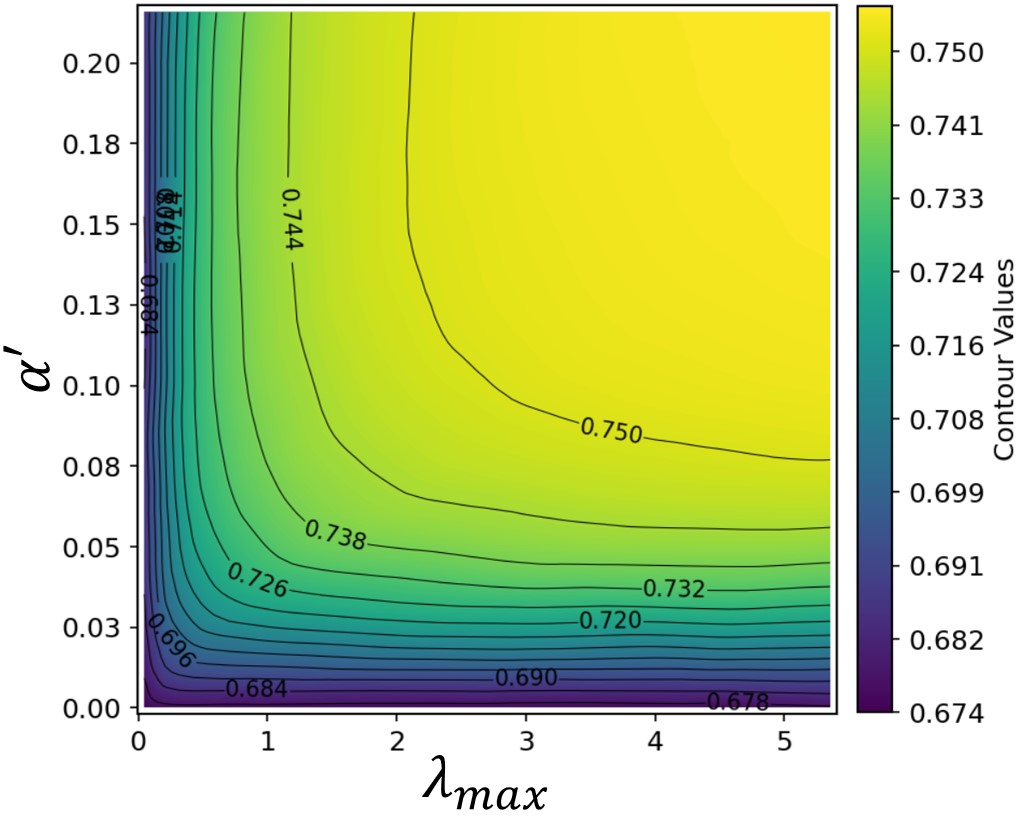}
    \caption{Hyperparameter surface map for \texttt{Hyper-TM-SPR}$_{\text{\textit{Beta}}}$, illustrating the results of a dense grid search over $\alpha'$ and $\lambda_{max}$ and their impact on registration performance, as measured by Dice scores, for the brain MRI registration task (IXI).}
    \label{fig:hyper_IXI}
  \end{minipage}
  \hfill
  \begin{minipage}[b]{0.48\textwidth}
  \centering
    \includegraphics[width=0.8\textwidth]{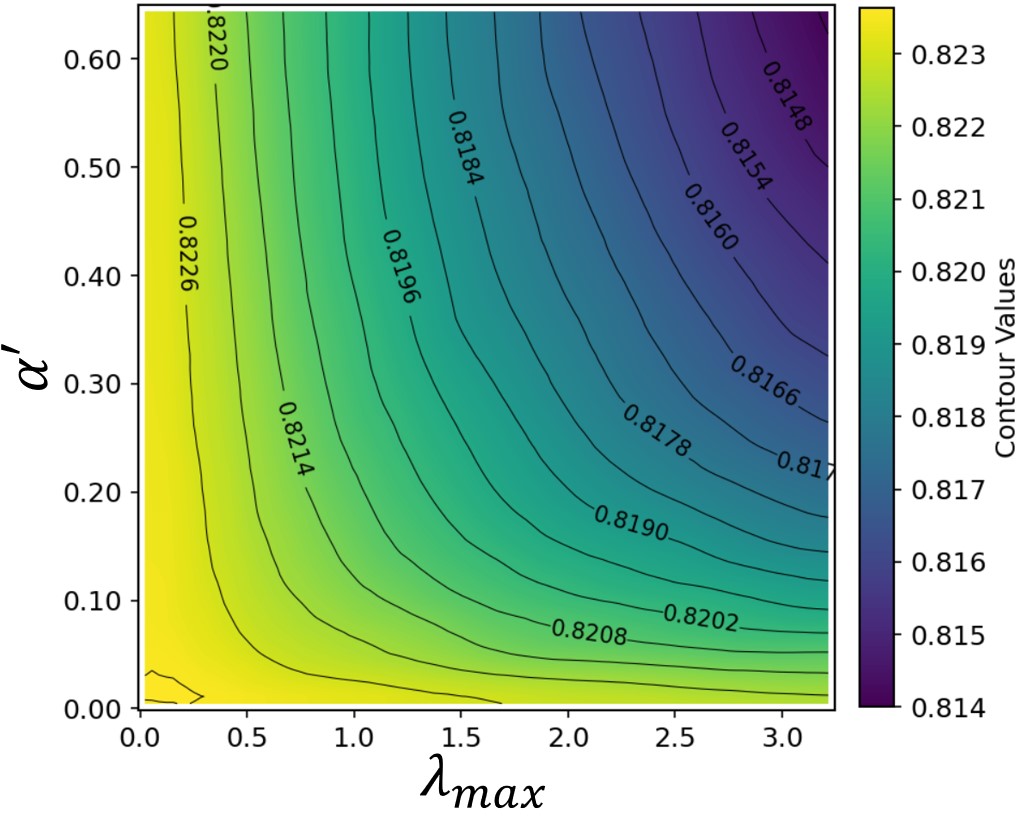}
    \caption{Hyperparameter surface map for \texttt{Hyper-TM-SPR}$_{\text{\textit{Beta}}}$, illustrating the results of a dense grid search over $\alpha'$ and $\lambda_{max}$ and their impact on registration performance, as measured by Dice scores, for the cardiac MRI registration task (ACDC M\&Ms).}
    \label{fig:hyper_ACDCMM}
  \end{minipage}
\end{figure*}

\subsubsection{Surface Maps for Different Prior Assumptions}
Since the proposed framework uses Bayesian hyperparameter optimization to tune the newly introduced hyperparameters by maximizing the posterior probability, it avoids the relatively more time-consuming and less efficient grid search approach. Figure~\ref{fig:hyper_contour} illustrates surface maps of Dice scores in the validation set for different hyperparameters under distinct prior distributions, specifically, $\lambda_{max}$ and $\alpha'$ for the beta prior and $\lambda_{mean}$ and $\sigma'$ for the Gaussian prior. Cyan-edged scatter points indicate hyperparameters pruned during training, while black-edged points denote those that completed the full training process, with the red cross marking the optimal performance.

One immediately noticeable trend is that different datasets favor significantly different hyperparameters. Not only were the optimal values distinctly varied, even under the same prior assumption (as shown in Sect.~\ref{sec:results}), but the contour patterns themselves also differ substantially. Under the beta prior (top row in Fig.~\ref{fig:hyper_contour}), whole-body CT image registration~(autoPET) favored smaller $\alpha'$ in the range $[0, 0.02]$ and larger $\lambda_{max}$ in the range $[2, 3]$. For brain registration, both $\alpha'$ and $\lambda_{max}$ were optimal at intermediate values, with $\alpha'$ in the range $[0.15, 0.2]$ and $\lambda_{max}$ within $[2, 3]$. This led to spatial weights that were low in regions with greater structural differences (indicating less regularization) but consistently close to 1 elsewhere, enforcing strong regularization in these areas, as illustrated in Fig.~\ref{fig:qual_results}. In contrast, cardiac image registration (ACDC M\&Ms) preferred larger $\alpha'$ values in the range $[0.4, 0.6]$ and smaller $\lambda_{max}$ values within $[0.5, 2]$, leading to spatial weights with a nearly constant high value of 1 across all voxels, as illustrated in Fig.~\ref{fig:qual_results}. \textcolor{black}{For lung CT registration (\textit{4DCT}), both hyperparameters again favored smaller values, with optimal $\alpha'$ in the range $[0, 0.05]$ and $\lambda_{max}$ within $[0, 0.5]$.}

\begin{figure}[t]
\begin{center}
\includegraphics[width=0.4\textwidth]{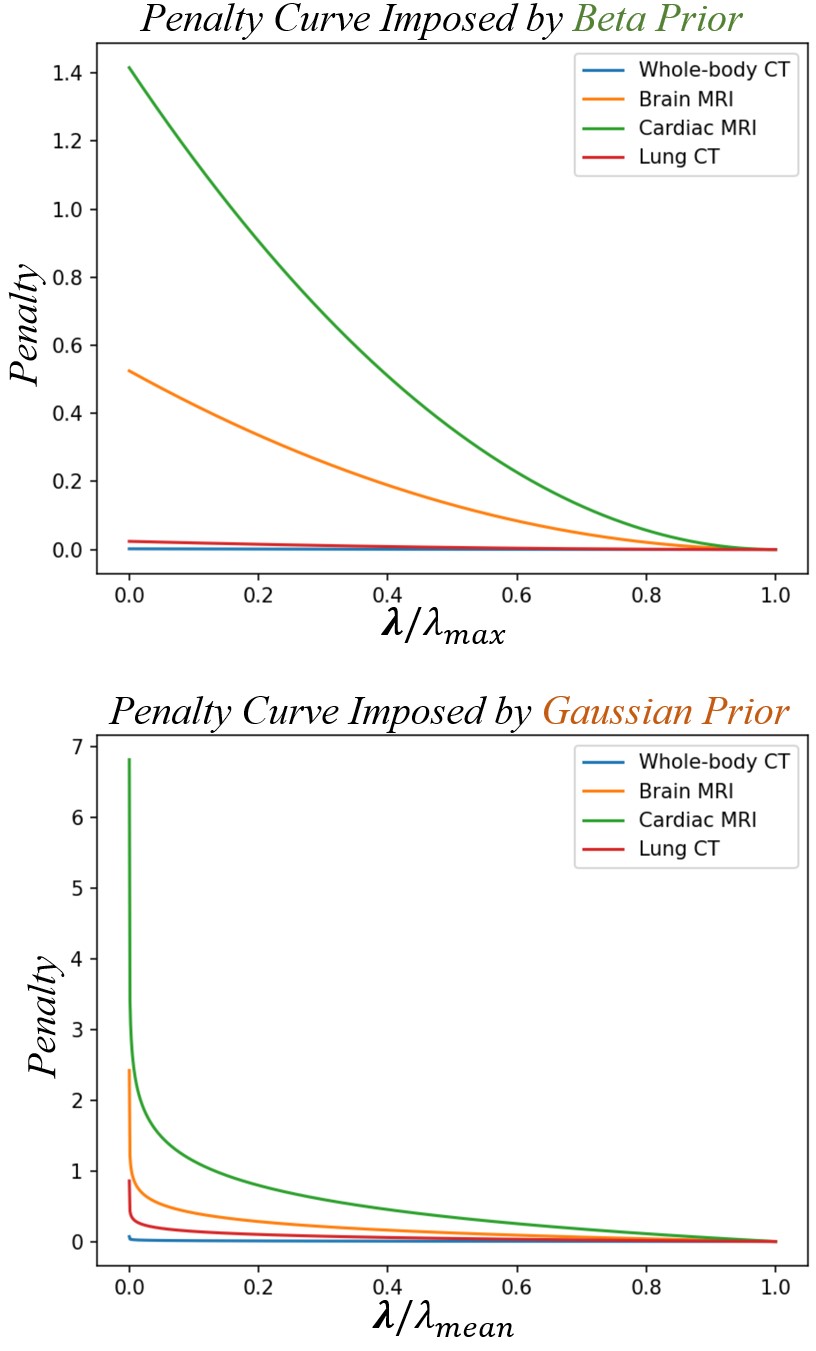}
\end{center}
   \caption{\textcolor{black}{Visualization of the penalty curves imposed by the beta prior (upper) and Gaussian prior (lower). The curves are generated using the optimal hyperparameters obtained from Bayesian optimization for each registration task, corresponding to the results reported in Table~\ref{tab:optimal_hyper}.}}
\label{fig:optimal_penalty}
\end{figure}

A similar pattern appears under the Gaussian prior (bottom row in Fig.~\ref{fig:hyper_contour}). For whole-body CT registration, large $\lambda_{mean}$ values in the range $[2, 3]$ and a small $\sigma'$ close to 0 were favored. This is reflected in the qualitative results in Fig.~\ref{fig:qual_results}, where the spatial weights are nearly zero within the body region. For brain registration, the preferred range for $\sigma'$ was $[0.5, 1.5]$, with $\lambda_{mean}$ within $[3, 4]$. For cardiac registration, $\sigma'$ was preferred within $[0.5, 1.25]$ and $\lambda_{mean}$ within $[1, 2]$. \textcolor{black}{Lastly, lung CT registration favored $\sigma'$ in a narrow range near zero ($[0, 0.05]$) and moderate values of $\lambda_{mean}$ in the range $[0.5, 1]$.} \textcolor{black}{We additionally included Fig.~\ref{fig:optimal_penalty}, which visualizes the penalty curves derived from the optimal hyperparameters for each registration task, as reported in Table~\ref{tab:optimal_hyper}. The figure shows the loss functions associated with the beta prior ($-\log p^{\mathcal{B}e}$) and the normal prior ($-\log p^{\mathcal{N}}$) as functions of the spatial weights $\pmb{\lambda}$. While the shapes of the penalty curves differ due to the underlying distributions, the learned hyperparameters consistently yield high penalties when $\lambda$ is near zero (indicating weak or no regularization), and low penalties when $\lambda$ is close to its maximum (indicating strong regularization). Notably, for whole-body CT registration, the penalty curve remains flat across all values of $\lambda$, indicating that minimal regularization is preferred in this task. This aligns with our qualitative observations in Fig.~\ref{fig:qual_results}, where the weight maps show low values across the body region.}

This variability in the surface maps underscores the necessity of dataset-specific hyperparameter tuning to achieve optimal performance across different tasks, supporting our assertion in the Introduction that different registration tasks favor spatially varying regularization differently at the voxel level. Consequently, no single universal set of hyperparameters is effective in all tasks.

\subsubsection{Hyperparameter Importance for Registration Accuracy}
The contribution of each hyperparameter to the accuracy of the registration is assessed using functional ANOVA~(fANOVA) from the Optuna package~\citep{akiba2019optuna}. This sensitivity analysis technique examines how much registration performance (in our case, the Dice score) changes when each parameter is varied while the others are held constant, yielding a relative importance score for each hyperparameter that sums to 1.

\begin{figure}[t]
\begin{center}
\includegraphics[width=0.48\textwidth]{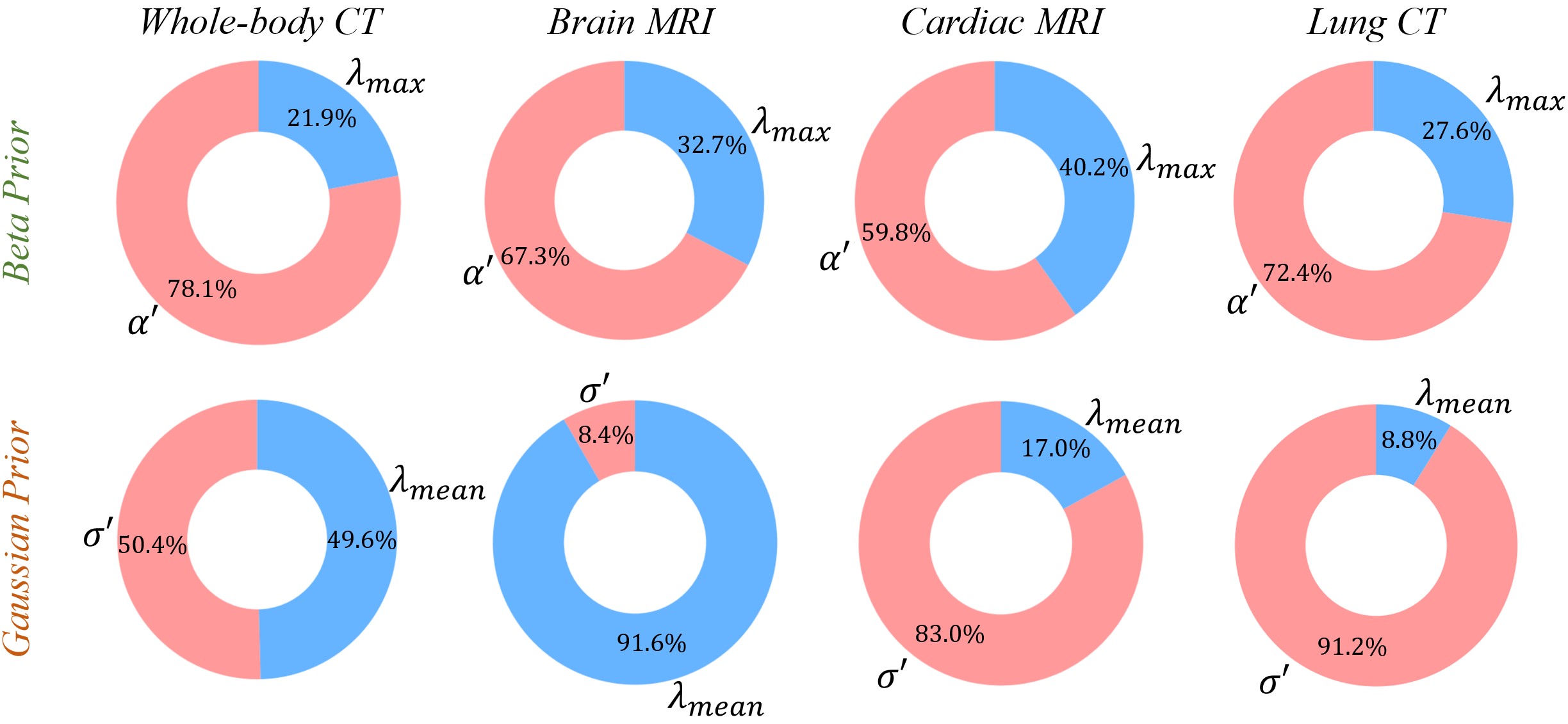}
\end{center}
   \caption{The pie charts illustrate the contribution of hyperparameters for the proposed spatially varying regularization to registration performance, evaluated for two different prior types across three registration tasks.}
\label{fig:hyper_import}
\end{figure}

The importance values for each registration task under two different prior assumptions are presented in Fig.~\ref{fig:hyper_import}. The results indicate that the importance of the hyperparameter varies depending on both the prior assumptions and the registration tasks, consistent with the trends observed in the surface maps.
For the beta prior, $\alpha'$, which controls the spatial coherence of the spatial weight volume $\pmb{\lambda}$, exerts a greater influence on the registration performance than the per-voxel regularization strength $\lambda_{max}$. In contrast, for the Gaussian prior, $\sigma'$ (which also controls spatial coherence) is less influential for brain registration on the IXI dataset than the per-voxel regularization strength $\lambda_{mean}$. However, for whole-body CT registration on the autoPET dataset, $\sigma'$ and $\lambda_{mean}$ are approximately equal importance.

\subsubsection{Impact of Upsampling Spatial Weight Volumes}
\textcolor{black}{
To examine how upsampling the spatial weight volumes influences registration performance, we conducted empirical experiments across different upsampling factors. Spatial weight volumes, $\pmb{\lambda}$, are estimated by a lightweight decoder at lower resolutions and subsequently upsampled to match the original image resolution, as illustrated in Fig.~\ref{fig:ConvNet_for_w}. We evaluated four upsampling factors: $1/8$, $1/4$, $1/2$, and full resolution. The decoder architecture was adjusted accordingly for each resolution.}

\textcolor{black}{Figure~\ref{fig:upsamp} presents the validation performance across the four registration tasks. We observe that upsampling has minimal impact on quantitative metrics for all tasks. This result is consistent with the patterns shown in Fig.~\ref{fig:qual_results}: for whole-body CT, the learned spatial weights are nearly zero inside the anatomical region, whereas for cardiac MRI, they are nearly at the maximum. In both cases, the smoothness of the weight map does not meaningfully affect the final registration outcome.}

\textcolor{black}{For tasks exhibiting more spatial heterogeneity in the spatial weight volumes, such as lung CT, slight differences emerged. Lung CT registration was somewhat more sensitive to the spatial resolution of the weight map, likely due to greater discontinuities inherent in lung deformation. Nevertheless, even for these scenarios, estimating spatial weights at lower resolutions offers significant computational benefits without substantially compromising accuracy.}

\textcolor{black}{Additionally, we provide synthetic illustrative examples in Fig.~\ref{fig:synth_upsamp} in the appendix to visually demonstrate how the choice of upsampling factor influences the resolution of the resulting spatial weight map. These examples clarify that finer-scale discontinuities in the deformation field may require higher-resolution spatial weight maps. However, our empirical results indicate that, in practical medical imaging contexts, the impact of upsampling factors on overall registration performance is relatively minor.}

\subsection{Hyperparameter Learning for Spatially Varying Regularization}
As this article introduces a novel deformation regularization strategy for image registration, our main focus was to rigorously evaluate the effects of the hyperparameters introduced on various registration tasks, with less emphasis on computational efficiency. As discussed earlier, the proposed method likely requires a unique set of optimal hyperparameters for different registration tasks, which poses a limitation due to the need for hyperparameter optimization for each application. Although we employed a more efficient BO approach for hyperparameter tuning, as opposed to a resource-intensive grid search, some level of optimization remains necessary for each specific registration task. Despite pruning many training trials before completing the full cycles, the model still needs to be trained from scratch multiple times during the optimization process.

\begin{figure*}[t]
\begin{center}
\includegraphics[width=0.99\textwidth]{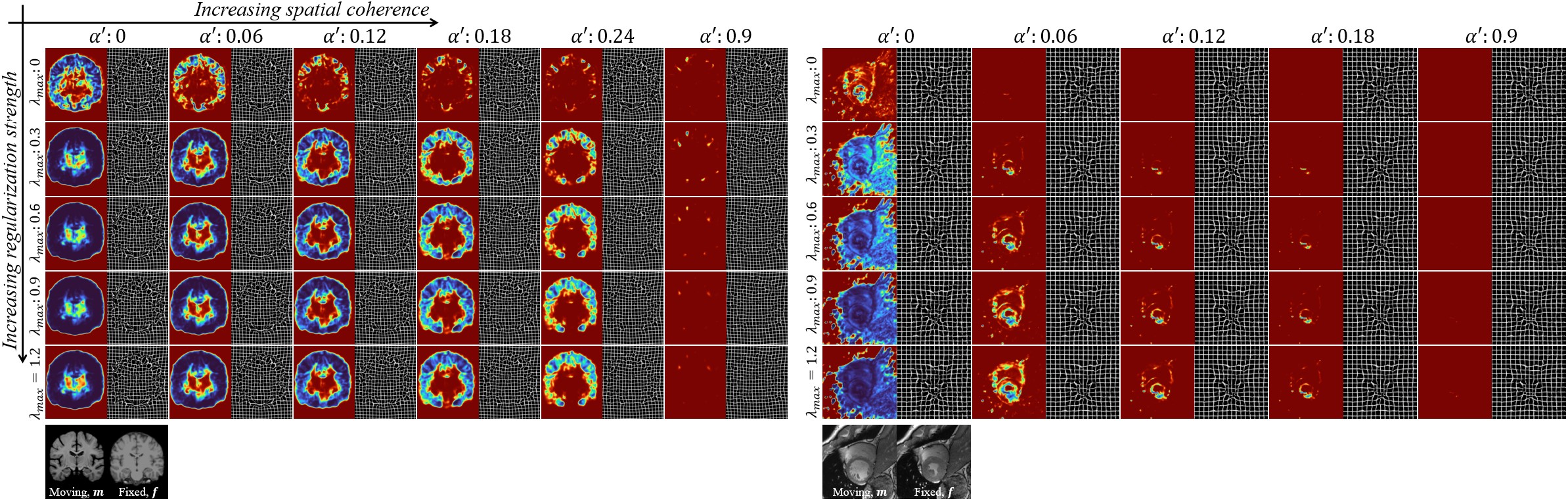}
\end{center}
   \caption{Visualization of spatial weight volumes and corresponding deformation fields through continuous tuning of the hyperparameters ($\alpha'$ and $\lambda_{max}$) using the hyperparameter learning method applied to the proposed spatially varying regularization (i.e., \texttt{Hyper-TM-SPR}$_{\text{\textit{Beta}}}$). The left panel presents results from brain MRI registration (IXI), while the right panel shows results from cardiac MRI registration (ACDC M\&Ms). For the cardiac MRI task, anatomical label maps were incorporated during training to demonstrate the proposed method's applicability to semi-supervised or supervised registration tasks.}
\label{fig:hyper_SPR}
\end{figure*}

In the following subsections, we demonstrate on the brain and cardiac MR image registration tasks how the proposed method can be seamlessly integrated within a hyperparameter learning framework, using the concepts from \texttt{HyperMorph}~\citep{hoopes2021hypermorph}. This setup requires only a single training cycle to enable continuous hyperparameter tuning, thereby avoiding potentially more computationally expensive (though more rigorous) hyperparameter optimization. For both tasks, we learn hyperparameters for spatially varying regularization under the beta prior (i.e., \texttt{Hyper-TM-SPR}$_{\text{\textit{Beta}}}$). To recap, the associated hyperparameters include \textit{1)} $\alpha'$, which controls the spatial coherence of the spatial weight volume $\pmb{\lambda}$, where larger values of $\alpha'$ result in a more spatially uniform regularization strength throughout the image, and \textit{2)} $\lambda_{max}$, which represents the maximum regularization strength applied to a voxel, thereby influencing the overall regularization strength.

While hyperparameter learning facilitates continuous tuning of hyperparameters with fewer training cycles and reduced training times, it is not without limitations. Recent neural network architectures, such as Transformers~\citep{zhai2022scaling, dehghani2023scaling} or large-kernel ConvNets~\citep{liu2022convnet, ding2022scaling, woo2023convnext}, which rely on large model scales (i.e., a substantial number of trainable parameters), introduce significant computational challenges when used as the registration network within the hyperparameter learning framework. Specifically, the auxiliary network responsible for predicting the model parameters also scales up proportionally with the number of layers and parameters in the registration network.

As a result, while hyperparameter learning can potentially decrease the overall training time, the computational demands for a single training cycle are considerably high. For instance, in the case of \texttt{Hyper-TM-SPR} applied to brain MRI registration, the training process required over 70~GB of GPU memory, presenting a significant challenge in practical applications. Although redesigning the auxiliary network to be more memory efficient is possible, it is beyond the scope of this study. This challenge further justify our choice of BO, which retains the same computational demands as running the original registration network. \textcolor{black}{In comparison, training the proposed \texttt{TM-SPR$_{Beta}$} and \texttt{TM-SPR$_{Gaussian}$} models required only 11.7 GB of GPU memory, slightly higher than the 9.8 GB used by the baseline \texttt{TransMorph} network. The increase in memory usage arises primarily from the convolutional decoder that predicts the spatial regularization weights, along with the computation of the auxiliary regularization loss function. Importantly, these computational additions are straightforward, and the decoder can be completely removed during inference since the spatial weight volume is not needed to produce the deformation field. Optimization of the decoder design could further reduce resource requirements, although such refinements are beyond the scope of this study.}

\subsubsection{Unsupervised Registration}
\label{sec:SPR_UnsupReg}
We first demonstrate the application of hyperparameter learning for spatially varying regularization in the brain MRI registration task within an unsupervised learning framework. Similar to the baseline models \texttt{HyperMorph} and \texttt{Hyper-TransMorph}, we employed an auxiliary network composed of fully connected subnetworks to predict the model parameters for \texttt{TM-SPR}${\text{\textit{Beta}}}$ based on the input hyperparameters, represented as a vector containing two scalars: $\alpha'$ and $\lambda_{max}$. The maximum values for $\alpha'$ and $\lambda_{max}$ sampled during training were set to match those used in the previously performed BO, specifically 0.2 for $\alpha'$ and 5 for $\lambda_{max}$. During training, the input values to the auxiliary network were normalized by these maximum values. All other training settings remained consistent with the original \texttt{TM-SPR}$_{\text{\textit{Beta}}}$. This new model is denoted as \texttt{Hyper-TM-SPR}$_{\text{\textit{Beta}}}$ to emphasize its hyperparameter learning capability.

\begin{table}[!t]
\centering
\caption{Quantitative results for hyperparameter learning applied to the proposed spatially varying regularization under the beta prior (denoted as \texttt{Hyper-TM-SPR}$_{\text{\textit{Beta}}}$), compared to models without hyperparameter learning. Results for models without hyperparameter learning are taken from Table~\ref{tab:IXI} and Table~\ref{tab:ACDC}. For cardiac MRI registration, "w/ Seg." denotes that anatomical label maps was incorporated during training for \texttt{Hyper-TM-SPR}$_{\text{\textit{Beta}}}$.}
\fontsize{8.5}{10}\selectfont
    \begin{tabular}{ c  c c c }
 \toprule
 \multicolumn{4}{c}{\textit{Brain MRI Registration (IXI)}}\\
 \cmidrule(lr){1-4}
 Method & Dice$\uparrow$ & \%$\vert J\vert\leq0\downarrow$ & \%NDV$\downarrow$ \\
 \cmidrule(lr){1-1}\cmidrule(lr){2-2}\cmidrule(lr){3-3}\cmidrule(lr){4-4}
 \rowcolor{Gray}
 \texttt{TM-SPR}$_{\text{\textit{Beta}}}$ & 0.767$\pm$0.129 & 0.00\% & 0.00\%\\

 \texttt{Hyper-TM-SPR}$_{\text{\textit{Beta}}}$ & 0.762$\pm$0.127 &0.00\%&0.00\%\\
 
 \hline
 \\[-0.6em]
 \toprule
 \multicolumn{4}{c}{\textit{Cardiac MRI Registration (ACDC M\&Ms)}}\\
 \cmidrule(lr){1-4}
 Method & Dice$\uparrow$ & \%$\vert J\vert\leq0\downarrow$ & \%NDV$\downarrow$ \\
 \cmidrule(lr){1-1}\cmidrule(lr){2-2}\cmidrule(lr){3-3}\cmidrule(lr){4-4}
 \rowcolor{Gray}
 \texttt{TM-SPR}$_{\text{\textit{Beta}}}$ & 0.735$\pm$0.119 & 0.00\% & 0.00\%\\

 \texttt{Hyper-TM-SPR}$_{\text{\textit{Beta}}}$  (w/ Seg.) & 0.848$\pm$0.091 & 0.00\% & 0.03\%\\
 \hline
\end{tabular}
\label{tab:hyperSPR}
\end{table}

After training, different values of $\alpha'$ and $\lambda_{max}$ were continuously sampled to assess their impact on the generated spatial weight volumes and the associated deformation fields. The left panel of Fig.~\ref{fig:hyper_SPR} illustrates these spatial weight volumes and deformation fields for an example image pair, demonstrating the effects of varying hyperparameters. For a fixed $\lambda_{max}$, increasing $\alpha'$ from 0 to 0.9 results in greater spatial coherence within the spatial weight volume, approaching uniform values throughout the image. This observation aligns with theoretical expectations, as the weight of the penalty term (Eqn.~\ref{eqn:log_beta}) that enforces higher spatial weights increases with $\alpha'$. On the other hand, when $\alpha'$ is fixed and $\lambda_{max}$ increases from 0 to 1.2, a different trend is observed. Larger $\lambda_{max}$ values, corresponding to stronger regularization, lead to lower spatial weights in specific regions. For example, at $\lambda_{max} = 0$, only a small portion of the cortical regions is assigned lower spatial weights (indicated by blue regions). As $\lambda_{max}$ increases to 1.2, these regions and their surroundings exhibit progressively darker shades of blue, reflecting further reductions in spatial weights. 

This empirical observation suggests that, while increasing $\lambda_{max}$ imposes a stronger overall regularization by setting a larger upper bound, it also decreases spatial weights, potentially counteracting the intended effect. \textit{To effectively enforce stronger regularization across the entire image, both $\lambda_{max}$ and $\alpha'$ need to be increased simultaneously, moving diagonally in the hyperparameter space.}

We also performed a dense grid search for $\alpha'$ and $\lambda_{max}$ with step sizes of 0.004 and 0.1, respectively, to generate a surface map of Dice scores on the validation data set. The resulting surface map, shown in Fig.~\ref{fig:hyper_IXI}, aligns visually with the surface map obtained using BO (Fig.~\ref{fig:hyper_contour}) for the beta prior on the same dataset. Specifically, the task favors larger values for both $\alpha'$ and $\lambda_{max}$, with the optimal values determined to be 0.212 and 5.4, respectively, which differ slightly from the BO-derived optimal values listed in Table~\ref{tab:optimal_hyper}. The quantitative comparison between \texttt{Hyper-TM-SPR}$_{\text{\textit{Beta}}}$ and \texttt{TM-SPR}$_{\text{\textit{Beta}}}$ (optimized with BO) is presented in Table~\ref{tab:hyperSPR}. While \texttt{Hyper-TM-SPR}$_{\text{\textit{Beta}}}$ achieves a slightly lower mean Dice score of 0.762 compared to 0.767 for \texttt{TM-SPR}$_{\text{\textit{Beta}}}$, the difference is statistically significant, as determined by the Wilcoxon signed-rank test. These results suggest that, while hyperparameter learning is effective for exploring a range of hyperparameter values and is computationally more efficient, BO remains preferable for identifying the optimal hyperparameter values and achieving the best-performing model.

\subsubsection{Semi-supervised Registration}
We next investigate the applicability of the proposed spatially varying regularization for semi-supervised image registration, a widely used approach that incorporates label map supervision during training to enhance registration performance~\citep{chen2024survey}. \textcolor{black}{Importantly, this framework is not limited to label maps or anatomical priors. Other types of prior information can also be incorporated, such as tissue composition (e.g., fat fraction maps, fibrosis probability, or water content from MR imaging) or metabolic activity from nuclear medicine imaging (e.g., PET or SPECT), by including them as additional similarity measures or likelihood terms, analogous to how label maps are used here.} We demonstrate the integration of label maps on the cardiac registration task, as it is the only task where learning-based models did not outperform the optimization-based method, as shown in Table~\ref{tab:ACDC}. Semi-supervised learning provides a distinct advantage for learning-based approaches by leveraging available anatomical label maps to guide the learning process, offering an opportunity to improve registration performance. Furthermore, we demonstrate that the proposed spatially varying regularization is not limited to unsupervised learning but can also be seamlessly integrated into semi-supervised registration frameworks.

The hyperparameter learning framework largely follows the approach described in Sect.~\ref{sec:SPR_UnsupReg}, with the primary difference being the inclusion of an additional loss term. Specifically, a Dice loss is introduced to compare the anatomical label map of the deformed moving image to that of the fixed image. This Dice loss is added to Eqn.~\ref{eqn:loss}, with its weight set equal to that of the image similarity measure, both assigned a value of 1. The hyperparameter values of $\alpha'$ and $\lambda_{max}$ are provided as input to the auxiliary network, which predicts the model parameters for the registration network. The maximum values for $\alpha'$ and $\lambda_{max}$ were set to 0.6 and 3, respectively. Similarly to the unsupervised learning setup in Sect.~\ref{sec:SPR_UnsupReg}, the sampled hyperparameter values were normalized by their maximum values before being input into the network.

The right panel of Fig.~\ref{fig:hyper_SPR} shows the spatial weight volumes and their associated deformation fields obtained for a sample cardiac MRI image pair using various $\alpha'$ and $\lambda_{max}$ values. Despite the additional Dice loss during training, the trends observed are consistent with those in the unsupervised learning task described in Sect.~\ref{sec:SPR_UnsupReg}. Specifically, increasing $\alpha'$ promotes greater spatial uniformity within the spatial weights in the image domain. In contrast, increasing $\lambda_{max}$, which raises the maximum regularization strength applied to individual voxel locations, results in smaller spatial weights in specific regions.

To further investigate the relationship between hyperparameters and registration performance, we performed a dense grid search for $\alpha'$ and $\lambda_{max}$ with step sizes of 0.012 and 0.06, respectively, and generated hyperparameter surface maps based on Dice scores, as shown in Fig.~\ref{fig:hyper_ACDCMM}.
Interestingly, these surface maps differ from those obtained by BO for the unsupervised learning task on cardiac MRI registration (as shown in Fig.~\ref{fig:hyper_contour}).
Specifically, the surface maps for the semi-supervised task favor smaller values for both $\alpha'$ and $\lambda_{max}$, with the optimal values identified as 0.012 and 0.12, respectively, indicating a bias towards minimal regularization. In contrast, the unsupervised task favors relatively larger values for both $\alpha'$ and $\lambda_{max}$.

This discrepancy likely arises from the bias introduced by the Dice loss during training. Since the surface maps were generated using Dice as the performance metric, the regularization inherently skews towards optimizing Dice alone, potentially reducing the influence of regularization. However, the resulting deformation fields remain smooth and almost free of folding, as indicated by "\%$\vert J\vert\leq0$" and "\%NDV" values in Table~\ref{tab:hyperSPR}, due to the enforcement of diffeomorphic registration.

\begin{figure}[t]
\begin{center}
\includegraphics[width=0.48\textwidth]{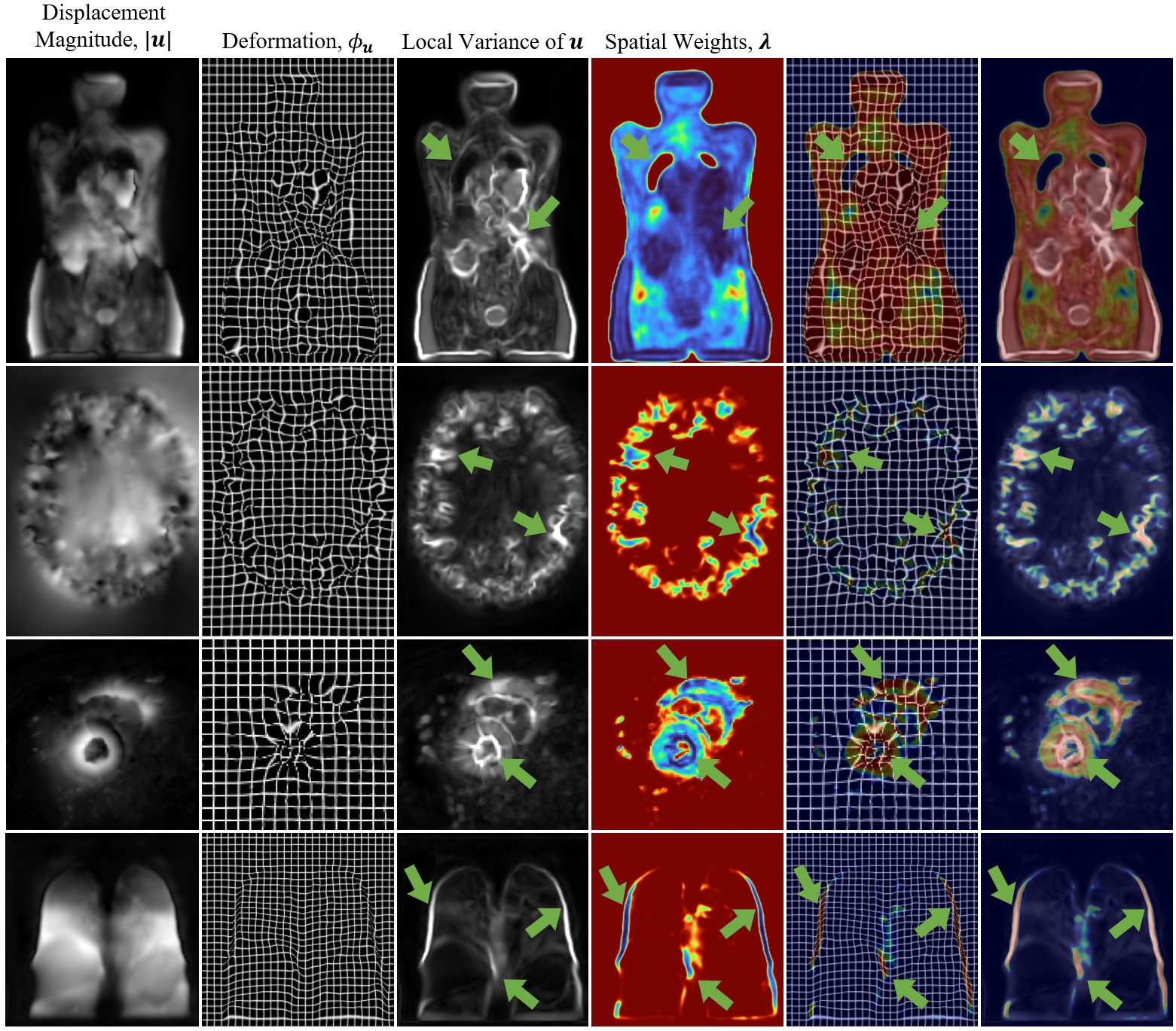}
\end{center}
   \caption{Visualization of spatial weight volumes and deformation fields, highlighting their correlation for improved interpretability. From left to right, the images depict: displacement magnitude, deformation field, local variance map, spatial weight volume, and spatial weight volume overlaid on the deformation field and local variance map (both with inverted colormaps for the spatial weight volume).}
\label{fig:Exp_SPR}
\end{figure}

This finding underscores a critical consideration: \textit{When applying the proposed method to semi-supervised tasks where diffeomorphic constraints are not imposed but Dice loss is used during training, relying solely on Dice as the success metric may lead to small hyperparameter values that impose weak to minimal regularization strength at each voxel location, potentially resulting in nonsmooth deformations that are often undesirable. As Dice is often considered a surrogate measure of registration performance~\citep{rohlfing2011image}, completely abandoning smoothness constraints may exacerbate true registration errors.}

It is worth noting, however, that the proposed method can be extended beyond Dice loss for label map matching. For instance, target registration errors, often considered the gold standard for registration accuracy, can be incorporated as an auxiliary loss function in place of Dice loss within the proposed spatially varying regularization framework when landmarks are available for training.

In terms of quantitative evaluation, Table~\ref{tab:hyperSPR} shows that \texttt{Hyper-TM-SPR}$_{\text{\textit{Beta}}}$, using the optimal hyperparameter values determined by grid search, achieved a significantly higher mean Dice score of 0.848, as measured by the Wilcoxon signed-rank test ($p$-values $\ll$ 0.001), compared to the unsupervised learning model \texttt{TM-SPR}$_{\text{\textit{Beta}}}$, which achieved 0.735. Furthermore, \texttt{Hyper-TM-SPR}$_{\text{\textit{Beta}}}$ outperformed the best performing optimization-based method for the cardiac MRI registration task, \texttt{deedsBCV}, which achieved a mean Dice score of 0.740.

\begin{figure*}[!t]
\begin{center}
\includegraphics[width=0.98\textwidth]{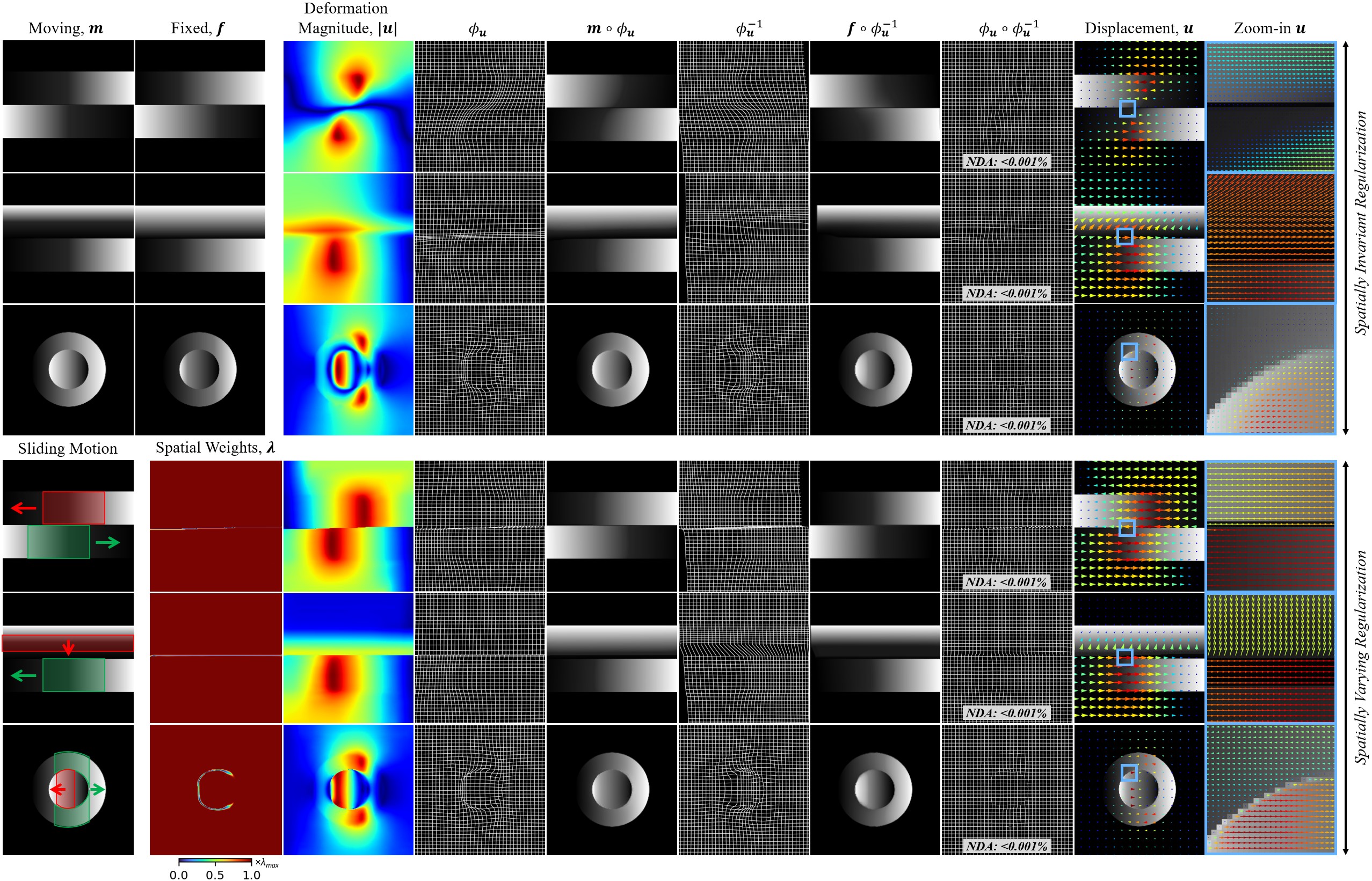}
\end{center}
   \caption{\textcolor{black}{Visualization of 2D synthetic examples illustrating sliding motion. The top panel presents results using spatially invariant regularization, while the bottom panel shows results from the proposed spatially varying regularization. In the top panel, the first two columns display the moving and fixed images. The first column of the lower panel visualizes the sliding motions used to generate the synthetic image pairs. For each method, we show (from left to right): the deformation magnitude, forward deformation field ($\phi_u$), the warped moving image ($m \circ \phi_u$), the inverse deformation field ($\phi_u^{-1}$), the warped fixed image ($f \circ \phi_u^{-1}$), the composition $\phi_u \circ \phi_u^{-1}$, displacement vectors, and a zoomed-in view of the displacement vectors. In the bottom panel, an additional column displays the learned spatial weights $\lambda$. Across all examples, the spatially varying regularization yields sharper transitions at sliding boundaries, effectively capturing discontinuous motion while maintaining invertibility, with non-diffeomorphic area (NDA)~\citep{liu2022finite} remaining below 0.001\%.}}
\label{fig:discontinuity}
\end{figure*}

\subsection{Explainable Image Registration}
Deep learning models are often perceived as "black boxes," lacking interpretability in their predictions, particularly in how they estimate deformation fields from input image pairs. Efforts to address this limitation for image registration have typically focused on exploring interpretability through various approaches, including examining registration uncertainty~\citep{chen2024survey, chen2024from,dalca2019unsupervised,luo2019applicability}, leveraging attention maps derived from self- or cross-attention mechanisms in attention-based models~\citep{shi2022xmorpher,chen2023deformable,liu2024vector}, and leveraging activation maps generated through methods such as gradient-weighted class activation mapping (Grad-CAM)~\citep{selvaraju2017grad,chen2023deformable}.

In this work, the proposed spatially varying regularization, facilitated by the spatial weight volume, introduces a novel framework for interpreting learning-based image registration, complementing existing approaches. The spatial weight volume offers valuable insights into how the deep learning model derives the deformation field. As described in Sect.~\ref{sec:pop_dist}, the spatial weight volume is intrinsically linked to the covariance of the displacement field, conditioned on the input images. This relationship effectively captures the complexity of the deformation field by indicating regions where stricter deformation control (higher spatial weights) or looser control (lower spatial weights) is preferred.

As shown in Fig.~\ref{fig:Exp_SPR} and indicated by the arrows, the spatial weight volume serves as a practical tool to understand how the model adjusts the deformation control to optimize registration performance. Regions with minimal structural differences between moving and fixed images, such as the background or inner lung regions, are assigned high spatial weights to suppress unnecessary deformations. In contrast, regions that require complex deformations, such as the cortical areas in brain images or the edges of cardiac structures, are assigned lower spatial weights, allowing the flexibility needed to accommodate intricate deformations.

However, it is important to note that the interpretability of the spatial weight volume diminishes if the hyperparameter controlling its spatial coherence is set too high. Excessive coherence results in near-uniform values across the entire image, reducing the association between the spatial weight volume and the complexity of the deformation field, thereby limiting its utility as an interpretability tool.

\subsection{Adaptive Regularization}
Prior work has demonstrated that different registration tasks or patient cohorts may require substantially different regularization strengths~\citep{simpson2012probabilistic}. However, existing learning-based methods often lack the ability to produce adaptive regularization specific to a given input image pair. Instead, regularization strength is typically pre-determined before training and remains fixed once the model is trained. Adjusting the regularization strength after training requires retraining the entire registration network from scratch. Although hyperparameter learning methods, such as \texttt{HyperMorph}~\citep{hoopes2021hypermorph}, allow for test-time adjustment of hyperparameters, the regularization itself remains non-adaptive and requires manual tuning for each specific image pair. 

\begin{figure*}[!t]
\begin{center}
\includegraphics[width=0.85\textwidth]{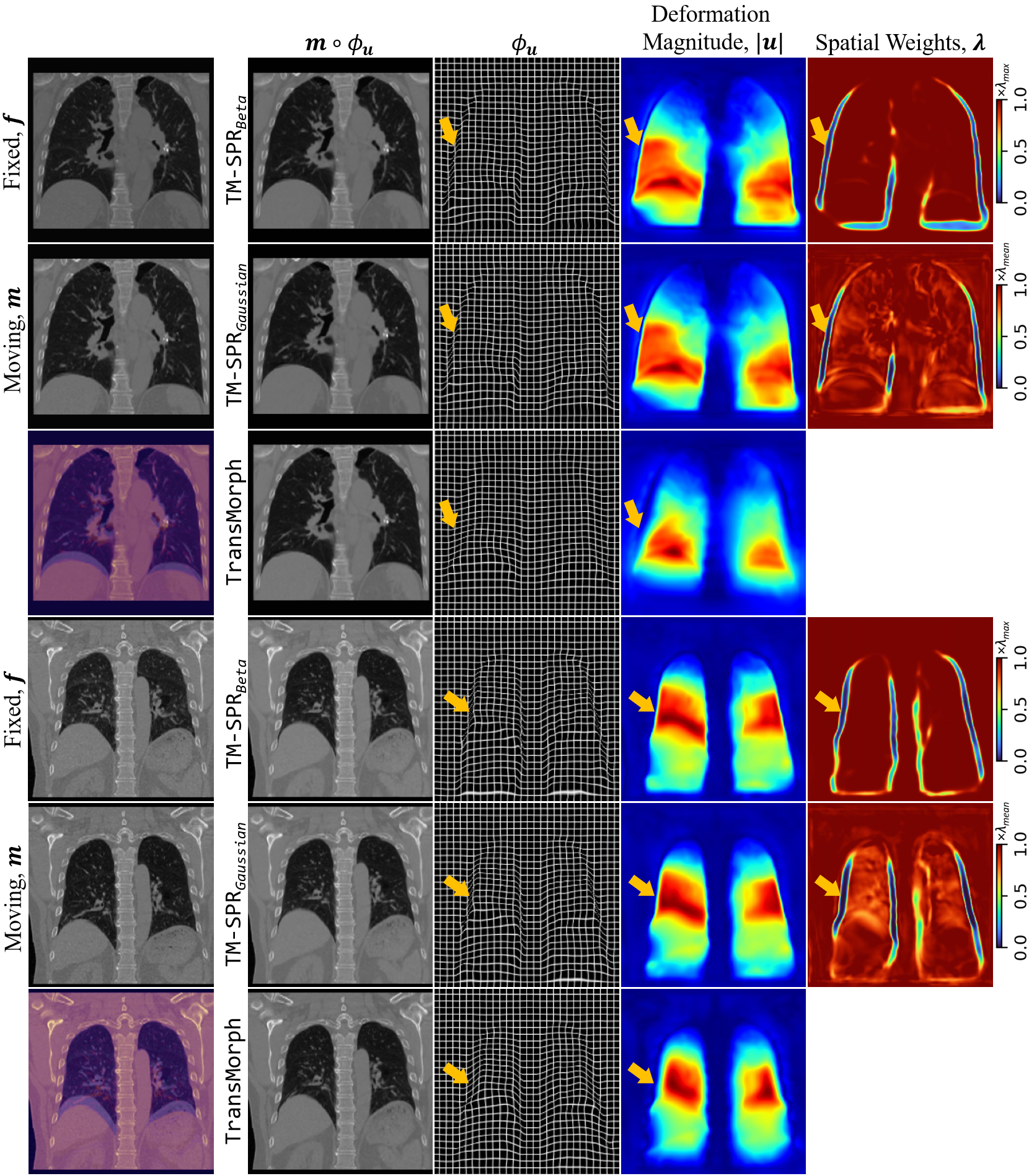}
\end{center}
   \caption{\textcolor{black}{Visualization of lung CT registration in the presence of sliding motion. The top three and bottom three rows correspond to two different image pairs. For each set, the first column shows the fixed image, moving image, and their overlay, highlighting the differences primarily in lung position while the surrounding rib structures remain largely stationary. We compare three methods: \texttt{TM-SPR}$_{\text{\textit{Beta}}}$ and \texttt{TM-SPR}$_{\text{\textit{Gaussian}}}$, both employing the proposed spatially varying regularization with their respective learned spatial weights shown in the last column, and the baseline \texttt{TransMorph}, which uses spatially invariant regularization. The proposed methods clearly capture the sliding motion along organ boundaries, as evidenced by the deformation grids, magnitude maps, and learned spatial weights.}}
\label{fig:discontinuity_lung}
\end{figure*}

In contrast, the proposed spatially varying regularization is inherently adaptive and tailored to each pair of images. While it involves two tunable hyperparameters---controlling the maximum regularization strength in the case of the beta prior, or the mean regularization strength in the case of the Gaussian prior, as well as the expected degree of spatial variation---these hyperparameters govern the framework rather than the specific regularization applied. The actual regularization strength is adaptively predicted for each image pair during the registration process, allowing a more flexible and precise alignment tailored to the given data.

Moreover, the proposed framework is designed to integrate seamlessly with most existing learning-based registration methods. With only minor modifications to the network architecture to incorporate a lightweight module for estimating the spatial weight volume, as well as an adaptable loss function compatible with various image similarity measures and deformation regularization strategies, this method offers a highly versatile enhancement to augment learning-based image registration.

\subsection{Discontinuity-Preserving Registration}
\label{sec:discont_reg}
Discontinuity-preserving image registration has been an important focus in medical image registration research. This is motivated by the observation that deformations in the human body do not always follow continuous motion assumptions. \textcolor{black}{At certain anatomical boundaries, abrupt changes in motion may occur. The most common example is respiratory motion, where lung tissue exhibits sliding behavior along the pleural interface, and adjacent structures such as the lung and rib bones move in different directions. Beyond sliding motion, discontinuous deformation may also be observed in scenarios such as tissue tearing in traumatic injuries, abrupt anatomical shifts following surgical intervention, or regions affected by pathological changes such as tumor resection or necrosis, where local displacements do not follow continuous trajectories.} Traditional global smoothing regularizers, such as the diffusion regularizer, are not well-suited to handle such scenarios. To address this limitation, previous studies have proposed modifications to the regularizer formulation~\citep{schmidt2009slipping,schmidt2012estimation,risser2013piecewise,pace2013locally}, or adaptations to the registration framework that explicitly accommodate discontinuities in the deformation field~\citep{zheng2024residual, chen2021deep}.

The proposed spatially varying regularization inherently supports discontinuities in the deformation field, while still accommodating diffeomorphic constraints when required. \textcolor{black}{Although diffeomorphism and discontinuity may seem incompatible, this contradiction arises only in the continuous domain. A diffeomorphic transformation is a smooth and invertible mapping with a smooth inverse, often generated by exponentiating a stationary velocity field. However, when applied to a discretely sampled image grid, such smooth deformations can appear discontinuous due to interpolation artifacts and limited resolution. For example, a diffeomorphic mapping may smoothly compress a large region into a very small one. If the resulting region is smaller than a voxel, the deformation will appear abrupt or sharp at the image level, even though it remains smooth and invertible in the continuous domain. Our method adopts a time-stationary velocity field framework to ensure diffeomorphic transformations via the scaling-and-squaring method. Within this framework, the spatial weight map controls where and how much regularization is enforced. By assigning near-zero weights at anatomical boundaries, the network is able to accommodate sharp displacement transitions in these regions. Although these transitions may appear discontinuous on the grid, the overall deformation remains diffeomorphic and globally invertible.} 

\textcolor{black}{We begin by demonstrating the deformation capability of our method in preserving discontinuities using 2D synthetic images that exhibit sliding motion, as shown in Fig.~\ref{fig:discontinuity}. The first two columns in the upper panel display the moving and fixed images, while the first column in the lower panel illustrates the sliding motion that transforms the moving image to the fixed image. The remaining columns in the upper panel show results from the baseline \texttt{TransMorph} model with spatially invariant regularization. As expected, the resulting deformation fields remain smooth but fail to capture the sliding motion, instead forcing a gradual, isotropic transition. In contrast, when spatially varying regularization is applied (as shown in the lower panel of Fig.~\ref{fig:discontinuity}), the learned spatial weights---estimated in an unsupervised manner---exhibit low values at object boundaries, enabling the deformation fields to appear discontinuous in the discrete image grid. Despite this apparent discontinuity, the deformation remains invertible and diffeomorphic, with only a small percentage of non-diffeomorphic areas (NDAs), owing to the use of velocity field exponentiation via the scaling-and-squaring method.}

\textcolor{black}{We further demonstrate the proposed method's ability to handle discontinuous motion in real-world 3D lung CT registration, using two qualitative examples shown in Fig.~\ref{fig:discontinuity_lung}. In both cases, the lung exhibits sliding motion, where the lung tissue contracts while adjacent anatomical structures, such as the rib bones, remain largely stationary. With spatially invariant regularization (\texttt{TransMorph}), the network is constrained by global smoothness and can only deform the inner lung region, gradually reducing the deformation magnitude near the boundary to avoid abrupt changes and preserve rib alignment. In contrast, the proposed spatially varying regularization models, i.e., \texttt{TM-SPR}$_{\text{\textit{Beta}}}$ and \texttt{TM-SPR}$_{\text{\textit{Gaussian}}}$, learn spatial weight maps with low values at the lung interface. This allows the network to model sharp displacement transitions and better represent the sliding motion across organ boundaries. These differences are visually highlighted with yellow arrows in Fig.~\ref{fig:discontinuity_lung}.}

Compared to existing regularizer-based approaches that handle discontinuities, the proposed method offers a clear advantage by adaptively learning the regularization strength directly from the data. Furthermore, discontinuities can be fully suppressed by enforcing spatial uniformity in the spatial weight volume, making the method highly flexible and adaptable to varying registration tasks.

\section{Conclusion}
\label{sec:conclusion}
In this study, we introduced an end-to-end, unsupervised framework for learning spatially varying regularization in medical image registration. Unlike existing learning-based methods that typically rely on spatially invariant regularization, the proposed approach generates a spatial weight volume that assigns individual weights to voxels, enabling spatially adaptive regularization. We also proposed a novel loss function formulation that incorporates two hyperparameters designed to compel the network to apply stronger regularization based on a specified maximum or mean regularization strength. Additionally, we introduced efficient hyperparameter tuning methods through Bayesian optmization and hyperparameter learning strategies to allow adjustment of these parameters during the test time.

The effectiveness of the proposed framework was demonstrated on three publicly available datasets that encompass diverse registration tasks and anatomical regions. Both qualitative and quantitative results highlight the advantages of the spatially varying regularization framework, positioning it as a valuable complement to existing learning-based medical image registration methods.

\section*{Disclosures}
The authors have no relevant financial interests or conflicts of interest to disclose related to the content of this paper. 

AI (OpenAI ChatGPT-4o) was used for grammar and language refinement. All scientific content, ideas, and references were developed by the authors and carefully reviewed to ensure accuracy and integrity.

\section*{Acknowledgments}
Junyu Chen and Yong Du were supported by grants from the National Institutes of Health~(NIH), United States, P01-CA272222~(PI: G.~Sgouros), R01-EB031023~(PI: Y.~Du), and U01-EB031798~(PI: G.~Sgouros).
Shuwen Wei, Yihao Liu, and Aaron Carass were supported by the NIH from National Eye Institute grants R01-EY024655~(PI:~J.L.~Prince) and R01-EY032284~(PI:~J.L.~Prince), as well as the National Science Foundation grant 1819326~(Co-PI: S.~Scott, Co-PI: A.~Carass).
Harrison Bai was supported by the NIH from National Cancer Institute grant R03-CA286693~(Co-PI: B.~Kimia, Co-PI: H.~Bai).
Junyu Chen and Harrison Bai were also supported by the United States Department of Defense grant HT94252510807 (PI: H.~Bai).
The views expressed in written conference materials or publications and by speakers and moderators do not necessarily reflect the official policies of the NIH; nor does mention by trade names, commercial practices, or organizations imply endorsement by the U.S. Government.

\onecolumn 
\appendix

\section{Delta Variational Bayes Derivation of the MAP Formulation}
\label{sec:derivation_DVB}

\textcolor{black}{
Let $q_\phi(\pmb{u},\pmb{\Lambda}|\pmb{f};\pmb{m})$ denote the variational approximation to the joint posterior. 
We model it in the delta limit as a point mass centered at the neural network outputs:
\[
q_\phi(\pmb{u},\pmb{\Lambda}|\pmb{f};\pmb{m}) 
\;\Rightarrow\; \delta\!\big((\pmb{u},\pmb{\Lambda}) - (h_\psi(\pmb{m},\pmb{f}),\,g_\theta(\pmb{m},\pmb{f}))\big),
\]
where the mappings
\[
h_\psi:\ (\pmb{m},\pmb{f}) \mapsto \hat{\pmb{u}}, 
\qquad 
g_\theta:\ (\pmb{m},\pmb{f}) \mapsto \hat{\pmb{\Lambda}}
\]
produce deterministic point estimates. The two networks, $h_\psi$ and $g_\theta$, share an encoder but have separate decoders,
consistent with our architecture design as shown in Fig.~\ref{fig:framework}.}

\textcolor{black}{
To formalize the delta limit, we first introduce a reparameterized location–scale family:
\[
q_\phi(\pmb{u},\pmb{\Lambda}|\pmb{f};\pmb{m})
= \mathcal{N}\!\Big(
\begin{bmatrix} h_\psi(\pmb{m},\pmb{f}) \\ g_\theta(\pmb{m},\pmb{f}) \end{bmatrix},
\;\tau^2 I\Big),
\]
with $\tau>0$ a variance control. As $\tau\to 0$, the distribution collapses to a delta concentrated at 
\[
(\pmb{u},\pmb{\Lambda}) \to \big(\hat{\pmb{u}},\,\hat{\pmb{\Lambda}}\big) 
= \big(h_\psi(\pmb{m},\pmb{f}),\, g_\theta(\pmb{m},\pmb{f})\big).
\]}

\textcolor{black}{
The ELBO for this joint factor is
\begin{equation*}\label{eq:joint-elbo}
\begin{aligned}
\mathcal{F}(\phi)
&= \E_{q_\phi}\!\big[\log p(\pmb{f}| \pmb{u},\pmb{\Lambda};\pmb{m})\big]
+ \E_{q_\phi}\!\big[\log p(\pmb{u}| \pmb{\Lambda})\big]
+ \E_{q_\phi}\!\big[\log p(\pmb{\Lambda})\big] - \E_{q_\phi}\!\big[\log q_\phi(\pmb{u},\pmb{\Lambda}| \pmb{f};\pmb{m})\big] \\
&= \E_{q_\phi}\!\big[\log p(\pmb{f}| \pmb{u},\pmb{\Lambda};\pmb{m})\big]
+ \E_{q_\phi}\!\big[\log p(\pmb{u}, \pmb{\Lambda})\big] - \E_{q_\phi}\!\big[\log q_\phi(\pmb{u},\pmb{\Lambda}|\pmb{f};\pmb{m})\big] \\
&= \E_{q_\phi}\!\big[\log p(\pmb f| \pmb u,\pmb\Lambda;\pmb m)\big]
+ \E_{q_\phi}\!\big[\log p(\pmb u,\pmb\Lambda)\big]
+ H\big(q_\phi(\pmb{u},\pmb{\Lambda}| \pmb{f};\pmb{m})\big)\\
&= \E_{q_\phi}\!\big[\log p(\pmb{f}| \pmb{u},\pmb{\Lambda};\pmb{m})\big]
- \KL\!\big(q_\phi(\pmb{u},\pmb{\Lambda}| \pmb{f};\pmb{m})||p(\pmb{u},\pmb{\Lambda})\big).
\end{aligned}
\end{equation*}}

\textcolor{black}{
For our Gaussian approximation with covariance $\Sigma_\tau=\tau^2 I$ and $d=\dim(\pmb u)+\dim(\pmb\Lambda)$,
\[
H\!\big(q_\phi(\pmb{u},\pmb{\Lambda}\mid \pmb{f};\pmb{m})\big)
= \tfrac{1}{2}\log\!\big((2\pi e)^d|\Sigma_\tau|\big)
= \tfrac{1}{2}\log\!\big((2\pi e)^d\big) + d\log\tau,
\]}
\textcolor{black}{
which depends only on $\tau$ and not on $q_\phi$. 
Therefore, it is constant with respect to optimization and can be omitted. 
Substituting into the ELBO gives:
\begin{equation*}
\begin{aligned}
\mathcal{F}(\phi)
&= \E_{q_\phi}\!\big[\log p(\pmb{f}\mid \pmb{u},\pmb{\Lambda};\pmb{m})\big]
+ \E_{q_\phi}\!\big[\log p(\pmb{u}\mid \pmb{\Lambda})\big]
+ \E_{q_\phi}\!\big[\log p(\pmb{\Lambda})\big] + H(q_\phi)\\
&= \log p(\pmb{f}\mid \hat{\pmb{u}},\hat{\pmb{\Lambda}};\pmb{m})
+ \log p(\hat{\pmb{u}}\mid \hat{\pmb{\Lambda}})
+ \log p(\hat{\pmb{\Lambda}}) + \text{const.}
\end{aligned}
\end{equation*}
}

\textcolor{black}{
Hence, maximizing $\mathcal{F}(\phi)$ is equivalent to maximizing the finite remainder:
\begin{equation*}
\log p(\pmb{f}\mid \hat{\pmb{u}},\hat{\pmb{\Lambda}};\pmb{m})
+ \log p(\hat{\pmb{u}}\mid \hat{\pmb{\Lambda}})
+ \log p(\hat{\pmb{\Lambda}}),
\end{equation*}
which is exactly the joint negative log-posterior evaluated at the network outputs (i.e., Eqn.~\ref{eqn:log_post}). Thus, the networks $(h_\psi,g_\theta)$ implement an \emph{amortized MAP} estimator of $(\pmb{u},\pmb{\Lambda})$: the dependence $\hat{\pmb{\Lambda}} = g_\theta(\pmb{m},\pmb{f})$ arises from the posterior mode, while the priors $p(\pmb{\Lambda})$ and $p(\pmb{u}|\pmb{\Lambda})$ remain \textit{input-independent}.}

\newpage
\section{Additional Qualitative and Quantitative Results for Whole-body CT Registration Task}

\begin{figure*}[!htp]
\begin{center}
\includegraphics[width=0.95\textwidth]{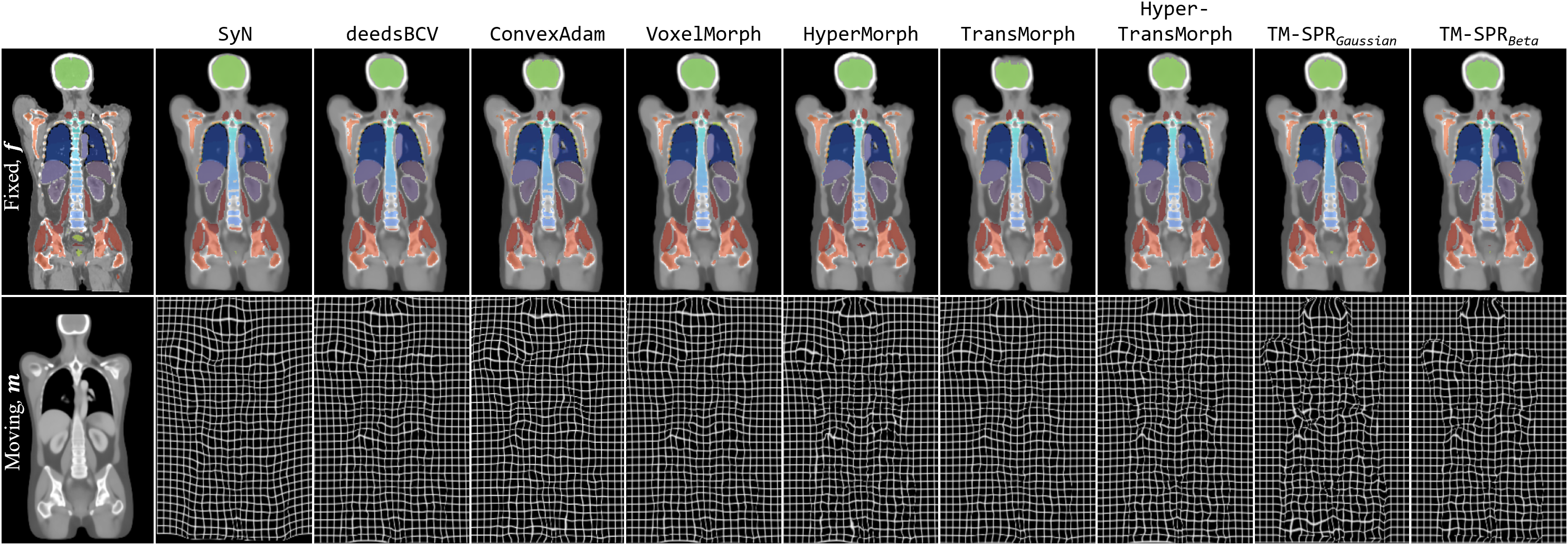}
\end{center}
   \caption{Qualitative comparison of registration methods for whole-body CT registration on the \textit{autoPET} dataset.}
\label{fig:autopet_qual}
\end{figure*}

\begin{figure*}[!htp]
\begin{center}
\includegraphics[width=0.95\textwidth]{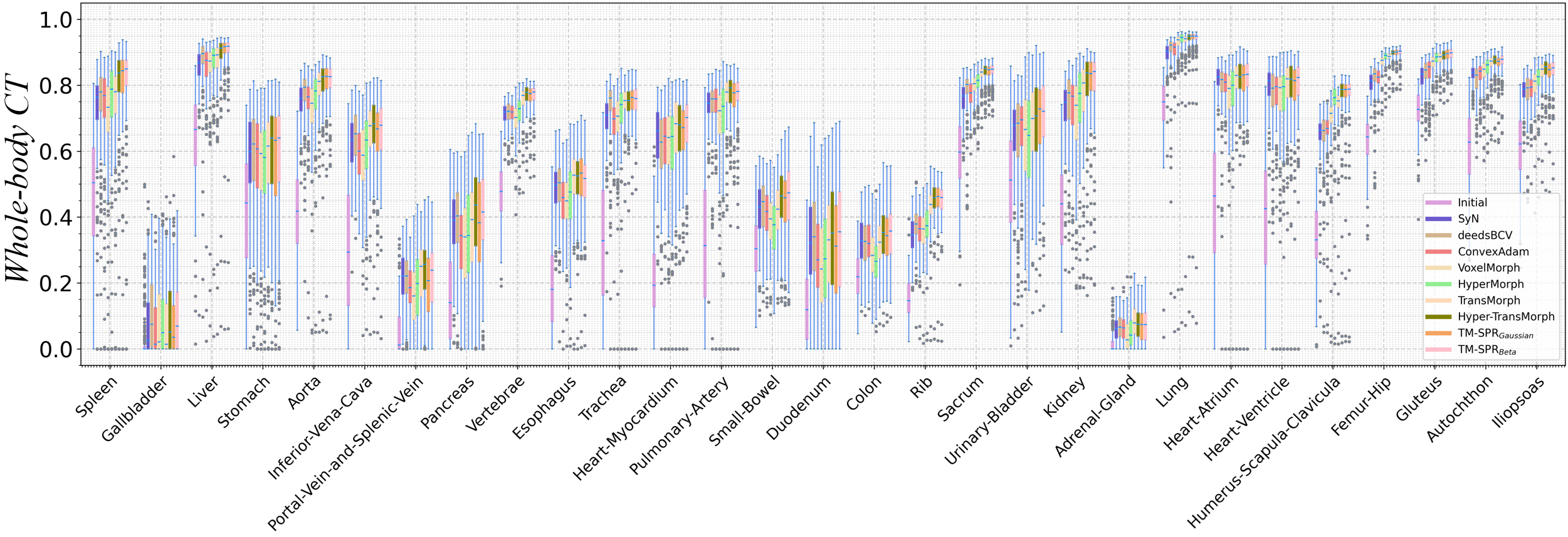}
\end{center}
   \caption{Quantitative comparison of registration methods for whole-body CT registration on the \textit{autoPET} dataset. Boxplots illustrate the Dice scores achieved for different organs in CT images.}
\label{fig:autopet_box}
\end{figure*}

\newpage
\section{Additional Qualitative and Quantitative Results for Brain MRI Registration Task}
\begin{figure*}[!htp]
\begin{center}
\includegraphics[width=0.95\textwidth]{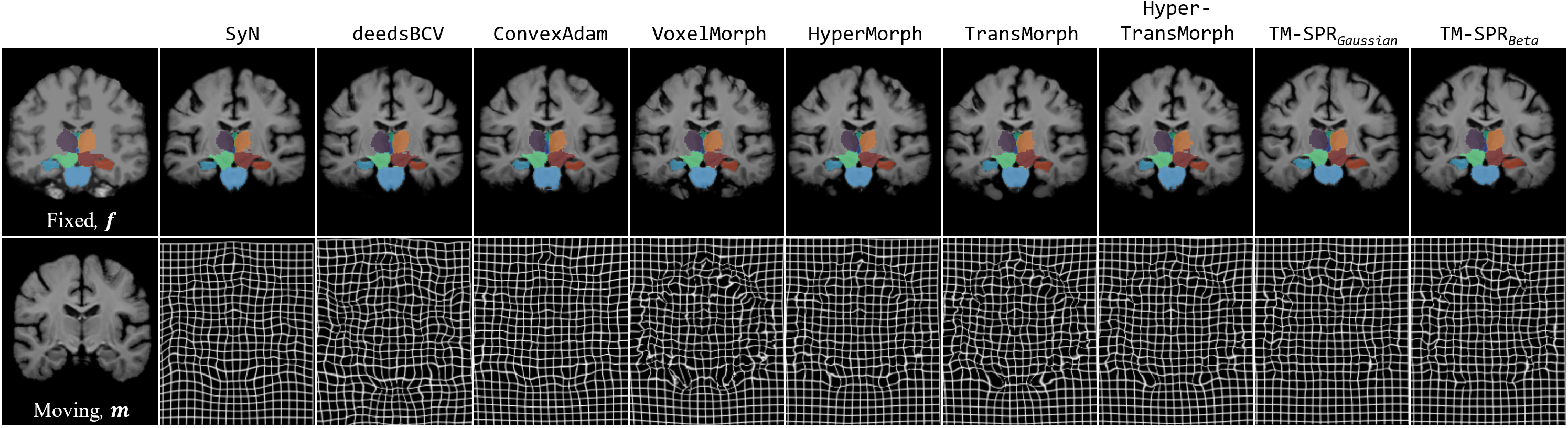}
\end{center}
   \caption{Qualitative comparison of registration methods for brain MRI registration on the \textit{IXI} dataset.}
\label{fig:ixi_qual}
\end{figure*}

\begin{figure*}[!htp]
\begin{center}
\includegraphics[width=0.95\textwidth]{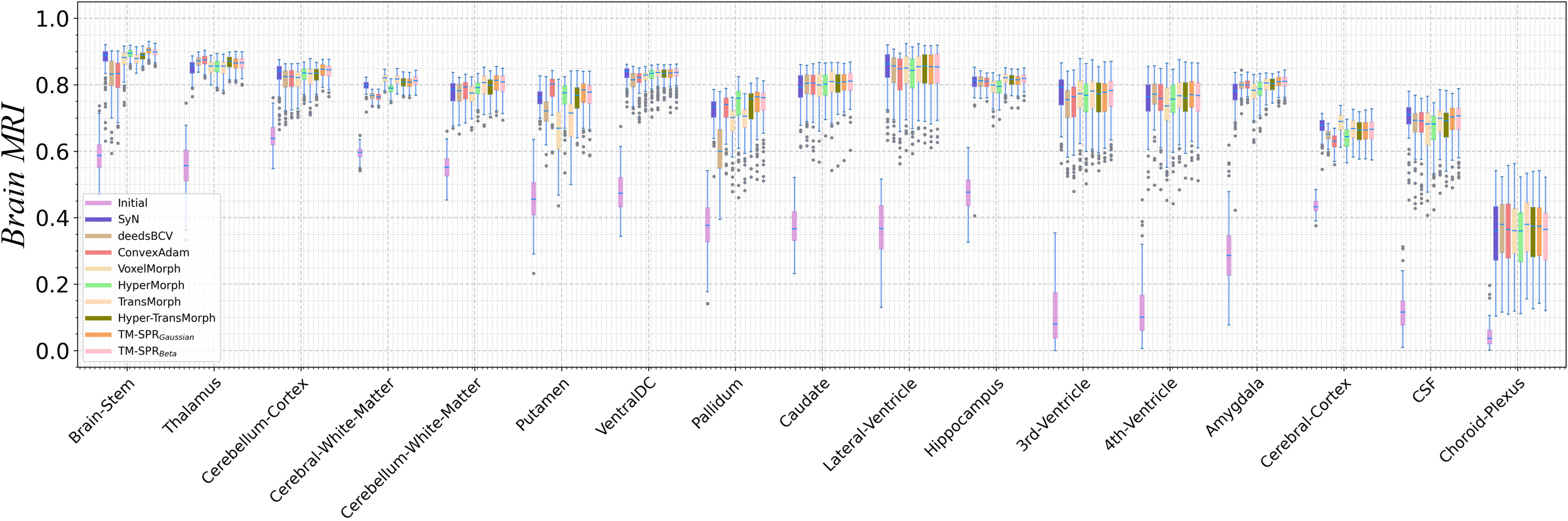}
\end{center}
   \caption{Quantitative comparison of registration methods for brain MRI registration on the \textit{IXI} dataset. Boxplots illustrate the Dice scores achieved for different brain structures in MRI images.}
\label{fig:ixi_box}
\end{figure*}

\section{Additional Qualitative and Quantitative Results for Cardiac MRI Registration Task}
\begin{figure*}[!htp]
\begin{center}
\includegraphics[width=0.95\textwidth]{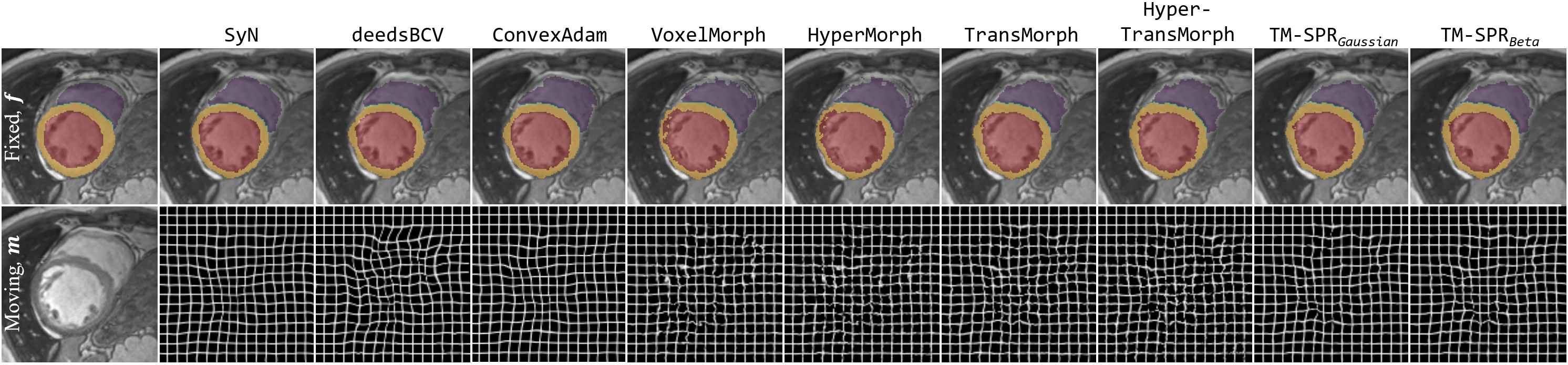}
\end{center}
   \caption{Qualitative comparison of registration methods for cardiac MRI registration on the \textit{ACDC and M\&Ms} dataset.}
\label{fig:acdcmm_qual}
\end{figure*}

\begin{figure*}[!htp]
\begin{center}
\includegraphics[width=0.95\textwidth]{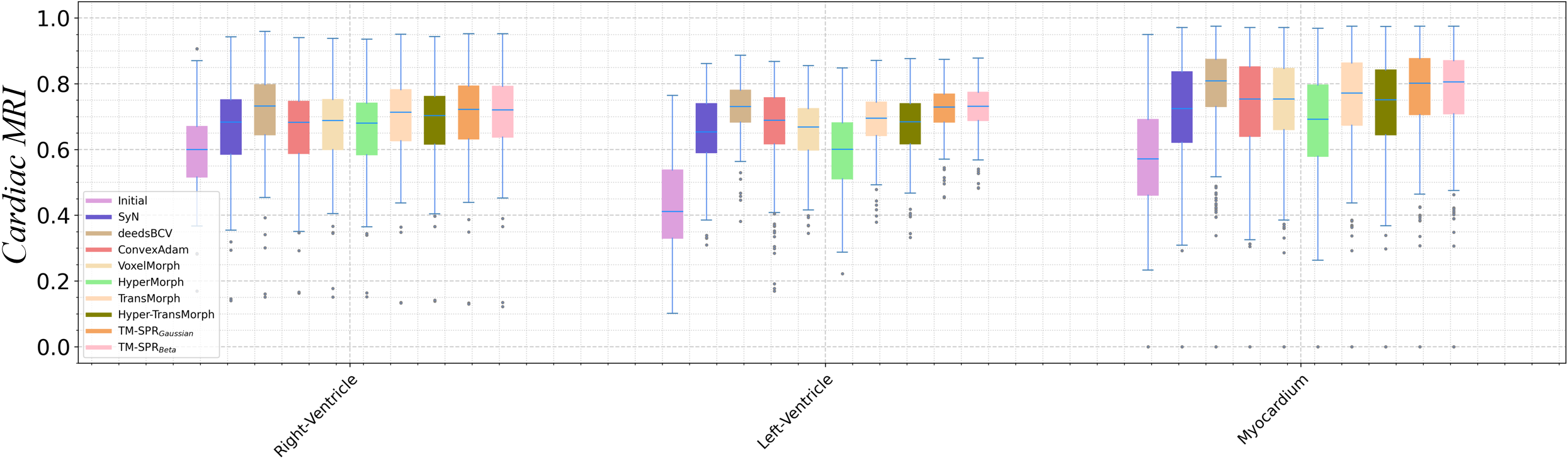}
\end{center}
   \caption{Quantitative comparison of registration methods for cardiac MRI registration on the \textit{ACDC and M\&Ms} dataset. Boxplots illustrate the Dice scores achieved for different cardiac structures in MRI images.}
\label{fig:acdcmm_box}
\end{figure*}

\newpage

\section{\textcolor{black}{Additional Qualitative Results for Lung CT Registration Task}}
\begin{figure*}[!htp]
\begin{center}
\includegraphics[width=0.95\textwidth]{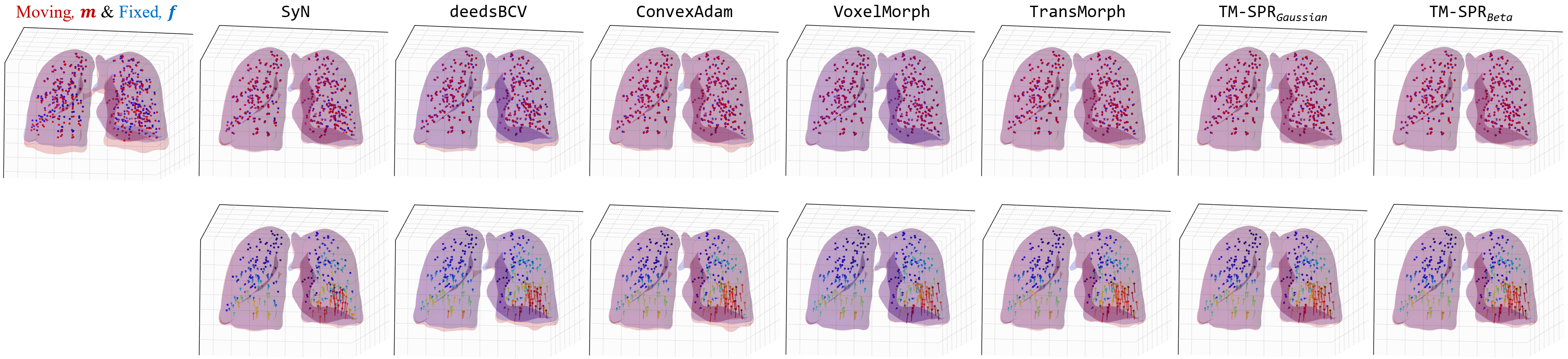}
\end{center}
   \caption{\textcolor{black}{Qualitative comparison of registration methods for lung CT registration on the \textit{4DCT} dataset. The first image on the left in the top row shows the lung mask and 300 landmarks, with red indicating the moving image and blue representing the fixed image. The remaining images in the top row show the displaced landmarks and warped lung masks for each registration method, where red denotes the warped results, while blue continues to represent the fixed image. The bottom row visualizes the corresponding landmark displacements, with color indicating the displacement magnitude.}}
\label{fig:lungct_qual}
\end{figure*}

\newpage
\section{Hyperparameter Landscape for HyperMorph and Hyper-TransMorph}
\begin{figure*}[!htp]
\begin{center}
\includegraphics[width=0.95\textwidth]{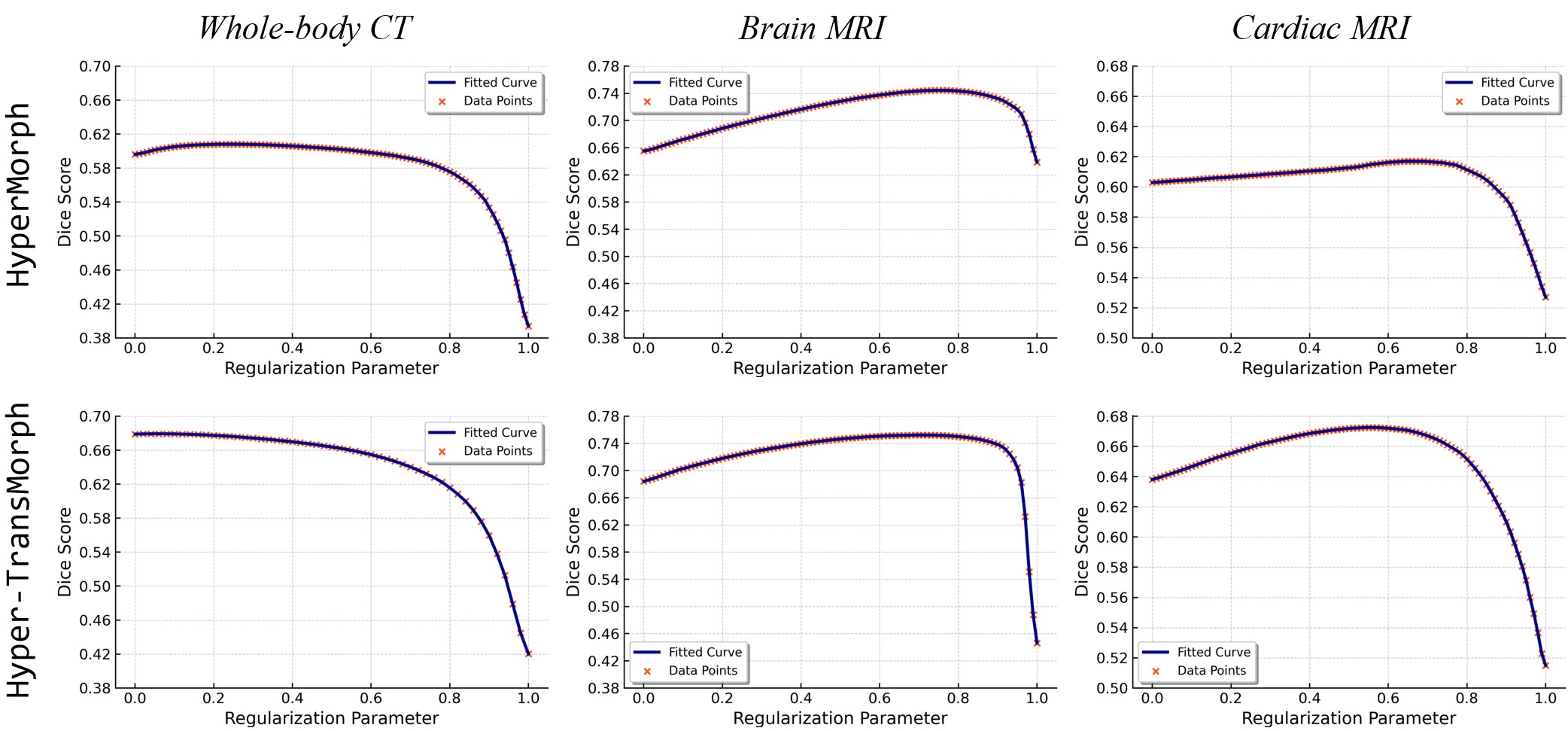}
\end{center}
   \caption{The plots show the grid search results for the regularization hyperparameter (with a step size of 0.01) and its impact on registration performance, as measured by the mean Dice scores. The top row corresponds to \texttt{HyperMorph}, while the bottom row corresponds to \texttt{Hyper-TransMorph}, evaluated across the three registration tasks.}
\label{fig:hypermorph}
\end{figure*}

\section{Effect of Upsampling Spatial Weight Volumes on Synthetic Images}
\begin{figure*}[!htp]
\begin{center}
\includegraphics[width=0.6\textwidth]{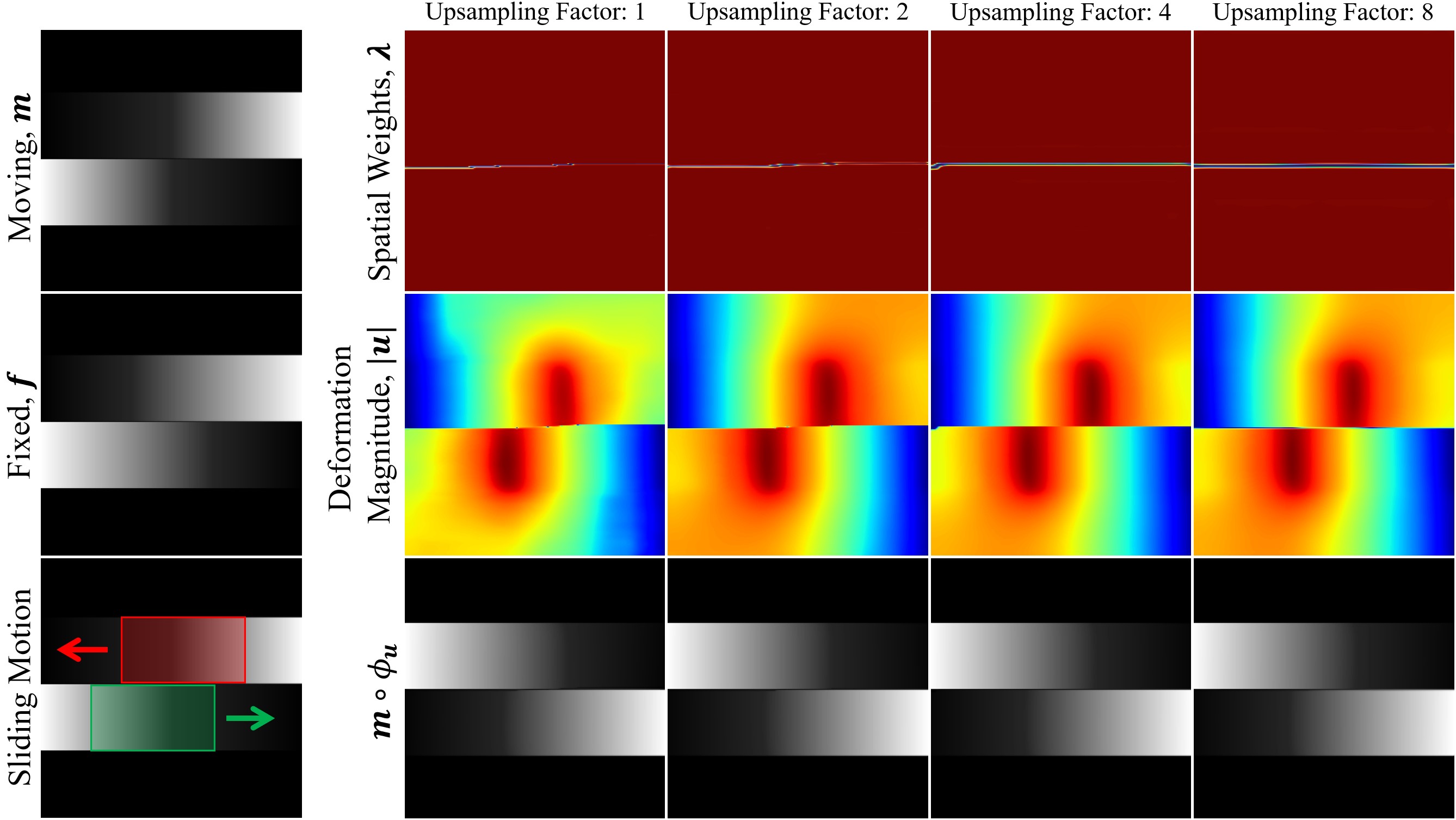}
\end{center}
   \caption{\textcolor{black}{Visualization of the spatial weight maps, deformation magnitudes, and corresponding deformed images under different upsampling factors, as defined by the network architecture in Fig.~\ref{fig:ConvNet_for_w}.}}
\label{fig:synth_upsamp}
\end{figure*}

\twocolumn
\bibliographystyle{model2-names.bst}
\biboptions{authoryear}
\bibliography{sample}

\end{document}